%% file: main.tex
\documentclass[sigconf]{acmart}

\setcopyright{none}
\renewcommand\footnotetextcopyrightpermission[1]{}
\input{math_commands.tex}

\usepackage{colortbl}       % \rowcolor / \cellcolor (xcolor[table] equivalent)
\usepackage{amsfonts}
\usepackage{nicefrac}
\usepackage{mathtools}
\usepackage{amsthm}
\usepackage{subcaption}
\usepackage[capitalize,noabbrev]{cleveref}
\usepackage{tabularx}
\usepackage{array}
\usepackage{threeparttable}
\usepackage{threeparttablex}
\usepackage{adjustbox}
\usepackage{tabularray}
\usepackage{nicematrix}
\usepackage{pifont}
\newcommand{\cmark}{\ding{51}}
\newcommand{\xmark}{\ding{55}}
\usepackage{multirow}
\usepackage{wrapfig}
\usepackage{enumitem}
\usepackage{algorithm}
\usepackage{algorithmic}
\usepackage{comment}
\usepackage{tikz}
\usepackage{longtable}
\usepackage{float}
\usepackage{placeins}    % \FloatBarrier (not a page break — flushes pending floats)

\theoremstyle{plain}

\theoremstyle{definition}

\theoremstyle{remark}

\extrafloats{200}

\definecolor{LightYellow}{RGB}{255, 255, 204}
\definecolor{LightGreen}{RGB}{220, 255, 220}
\definecolor{LightGray1}{RGB}{240, 240, 240}
\definecolor{LightGray2}{RGB}{230, 230, 230}
\definecolor{LightGray3}{RGB}{220, 220, 220}
\definecolor{LightGray4}{RGB}{210, 210, 210}
\definecolor{LightTextGray}{RGB}{100,100,100}

\definecolor{darkblue}{RGB}{0,0,139}
\definecolor{darkgreen}{RGB}{0,100,0}
\definecolor{myred}{rgb}{0.9,0.1,0.1}
\definecolor{myblue}{rgb}{0.1,0.3,0.9}
\definecolor{codeblue}{rgb}{0.25,0.5,0.5}
\definecolor{rebuttalred}{RGB}{200,0,0}

\newcommand{\first}[1]{\textbf{\textcolor{black}{#1}}}
\newcommand{\second}[1]{\underline{\textcolor{black}{#1}}}

\title[Not All Retrievals are Useful: Cross-Attention for Input-Aware RAG in Time Series Forecasting]{Not All Retrievals are Useful: Cross-Attention for \\Input-Aware RAG in Time Series Forecasting}

\author{Seunghan Lee}
\affiliation{%
  \institution{LG AI Research}
  \city{Seoul}
  \country{South Korea}
}
\author{Jaehoon Lee}
\affiliation{%
  \institution{LG AI Research}
  \city{Seoul}
  \country{South Korea}
}
\author{Jun Seo}
\affiliation{%
  \institution{LG AI Research}
  \city{Seoul}
  \country{South Korea}
}
\author{Sungdong Yoo}
\affiliation{%
  \institution{LG AI Research}
  \city{Seoul}
  \country{South Korea}
}
\author{Minjae Kim}
\affiliation{%
  \institution{LG AI Research}
  \city{Seoul}
  \country{South Korea}
}
\author{Tae Yoon Lim}
\affiliation{%
  \institution{LG AI Research}
  \city{Seoul}
  \country{South Korea}
}
\author{Dongwan Kang}
\affiliation{%
  \institution{LG AI Research}
  \city{Seoul}
  \country{South Korea}
}
\author{Hwanil Choi}
\affiliation{%
  \institution{LG AI Research}
  \city{Seoul}
  \country{South Korea}
}
\author{SoonYoung Lee}
\affiliation{%
  \institution{LG AI Research}
  \city{Seoul}
  \country{South Korea}
}
\author{Wonbin Ahn}
\affiliation{%
  \institution{LG AI Research}
  \city{Seoul}
  \country{South Korea}
}

\renewcommand{\shortauthors}{Lee et al.}

\acmConference[KDD '26 MILETS Workshop]%
  {12th International Workshop on Mining and Learning from Time Series}%
  {August 9--13, 2026}{Jeju, Korea}

\begin{document}

%% Abstract MUST come BEFORE \maketitle in acmart (opposite of NeurIPS).
\begin{abstract}
Retrieval-augmented generation (RAG) enhances zero-shot time series (TS) forecasting
by leveraging external knowledge bases,
yet existing approaches overlook input-level relevance when fusing retrieved samples with the query.
We argue that \textit{not all retrievals are equally useful}, and irrelevant ones can degrade performance.
To this end, we propose \textbf{Cross-RAG}, a zero-shot RAG-based forecasting framework
that \textit{selectively} attends to query-relevant retrieved samples via query–retrieval \textit{cross-attention}.
By modeling input-level relevance between the query and retrieved samples, Cross-RAG jointly incorporates
three sources of information:
1) the query itself,
2) the retrieved samples, and
3) their relational interactions.
In particular, this input-aware design enables Cross-RAG to remain stable as the number of retrieved samples $k$ grows, whereas prior methods without cross-attention require careful $k$ tuning to avoid degradation from irrelevant retrievals.
Extensive experiments demonstrate that Cross-RAG consistently improves zero-shot forecasting performance across multiple TSFM backbones and various RAG methods, with additional analyses confirming
its effectiveness
across various retrieval scenarios.
Code is available at: \url{https://github.com/seunghan96/Cross-RAG}.

\end{abstract}

\maketitle

\input{icml_01_intro}
\input{icml_02_related_works}
\input{icml_03_methodology}

\input{icml_04_experiments_settings}
\input{icml_04_experiments_main}
\input{icml_04_experiments_ablation_modules}

\input{icml_05_analysis1.tex}
\input{icml_05_analysis2.tex}
\input{icml_06_conclusion}

\FloatBarrier   % flush pending floats from main body before References

%% Impact Statement is not required by the MILETS 2026 CFP — commented out.
% \section*{Impact Statement}
% This paper improves time series forecasting by incorporating retrieval-augmented generation into time series foundation models.
% The impact extends to domains where reliable forecasts are essential, including finance and healthcare, where improved predictions support informed decision-making and efficient resource allocation.
% We note that deploying automated zero-shot forecasting systems in high-stakes domains without sufficient validation may lead to overconfident predictions and consequential decision errors. Human oversight remains essential when applying such methods in safety-critical settings.
% We do not identify direct ethical concerns associated with this work.
% Overall, this study facilitates broader adoption of foundation models for complex time series analysis.

%% ACM bibliography style (provided by ACM-Reference-Format.bst)
\bibliographystyle{ACM-Reference-Format}
\bibliography{icml2026_conference}

%%%%%%%%%%%%%%%%%%%%%%%%%%%%%%%%%%%%%%%%%%%%%%%%%%%%%%%%%%%%
% \clearpage
\FloatBarrier
\appendix
\renewcommand{\thefigure}{\thesection.\arabic{figure}}
\renewcommand{\thetable}{\thesection.\arabic{table}}
\counterwithin{figure}{section}
\counterwithin{table}{section}
\input{icml_99_appendix}

\end{document}

%% file: math_commands.tex
\usepackage{amsmath,amsfonts,bm}

\def\eqref#1{equation~\ref{#1}}
\def\1{\bm{1}}

\DeclareMathAlphabet{\mathsfit}{\encodingdefault}{\sfdefault}{m}{sl}
\SetMathAlphabet{\mathsfit}{bold}{\encodingdefault}{\sfdefault}{bx}{n}

%% file: icml_01_intro.tex
% \vspace{-8pt}
\section{Introduction}
% \vspace{-8pt}
Time series (TS) forecasting is widely adopted across diverse domains, including finance \cite{rezaei2021stock} and traffic systems \cite{cirstea2022towards}.
Deep learning--based approaches, such as recurrent \cite{xu2025time}, convolutional \cite{luo2024moderntcn}, graph-based \cite{huang2023crossgnn}, and transformer-based models \cite{liu2023itransformer}, effectively capture temporal dependencies, while their training remains largely confined to domain-specific datasets.

\begin{figure}[t]
\vspace{15pt}
\centering
\begin{adjustbox}{max width=\linewidth}
\includegraphics[width=1.0\columnwidth]{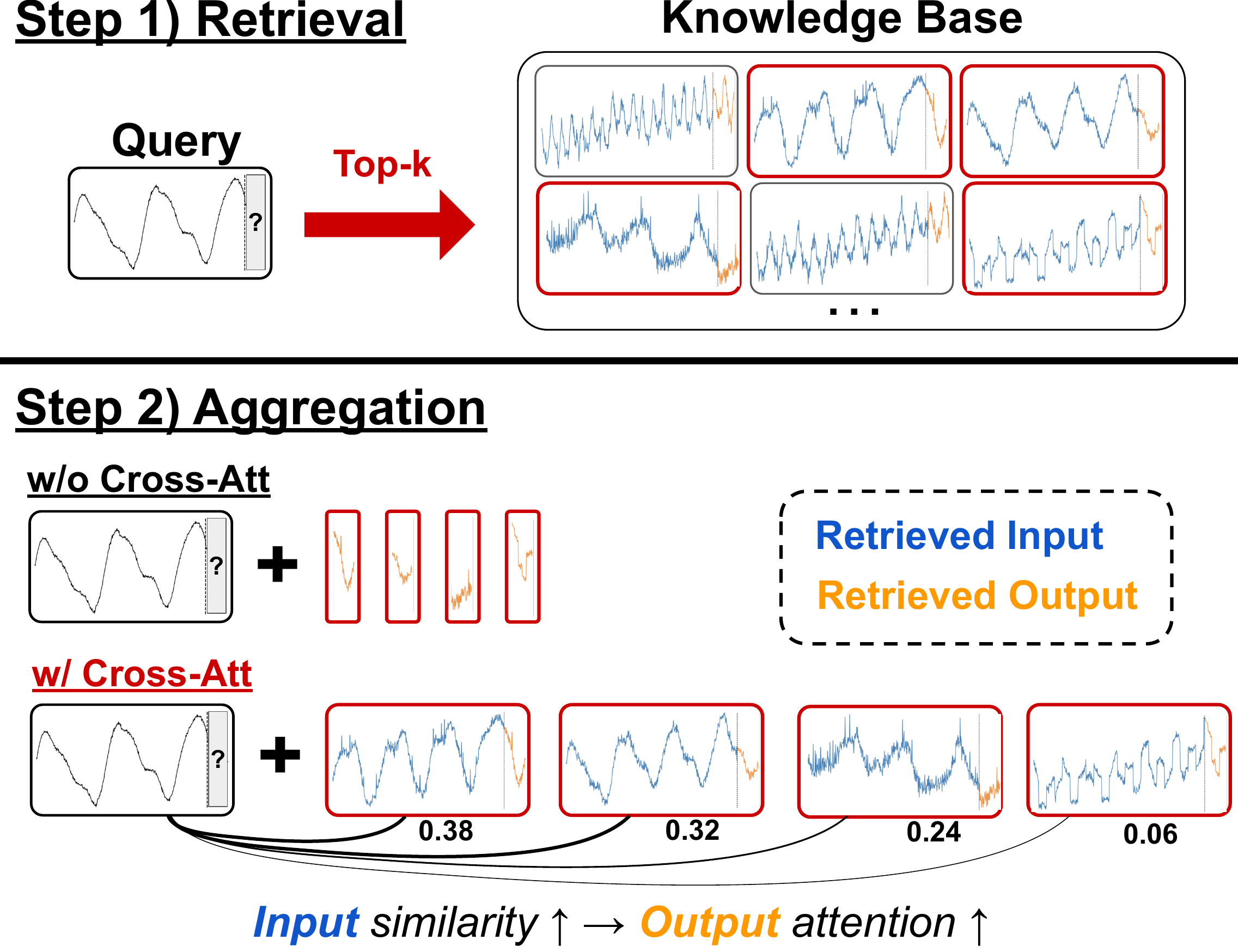}
\end{adjustbox}
\caption{\textbf{Cross-attention btw query \& retrieved inputs.}
Cross-RAG performs \textit{input-aware} fusion with \textit{cross-attention} to weight retrieved samples based on the input similarity.
}
\label{fig:motivation}
\vspace{-8pt}
\end{figure}

Recent studies investigate
time series foundation models (TSFMs) to enable more generalizable forecasting through large-scale pretraining \cite{jiangfstllm,tan2024language,ansari2024chronos, das2024decoder, lee2026dataset}.
Despite their expressive capacity, these models exhibit limited robustness in zero-shot settings due to distribution mismatch between the pretraining and unseen datasets, incur substantial adaptation cost, and lack mechanisms for dynamic domain conditioning and interpretability \cite{ning2025ts}.

Recently, retrieval-augmented generation (RAG) enhances context awareness across diverse tasks by retrieving relevant contextual evidence and incorporating it into model inputs.
Several studies extend this paradigm to TS forecasting \cite{ning2025ts, han2025retrieval, yang2025timerag},
seeking to leverage historical patterns
through retrieval mechanisms. However, existing 
%RAG-based 
approaches for TS forecasting either rely on dataset-specific adaptation rather than zero-shot inference or fail to fully exploit the input structure of retrieved samples. %during forecasting.

\begin{figure*}[t]
\centering
\begin{minipage}[b]{0.28\textwidth}
  \centering
  \begin{adjustbox}{max width=\linewidth}
    \includegraphics[width=\linewidth]{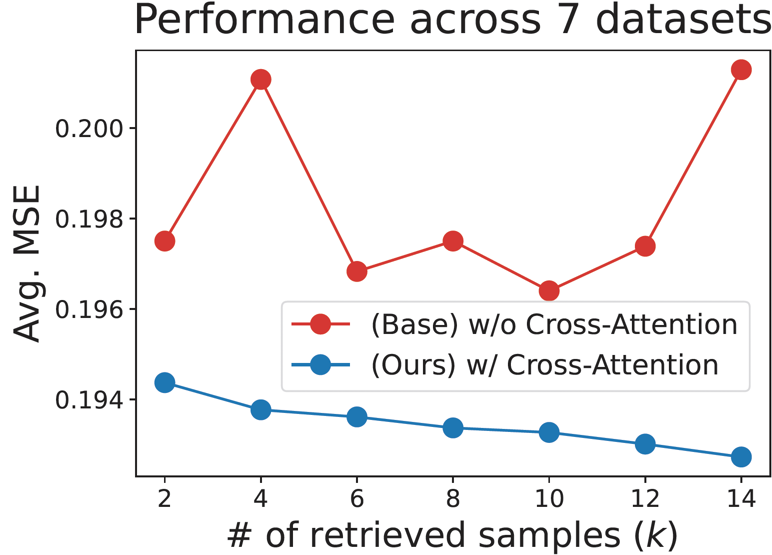}
  \end{adjustbox}
  \captionof{figure}{\textbf{Performance by $k$.} Selective attention enables Cross-RAG to benefit from larger $k$, even when irrelevant samples are included.}
  \label{fig:motivation_k}
\end{minipage}
\hfill
\begin{minipage}[b]{0.70\textwidth}
%   \vspace{10pt}
  \centering
  \begin{threeparttable}
  \begin{adjustbox}{max width=\linewidth}
  \begin{NiceTabular}{c|l|c|c|c}
  \toprule
  \multicolumn{2}{c}{Method}
  & \textbf{1. How to retrieve?}
  & \textbf{2. What to retrieve?}
  & \textbf{3. How to fuse?} \\
  \midrule
%   \cellcolor{LightGray1} 
  \multirow{3}{*}{Full-shot}
  & \cellcolor{LightGray1} \textcolor{LightTextGray}{RAFT (ICML 2025)}
  & \cellcolor{LightGray1} \textcolor{LightTextGray}{$Q$ vs. $R_x$}
  & \cellcolor{LightGray1} \textcolor{LightTextGray}{$R_y$}
  & \cellcolor{LightGray1} \textcolor{LightTextGray}{Concatenate} \\
%   \cellcolor{LightGray1}
  & \cellcolor{LightGray1} \textcolor{LightTextGray}{RATD (NeurIPS 2024)}
  & \cellcolor{LightGray1} \textcolor{LightTextGray}{$Q$ vs. $h(R_x)$}
  & \cellcolor{LightGray1} \textcolor{LightTextGray}{$(R_x, R_y)$}
  & \cellcolor{LightGray1} \textcolor{LightTextGray}{Cross-Attention$^{(1)}$} \\
%   \cellcolor{LightGray1}
  & \cellcolor{LightGray1} \textcolor{LightTextGray}{TimeRAG (ICASSP 2025)}
  & \cellcolor{LightGray1} \textcolor{LightTextGray}{$Q$ vs. $R_x$}
  & \cellcolor{LightGray1} \textcolor{LightTextGray}{$R_x$}
  & \cellcolor{LightGray1} \textcolor{LightTextGray}{\xmark$^{(2)}$} \\
  \midrule
  \multirow{3}{*}{\textbf{Zero-shot}}
  & RAF (arXiv 2025) & $Q$ vs. $h(R_x)$ & $(R_x, R_y)$ & \xmark$^{(3)}$ \\
  & TS-RAG (NeurIPS 2025) & $Q$ vs. $h(R_x)$ & $R_y$ & Self-Attention \\
  & \cellcolor{LightYellow} Cross-RAG (Ours) & \cellcolor{LightYellow} $Q$ vs. $R_x$ & \cellcolor{LightYellow} $(R_x, R_y)$ & \cellcolor{LightYellow} Cross-Attention \\
  \bottomrule
  \end{NiceTabular}
  \end{adjustbox}
  \begin{tablenotes}
  \footnotesize
  \item[(1)] The cross-attention mechanism in \textbf{RATD} is fundamentally different from ours, as it attends over retrieved samples using both $(R_x, R_y)$ as keys and dataset-level auxiliary statistics as values, rather than modeling query--retrieval interactions.
  \item[(2)] \textbf{TimeRAG} converts retrieved input sequences into text prompts and feeds them into an LLM backbone.
  \item[(3)] \textbf{RAF} forms a single sequence by prepending retrieved samples to the query without modeling interaction between them.
  \end{tablenotes}
  \end{threeparttable}
  \captionof{table}{
%   \textbf{Comparison of RAG in TS.}
  \textbf{RAG in TS forecasting.}
  (1) \emph{How to retrieve?} Defines the similarity space for retrieval,
  %(data vs. latent), 
  depending on 
%   usage of 
% the
  retrieval encoder $h(\cdot)$.
  (2) \emph{What to retrieve?} Specifies the retrieved content $R$.
  (3) \emph{How to fuse?} Describes how the retrieved info is integrated with query $Q$.}
  \label{tbl:compare_RAG}
\end{minipage}
\end{figure*}
\begin{figure*}[t]
    \vspace{10pt}
    \centering
    \begin{subfigure}[t]{0.1885\textwidth}
        \centering
        \begin{adjustbox}{max width=\linewidth}
        \includegraphics[width=\textwidth]{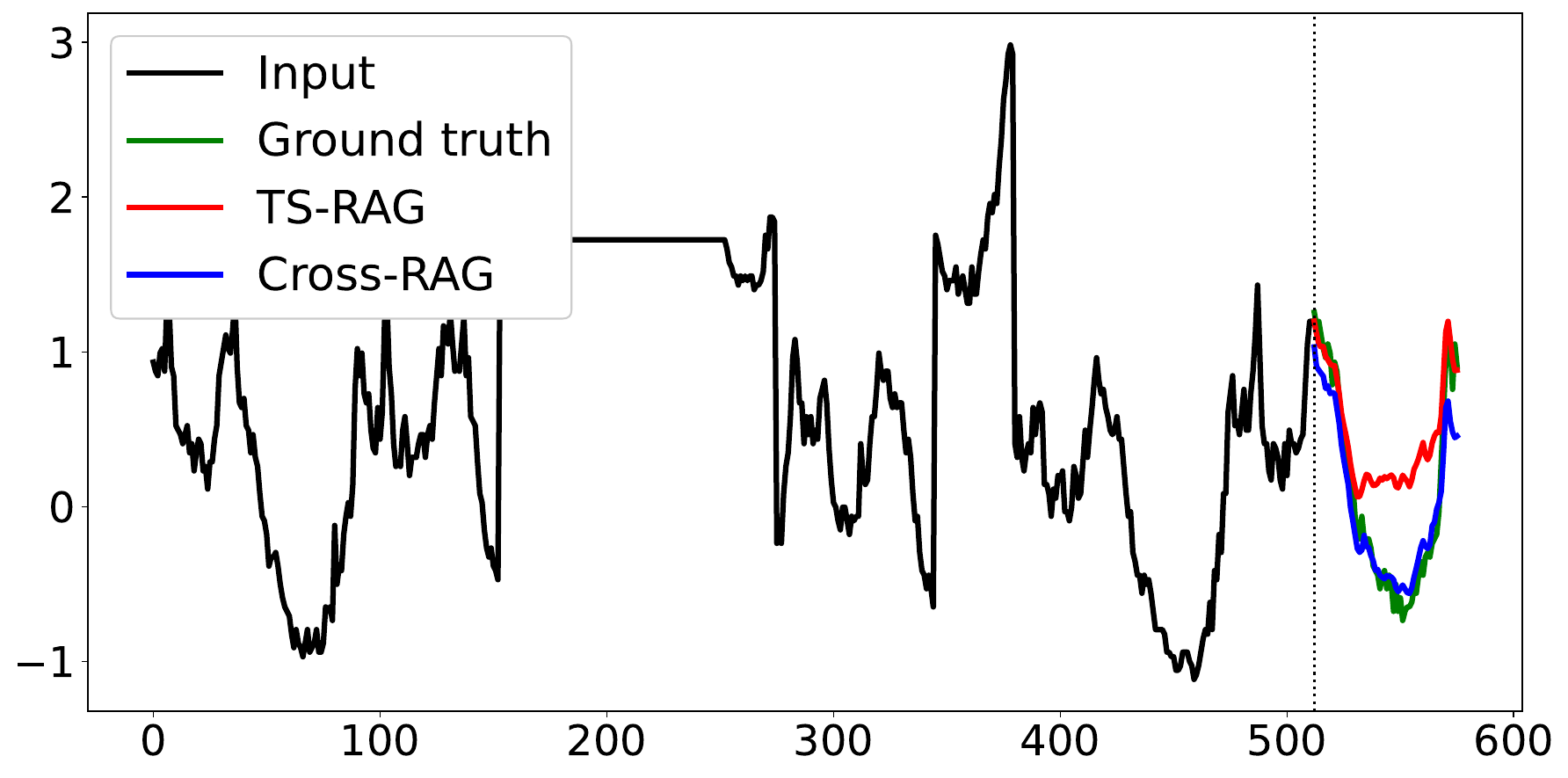}
        \end{adjustbox}
        \caption{ETTm1}
    \end{subfigure}
    \begin{subfigure}[t]{0.1885\textwidth}
        \centering
        \begin{adjustbox}{max width=\linewidth}
        \includegraphics[width=\textwidth]{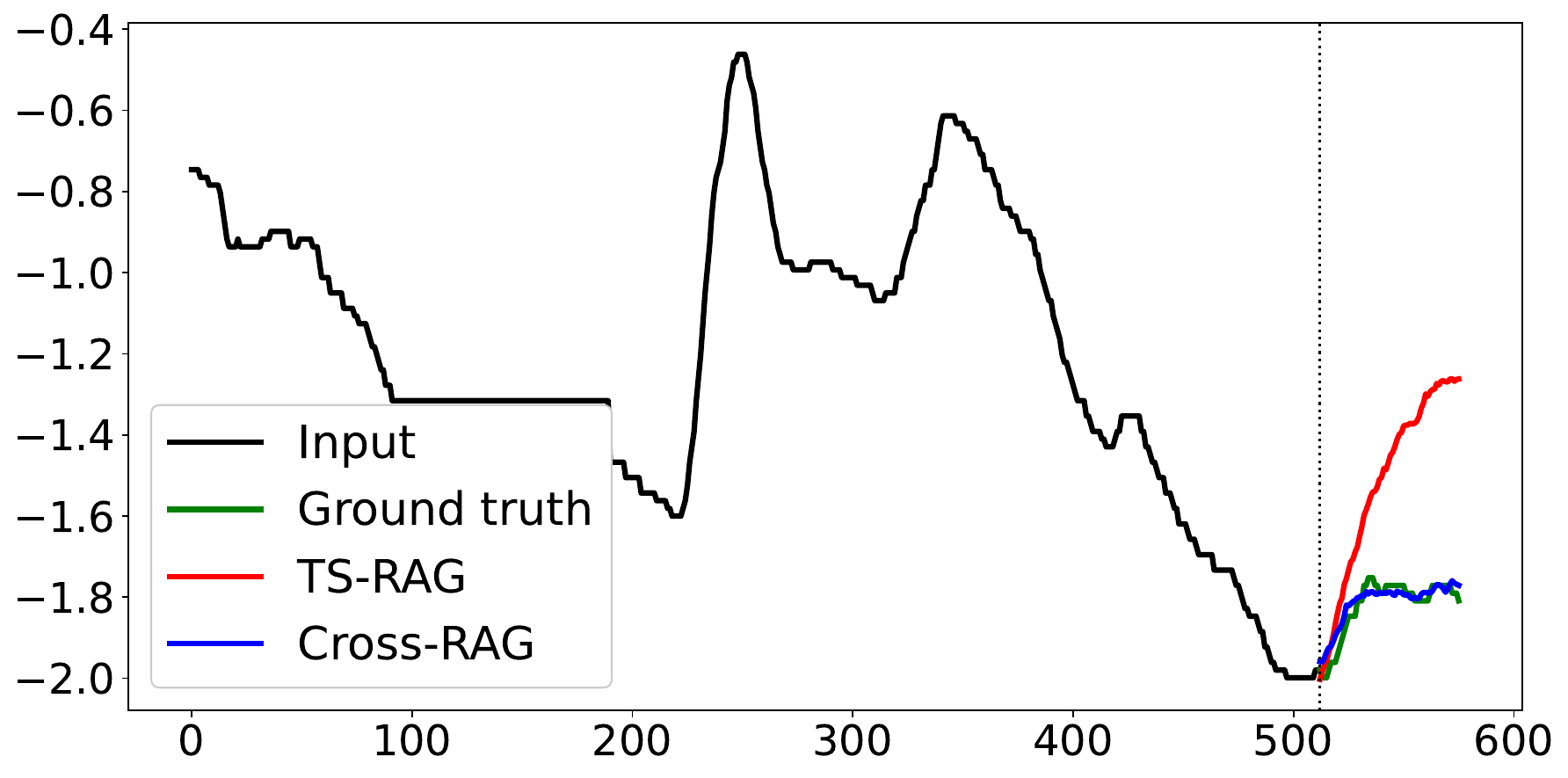}
        \end{adjustbox}
        \caption{ETTm2}
    \end{subfigure}
    \begin{subfigure}[t]{0.1885\textwidth}
        \centering
        \begin{adjustbox}{max width=\linewidth}
        \includegraphics[width=\textwidth]{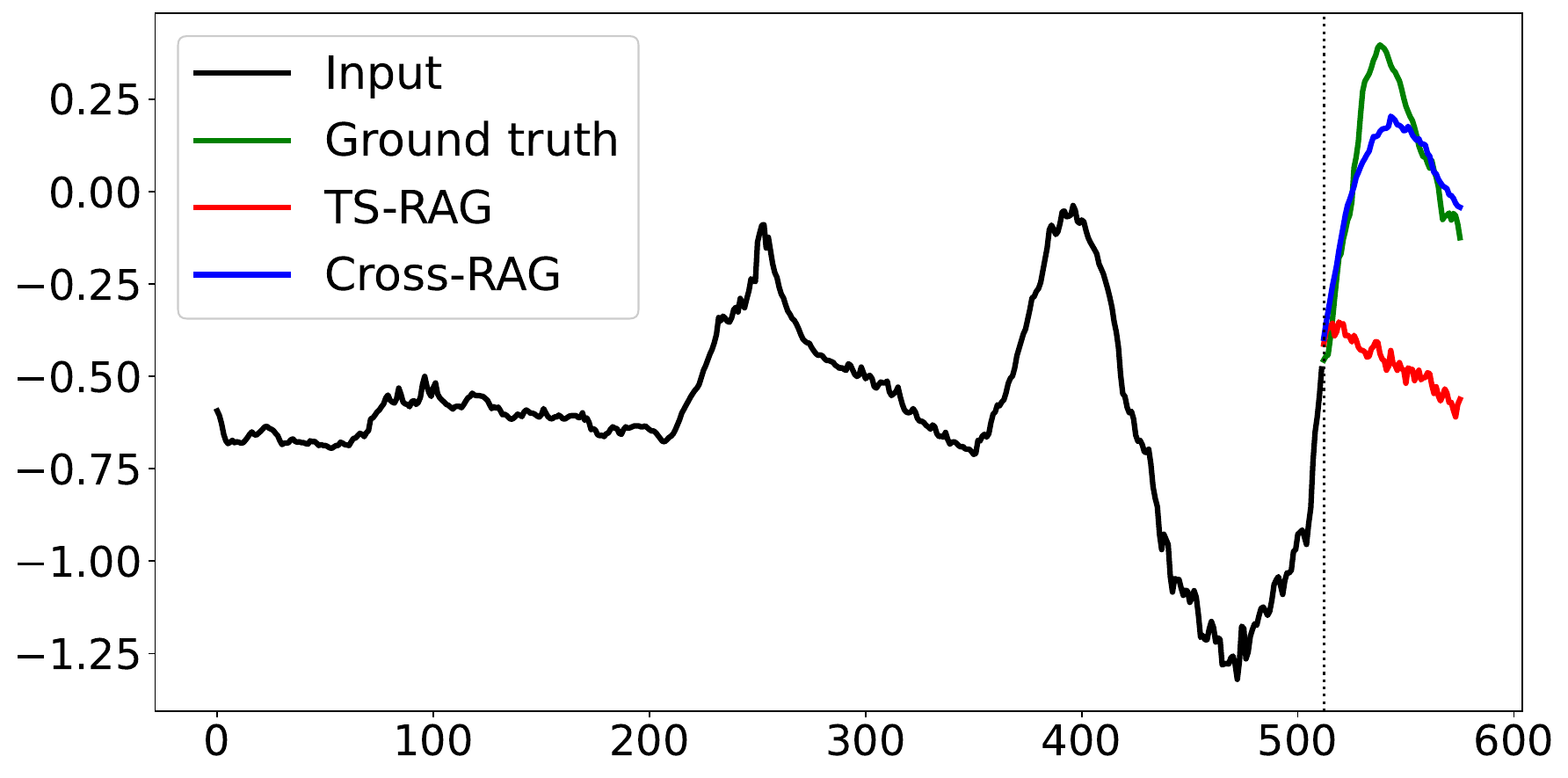}
        \end{adjustbox}
        \caption{Weather}
    \end{subfigure}
    \begin{subfigure}[t]{0.1885\textwidth}
        \centering
        \begin{adjustbox}{max width=\linewidth}
        \includegraphics[width=\textwidth]{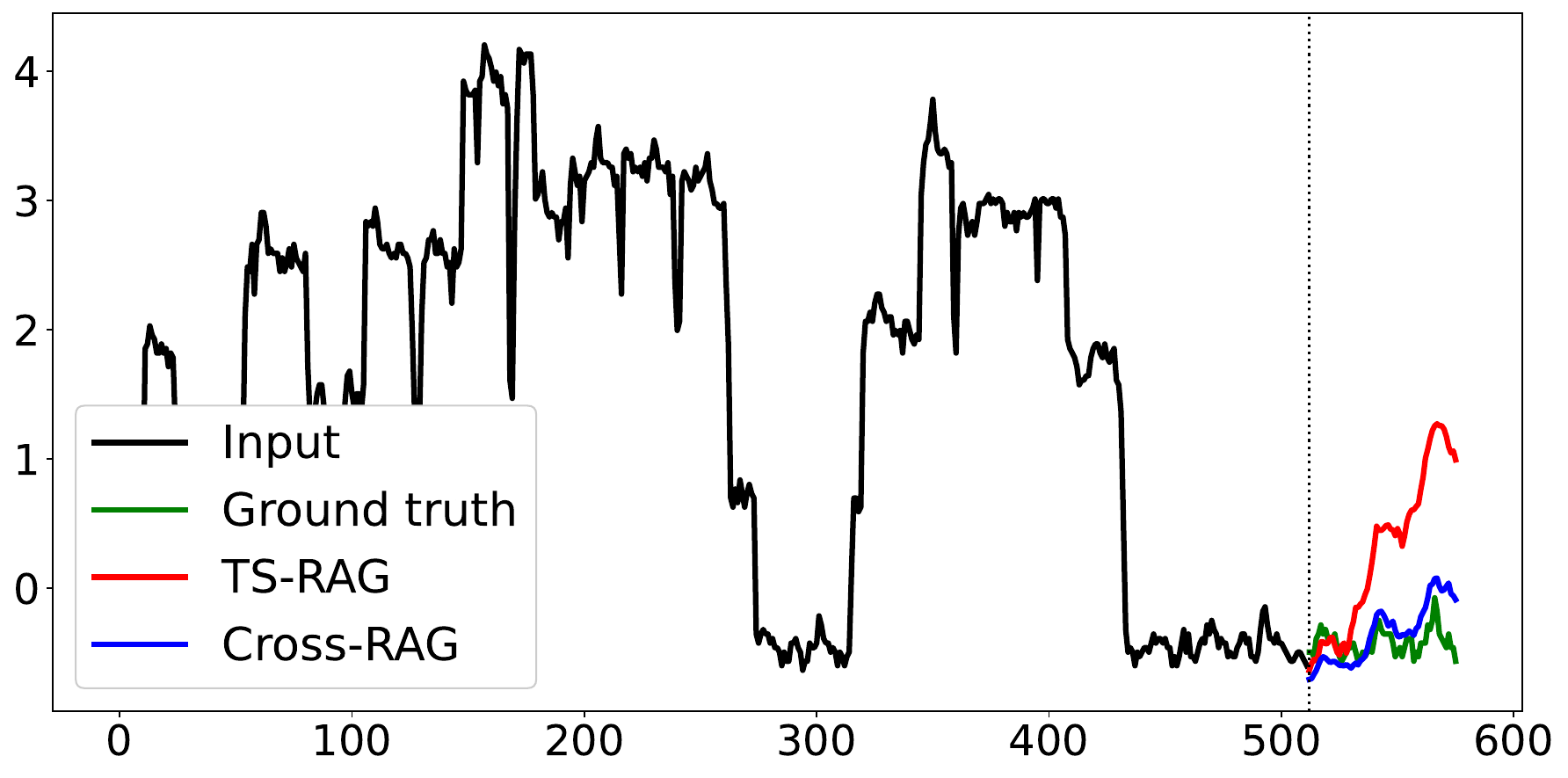}
        \end{adjustbox}
        \caption{Electricity}
    \end{subfigure}
    \begin{subfigure}[t]{0.1885\textwidth}
        \centering
        \begin{adjustbox}{max width=\linewidth}
        \includegraphics[width=\textwidth]{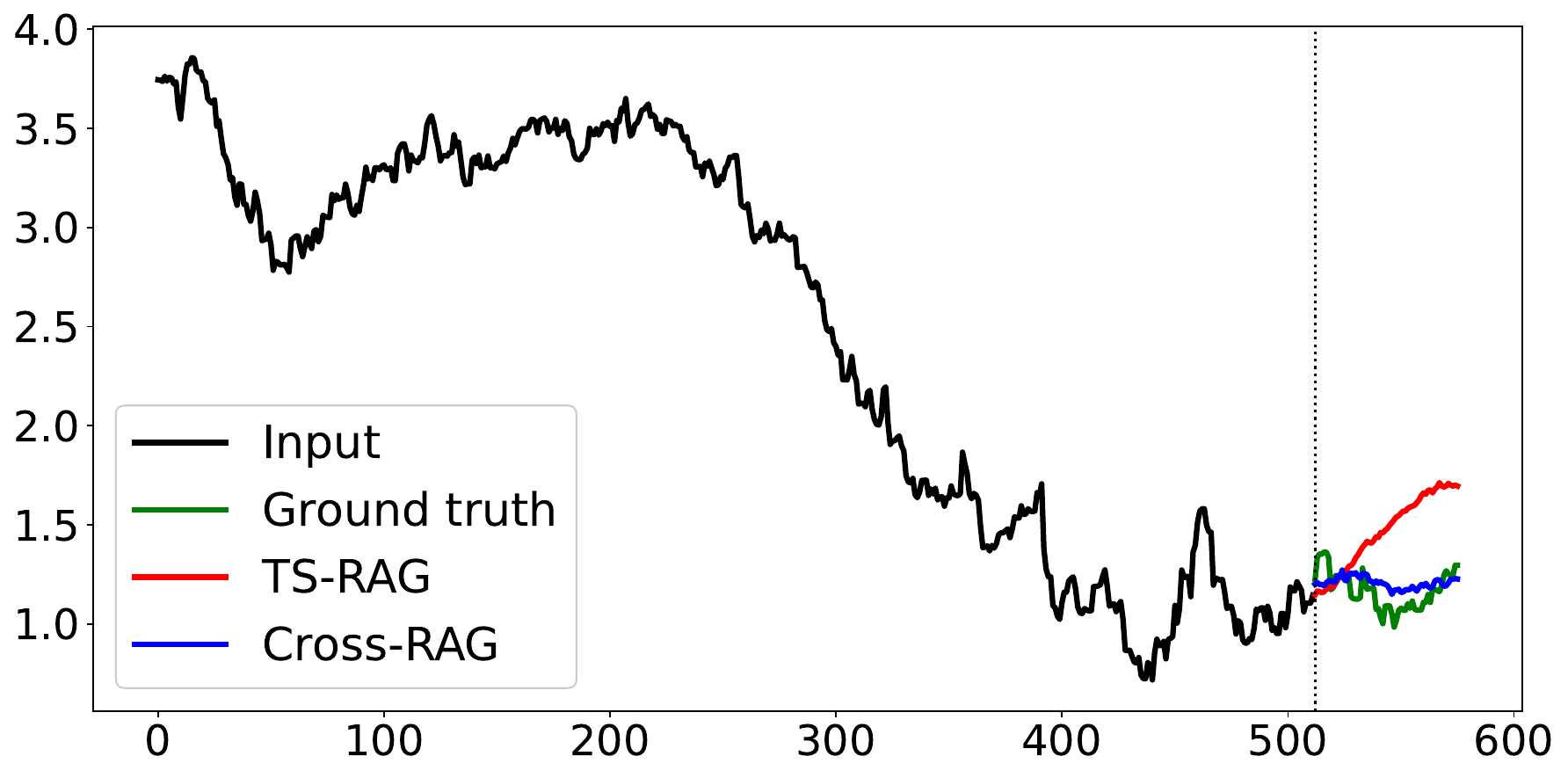}
        \end{adjustbox}
        \caption{Exchange}
    \end{subfigure}
    \caption{
    \textbf{Visualization of zero-shot TS forecasting results.} The figure shows the forecasting results across five datasets for models with cross-attention (Ours) and without cross-attention (TS-RAG), demonstrating that Cross-RAG follows the ground truth more closely than TS-RAG across diverse domains.
    }
    \label{fig:viz_ts}
% \vspace{-8pt}    
\end{figure*}

To this end, we propose \textbf{Cross-RAG}, a zero-shot retrieval-augmented forecasting framework
which
models input-level relevance between the query and retrieved samples by attending to their inputs via (query--retrieval) \textit{cross-attention},
as illustrated in Figure~\ref{fig:motivation}. Specifically, the prediction integrates three sources of information: 1) the query itself, 2) the retrieved samples themselves, and 3) the relational information between the query and retrieved samples.

As shown in Figure~\ref{fig:motivation_k},
Cross-RAG consistently improves performance (average MSE across seven datasets) as the number of retrieved samples ($k$) increases by selectively leveraging relevant samples
even when irrelevant ones may be included,
whereas without cross-attention~\cite{ning2025ts}, performance does not improve monotonically with increasing $k$, requiring careful selection of $k$.
Figure~\ref{fig:viz_ts} visualizes the zero-shot forecasting results, 
showing that incorporating cross-attention more closely follows the ground truth than the variant without cross-attention~\cite{ning2025ts} across diverse datasets.

The main contributions are as follows:
%showing that employing cross-attention (Cross-RAG) follows the ground truth more closely than without cross-attention (TS-RAG~\cite{ning2025ts}) across diverse datasets.
% \setlist[itemize]{leftmargin=0.3cm,itemsep=-1.5pt,topsep=-1.5pt, partopsep=0pt}
\setlist[itemize]{leftmargin=0.3cm}
\begin{itemize}
    \item We propose \textbf{Cross-RAG}, a plug-in method that enhances retrieval-augmented 
    %TS 
    forecasting by \textit{selectively} attending to retrieved samples according to their relevance to the query,
    using \textit{cross-attention} between the query and the inputs of the retrieved samples.
    \item To the best of our knowledge, Cross-RAG is the first zero-shot RAG approach to model relationships between the query and retrieved inputs, as shown in Table~\ref{tbl:compare_RAG}.
    \item
    We validate Cross-RAG through extensive experiments, achieving state-of-the-art (SoTA) performance across diverse datasets and backbones.
    The proposed method supports zero-shot forecasting, enabling efficient integration without requiring additional retraining of target datasets.
    \item
    We conduct in-depth analyses under various retrieval conditions, demonstrating its ability to selectively attend to relevant samples while remaining robust to 
%     the 
    %inclusion of 
    irrelevant 
    %retrieved 
    samples.
\end{itemize}

%% file: icml_02_related_works.tex
% \vspace{-3pt}
\section{Related Works}
% \vspace{-5pt}
\textbf{Time series foundation models (TSFMs).} 
%Time series foundation models 
TSFMs
have emerged as a scalable alternative to LLM-based forecasters, which incur high computational costs and domain adaptation challenges 
%\cite{zhou2023one, xue2023promptcast, chang2023llm4ts, jin2023time, pan2024s, sun2023test,niu2025langtime}. 
\cite{zhou2023one, xue2023promptcast, chang2023llm4ts, sun2023test,niu2025langtime}. 
Early works introduce large-scale pretraining for TS forecasting \cite{LagLlama,TimeGPT1}, followed by Transformer-based extensions \cite{liutimer,liu2025sundial}. Recent models commonly adopt patch-based tokenization to embed continuous TS, where TimesFM \cite{das2024decoder} and MOMENT \cite{goswami2024moment} model patch-level representations to enable forecasting across diverse datasets, while TTM \cite{ekambaram2024tiny} improves efficiency through compact architectures with reduced parameterization. Time-MoE \cite{shi2024time} further enhances scalability by adopting a sparse mixture-of-experts design.
%that activates only a subset of experts during inference. 
Beyond deterministic forecasting, Moirai \cite{woo2024unified} learns mixtures of predictive distributions to capture uncertainty. Chronos \cite{ansari2024chronos} formulates forecasting as a language modeling task by discretizing scaled TS into categorical tokens, 
%and modeling them with cross-entropy loss, 
and Chronos-Bolt replaces discrete tokenization with patch-based inputs.
%and produces multi-step quantile forecasts.
%from decoder representations.
%to improve inference efficiency and accuracy. 
Despite strong generalization, these models rely on internal representations and lack mechanisms to dynamically incorporate external knowledge, which limits interpretability and zero-shot adaptability \cite{ning2025ts}.

\begin{figure*}[t]
\centering
\begin{adjustbox}{max width=0.95\linewidth}
\includegraphics[width=\textwidth]{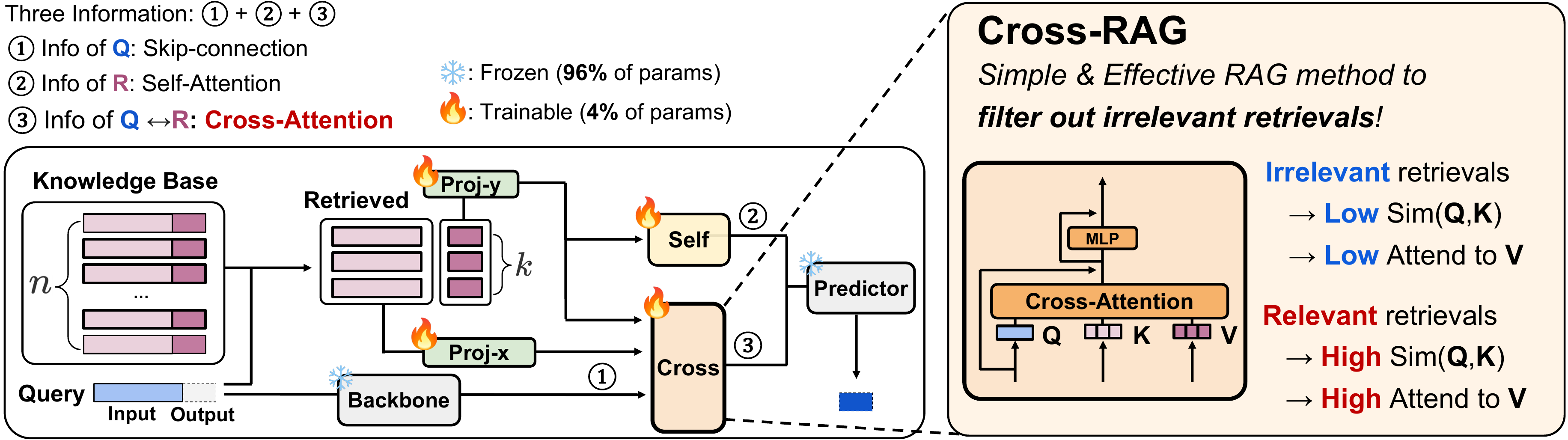}
\end{adjustbox}
\caption{
% \textbf{Overall framework of Cross-RAG.}
\textbf{Framework of Cross-RAG.}
Cross-RAG models \textit{query–retrieval relevance} through \textit{cross-attention}, which selectively aggregates retrieved outputs conditioned on their relevance to the query.
Additionally, retrieval self-attention is employed to provide (query-independent) contextual information among retrieved samples.
The backbone and predictor are frozen.
%during pretraining.
}
\label{fig:main}
\end{figure*}

\textbf{RAG in TS.} 
RAG \cite{lewis2020retrieval} provides a principled way to incorporate external knowledge and improve robustness, and recent studies extend this paradigm to TS forecasting.
ReTime \cite{jing2022retrieval} introduces relational retrieval with content synthesis, 
RATD \cite{liu2024retrieval} leverages retrieved historical series to guide diffusion-based denoising, 
and RAFT \cite{han2025retrieval} combines retrieval with multi-resolution forecasting. 
TimeRAG \cite{yang2025timerag} incorporates retrieved sequences through a frozen LLM backbone with a trainable reprogramming layer to align modalities. 
However, these approaches are limited in that they \textit{require fine-tuning on the target set} for adaptation to unseen datasets.
% Zero-shot refers to settings where the target task dataset is not used during training, and models rely only on pretraining over large-scale task-independent time series datasets.
% Recently, zero-shot\footnote{In the time series RAG literature, zero-shot refers to settings where no data from the target task dataset is used, and models are pretrained solely on large-scale task-independent time series corpora (e.g., the Chronos dataset).} RAG methods have also been introduced for TS forecasting.
%leverage the pretraining dataset of
%Chronos

Recently, zero-shot\footnote{
Zero-shot in RAG refers to settings where no data from the \textit{target task dataset} is used for parameter updates, 
and models are pretrained 
%solely 
on 
%large-scale, 
\textit{task-independent TS corpora}.} RAG methods have also been introduced for TS forecasting.
RAF \cite{tire2024retrieval} constructs an augmented input by 
%directly 
concatenating retrieved sequences with the query sequence in the raw data space.
TS-RAG \cite{ning2025ts} retrieves 
%semantically relevant 
TS segments using a pretrained encoder
%and integrates them into the forecasting process by conducting 
and conducts self-attention with concatenated query and retrieved outputs.
However, these approaches \textit{do not explicitly account for the relationship between the query and the retrieved inputs}
when aggregating retrieved samples.
% Comparison of RAG methods in TS is shown in Table~\ref{tbl:compare_RAG}.

%RAF: retrieve input하지만, 이를 query와 함께 사용하지 않고, 단순히 별도의 augmentation sample로써 활용하여 query input 앞에 붙여서 긴 long-sequence 로써 만듬.
%RAFT \cite{han2025retrieval} 또한 input 자체를 입력으로 받지 않고, 단순히 입력와의 유사도에 가중합을 한 output만을 retrieve한다. 
%RAF constructs augmented inputs by concatenating retrieved samples as a long sequence, without explicitly modeling interactions between the query and retrieved contexts. RAFT retrieves similarity-weighted outputs rather than integrating the input query itself, which limits fine-grained query-dependent adaptation.

%% file: icml_03_methodology.tex
\begin{figure*}[t]
    \centering
    \begin{subfigure}[t]{0.47\textwidth}
        \centering
        \begin{adjustbox}{max width=\linewidth}
        \includegraphics[width=\textwidth]{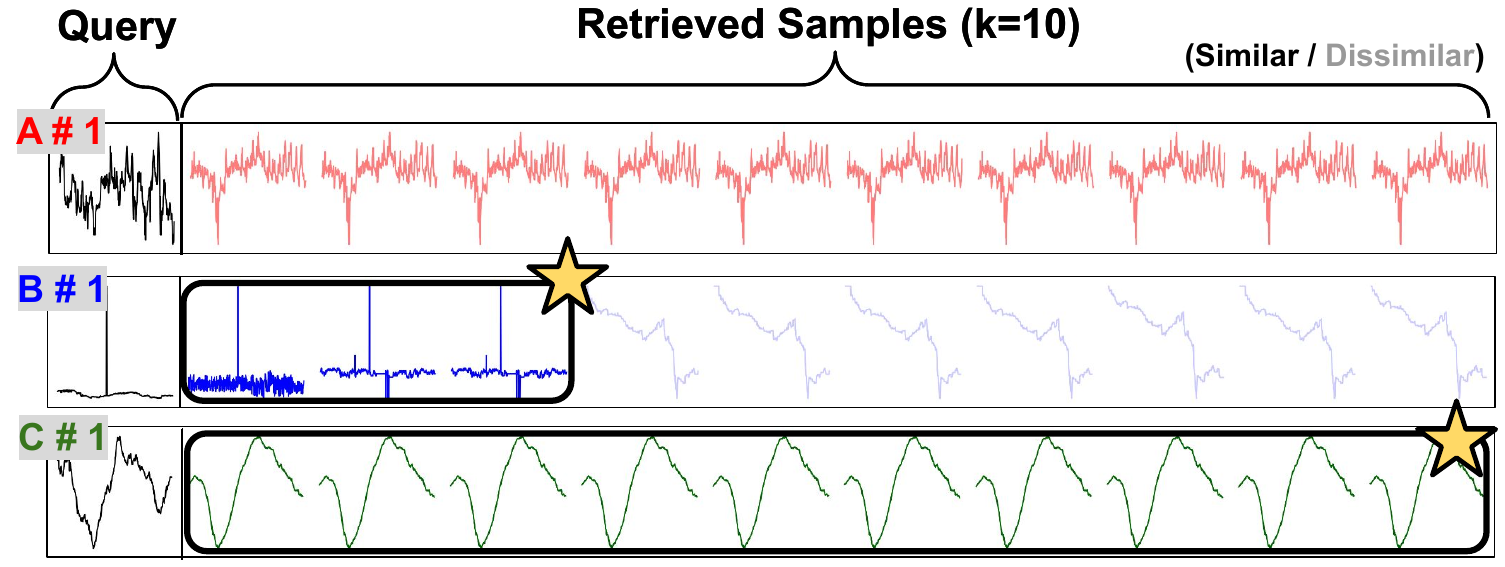}
        \end{adjustbox}
        \caption{Optimal $k$ varies \textbf{across} datasets.}
        \label{fig:optimal_k_differs2}
    \end{subfigure}
    \hfill
    \begin{subfigure}[t]{0.47\textwidth}
        \centering
        \begin{adjustbox}{max width=\linewidth}
        \includegraphics[width=\textwidth]{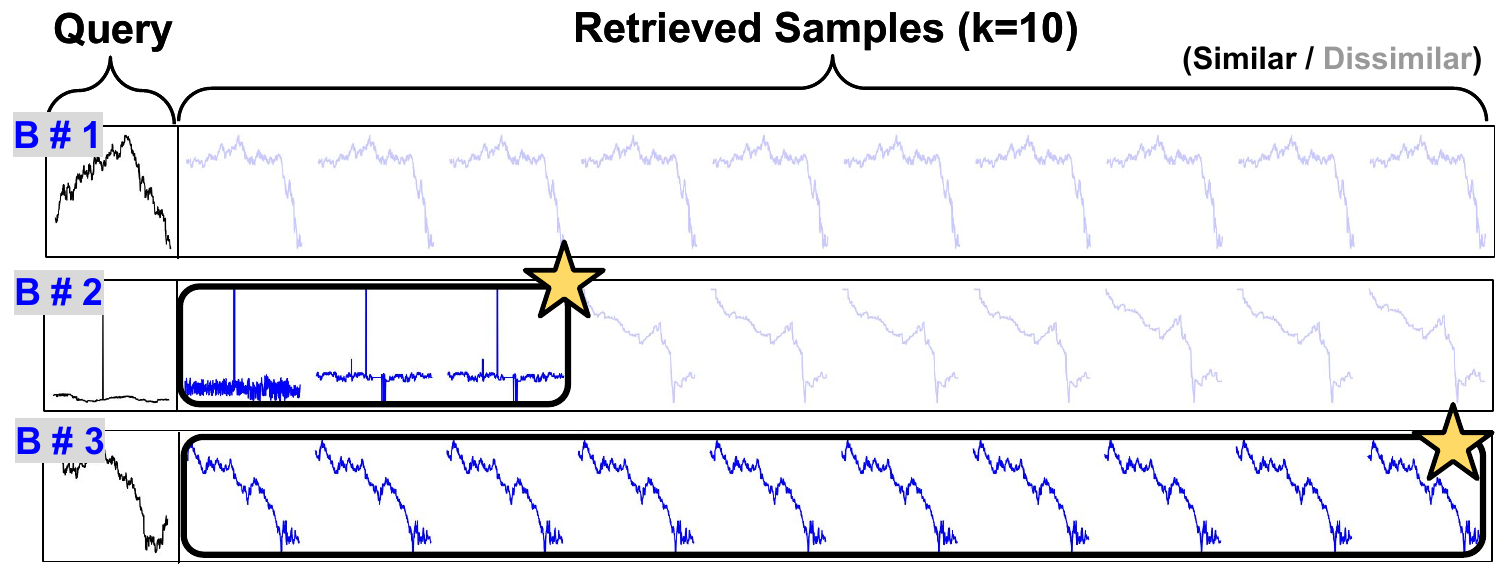}
        \end{adjustbox}
        \caption{Optimal $k$ varies \textbf{within} a dataset.}
        \label{fig:optimal_k_differs1}
    \end{subfigure}
    \vspace{-7pt}
    \caption{
    \textbf{Necessity of selective attention.}
    The figure shows that the optimal $k$ varies both \textit{across} and \textit{within} datasets (\textcolor{red}{[A] ETTh1}, \textcolor{darkblue}{[B] Exchange}, \textcolor{darkgreen}{[C] Weather}), highlighting the necessity of selectively attending to
    samples among the retrieved samples.
    }
    \label{fig:optimal_k_differs}
    \vspace{-3pt}
\end{figure*}

\section{Methodology}
\textbf{TS forecasting with RAG.} We consider the task of TS forecasting with retrieval augmentation.
Let $\mathcal{D} = \{(\boldsymbol{x}_j, \boldsymbol{y}_j)\}_{j=1}^{n}$ denote a retrieval knowledge base,
where $\boldsymbol{x}_j \in \mathbb{R}^{T}$ is a input window of length $T$
and $\boldsymbol{y}_j \in \mathbb{R}^{L}$ is the corresponding forecasting horizon of length $L$.
Given a query input $\boldsymbol{x}_i$, our goal is to predict $\hat{\boldsymbol{y}}_i$
by leveraging both the query input and a set of retrieved samples from $\mathcal{D}$.

In this section, we
organize our retrieval-augmented forecasting framework into three key aspects:
\begin{itemize}
    \item \mbox{Section~\ref{sec:how_to_retrieve}} (\textbf{\textit{How to Retrieve}}) covers the choice of 
    %similarity 
    space and metric for selecting relevant samples.
    \item \mbox{Section~\ref{sec:what_to_retrieve}} (\textbf{\textit{What to Retrieve}}) specifies which parts of the retrieved samples ($X$ or $Y$)
    %—inputs, outputs, or both—
    are used.
    \item \mbox{Section~\ref{sec:how_to_fuse}} (\textbf{\emph{How to Fuse}})
describes how query and retrieved information are aggregated to produce a final representation.
% for forecasting.
\end{itemize}
The overall framework of our method is shown in Figure~\ref{fig:main}.

\begin{table}[t]
\vspace{-3pt}
\caption{
\textbf{\emph{1) How to retrieve?}}
Robustness to similarity space and retrieval metric 
% (Average MSE across seven datasets).
(Avg. MSE across seven datasets).
}
\centering
\begin{adjustbox}{max width=\linewidth}
\begin{NiceTabular}{c|c|c|c}
\toprule
% \multicolumn{2}{c}{\multirow{2}{*}{\shortstack{Retrieval space \\ \& Metric}}} & \multirow{2}{*}{\shortstack{Retrieval\\encoder}} & \multirow{2}{*}{\shortstack{Avg.\\MSE}}\\
\multicolumn{2}{c}{Retrieval space \& Metric} & Retrieval encoder & Avg. MSE\\
\midrule
\multirow{3}{*}{Data}
& Cosine & \multirow{3}{*}{\xmark}  & 0.191 \\
& Euclidean & & 0.193 \\
& Correlation & & 0.194 \\
\midrule
\multicolumn{2}{c}{Latent} & \cmark & 0.193\\
\bottomrule
\end{NiceTabular}
\end{adjustbox}
\label{tbl:retrieval_space}
\vspace{-8pt}
\end{table}
\subsection{How to Retrieve} \label{sec:how_to_retrieve}
A key design choice in RAG for TS forecasting is the retrieval space,
where we analyze retrieval performed in the \textit{data space} and the \textit{latent space}.
Table~\ref{tbl:retrieval_space} reports the average MSE across seven datasets using four different similarity metrics, showing that our method remains robust across retrieval spaces and metrics.
This result contrasts with prior studies~\cite{ning2025ts, liu2024retrieval}, which report inferior performance with data-space retrieval.
In contrast, our method achieves comparable performance in both spaces, while data-space retrieval avoids the need for an additional retrieval encoder and thus offers a more efficient alternative, which is further discussed in Section~\ref{sec:efficiency}.
% Based on this observation, we adopt data-space retrieval in our framework.
% Formally,
% given a cosine similarity on the input space,
% we retrieve the top-$k$ most similar samples to $\boldsymbol{x}_i$ from the dataset $\mathcal{D}$, denoted as $\mathcal{R}_i$.
Based on this observation, we adopt data-space retrieval by selecting the top-$k$ samples from $\mathcal{D}$ that are most similar to $\boldsymbol{x}_i$ under cosine similarity in the input space, denoted as $\mathcal{R}_i$.

\subsection{What to Retrieve} \label{sec:what_to_retrieve}
Selecting a fixed 
%top-$k$ 
$k$
%retrieved 
samples can be problematic, as 
%the 
optimal $k$ depends on how many samples in the knowledge base are similar to the query, and a fixed $k$ may include \textit{irrelevant ones} when this number varies.
Figure~\ref{fig:viz_ts} illustrates this issue, where Cross-RAG follows the ground truth more closely than the previous method \cite{ning2025ts}.

As shown in Figure~\ref{fig:optimal_k_differs2}, the optimal $k$ differs \textit{across} datasets.
Moreover, Figure~\ref{fig:optimal_k_differs1} shows that even \textit{within} the same dataset, the number of samples similar to a given query varies across queries,
highlighting the necessity to \textit{selectively attend to relevant samples} among the retrieved ones.

To address this issue, we fuse information between the query and the retrieved samples
by explicitly modeling the relationship between the query and the retrieved inputs.
Each retrieved element in $\mathcal{R}_i$ is represented as an input–output pair $(\boldsymbol{x}_j, \boldsymbol{y}_j)$, which are mapped into $d$-dimensional latent spaces via MLP-based projectors.
\begin{comment}
Formally, for each retrieved pair $(\boldsymbol{x}_j, \boldsymbol{y}_j) \in \mathcal{R}_i$,
we compute
\begin{equation}
\boldsymbol{r}_{x,j} = f_x(\boldsymbol{x}_j), \qquad
\boldsymbol{r}_{y,j} = f_y(\boldsymbol{y}_j),
\end{equation}
where $f_x(\cdot)$ and $f_y(\cdot)$
denote the input and output projectors, respectively.
\end{comment}
Formally,
we compute
\begin{equation}
\boldsymbol{r}_{x,j} = f_x(\boldsymbol{x}_j), \qquad
\boldsymbol{r}_{y,j} = f_y(\boldsymbol{y}_j)
\end{equation}
for each retrieved pair $(\boldsymbol{x}_j, \boldsymbol{y}_j) \in \mathcal{R}_i$,
where $f_x(\cdot)$ and $f_y(\cdot)$
denote the input and output projectors.
%, respectively.

\subsection{How to Fuse} \label{sec:how_to_fuse}
Given a query $\boldsymbol{x}$ and its retrieved set
$\mathcal{R} = \{(\boldsymbol{r}_{x,j}, \boldsymbol{r}_{y,j})\}_{j=1}^{k}$,
we integrate query and retrieved information by explicitly modeling
their input-level relevance.
%\footnote{For simplicity, we omit the query index in this section.}.
For simplicity, we omit the query index in this section.
Let $\boldsymbol{h} \in \mathbb{R}^{d}$ denote the query representation
produced by 
%an arbitrary 
TSFM backbone, and let
$\boldsymbol{R}_{x} \in \mathbb{R}^{k \times d}$ and
$\boldsymbol{R}_{y} \in \mathbb{R}^{k \times d}$
denote the stacked representations of retrieved inputs and outputs.

\textbf{(Query--Retrieval) Cross-Attention.}
The proposed method applies cross-attention
to selectively incorporate retrieval information relevant to the query,
using the query representation as the \textit{query} and the retrieved inputs and outputs as the \textit{keys} and \textit{values} of attention-mechanism, respectively:
\begin{equation}
\boldsymbol{c}
= \mathrm{Attn}\!\left(
\boldsymbol{h},\;
\boldsymbol{R}_{x},\;
\boldsymbol{R}_{y}
\right) + \boldsymbol{h}.
\end{equation}
Here, $\mathrm{Attn}(\cdot)$ denotes a multi-head attention block, and the residual connection preserves the information of the query itself.
The attended representation is further processed by a feed-forward network
with a residual connection:
\begin{equation}
\tilde{\boldsymbol{c}}
= \mathrm{FFN}_{\mathrm{cross}}(\boldsymbol{c}) + \boldsymbol{c}.
\end{equation}
This operation selectively extracts information from the retrieved outputs by weighting them according to the relevance of their corresponding inputs to the query.

\textbf{(Retrieval) Self-Attention.}
To summarize the retrieved samples independently of the query,
Cross-RAG applies self-attention over their outputs:
\begin{equation}
\boldsymbol{s}
= \mathrm{Attn}\!\left(
\boldsymbol{R}_{y},\;
\boldsymbol{R}_{y},\;
\boldsymbol{R}_{y}
\right) + \boldsymbol{R}_{y},
\end{equation}
followed by a feed-forward network and aggregation:
\begin{equation}
\tilde{\boldsymbol{s}}
= \mathrm{Pool}\!\left(
\mathrm{FFN}_{\mathrm{self}}(\boldsymbol{s}) + \boldsymbol{s}
\right),
\end{equation}
where $\mathrm{Pool}(\cdot)$ denotes mean pooling over the retrieved samples.
This operation captures retrieval-driven information by aggregating the outputs of the top-$k$ retrieved samples.

Finally, we combine the two sources of information as:
\begin{equation}
\boldsymbol{z}
= \lambda \tilde{\boldsymbol{c}}
+ (1 - \lambda) \tilde{\boldsymbol{s}} ,
\end{equation}
where $\lambda \in [0,1]$ is a fixed hyperparameter controlling
the relative contribution of each source of information.
The fused representation $\boldsymbol{z}$,
which encodes (i) the query, (ii) the retrieved samples, and (iii) their relational information,
is passed to the forecasting
head to produce the prediction $\hat{\boldsymbol{y}}$.

%% file: icml_04_experiments_settings.tex
\section{Experiments}

\subsection{Experimental Setup}
\textbf{Pretraining datasets.} For pretraining, we use the Chronos pretraining dataset~\cite{ansari2024chronos}, from which 50M data points are uniformly sampled, following the setup of prior work~\cite{ning2025ts}. A subset of 5M data points is further selected to construct a retrieval knowledge base. Both the pretraining dataset and the retrieval knowledge base are segmented using a fixed input window, resulting in 26M pretraining pairs and 2.8M retrieval pairs.
Specifically, the 50M data points are sampled from the full Chronos pretraining corpus, which includes both real-world and synthetic datasets, and 5M of them are randomly selected to construct the retrieval knowledge base. 
The training splits of downstream evaluation datasets are excluded from the retrieval knowledge base to ensure fair zero-shot evaluation.

\textbf{Evaluation datasets.}
Zero-shot evaluation is conducted on widely used time series benchmarks spanning diverse domains, four ETT datasets (ETTh1, ETTh2, ETTm1, ETTm2) \cite{zhou2021informer},
Weather, Electricity, and Exchange \cite{wu2021autoformer}.
We maintain chronological order when separating the training, validation, and test sets, with split ratios of 6:2:2 for the ETT datasets and 7:1:2 for the remaining datasets.
Details of datasets are provided in Appendix~\ref{sec:data}.

\textbf{Baseline methods.}
We compare our method with several TSFMs,
including
Chronos, Chronos-Bolt~\cite{ansari2024chronos},
MOMENT~\cite{goswami2024moment},
TTM~\cite{ekambaram2024tiny},
Moirai~\cite{woo2024unified},
TimesFM~\cite{das2024decoder},
and Time-MoE~\cite{shi2024time}.
Details of baselines are provided in Appendix~\ref{sec:baseline}.

\textbf{Experimental settings.}
To remain consistent with standard settings, we fix the input length and forecasting horizon to 512 and 64, respectively, for both pretraining and evaluation,
and use Chronos-Bolt as the TSFM backbone.
For the evaluation metrics, we use mean squared error (MSE) and mean absolute error (MAE), and set $k=15$ and $\lambda=0.7$.
Note that we use $k=5$ for TS-RAG, which yields the best performance among values in [1, 15].
Details 
%of experimental settings 
are reported in Appendix~\ref{sec:exp_details}.

%% file: icml_04_experiments_main.tex
\begin{table*}[t]
% \vspace{-10pt}
\caption{\textbf{Results of zero-shot forecasting across TSFMs.}
The proposed method outperforms existing TSFMs across diverse datasets.
%, achieving up to a 4.8\% MSE improvement over the best baseline.
The 1$^{\mathrm{st}}$ and 2$^{\mathrm{nd}}$ results are indicated by \textbf{bold} and \underline{underline}, respectively.
}
\centering
% \vspace{-3pt}
\begin{adjustbox}{max width=1.000\textwidth}
\begin{NiceTabular}{lcccccccccccccccc}
          \toprule
\multirow{2}{*}{\textbf{Methods}} &
\multicolumn{2}{c}{\multirow{2}{*}{\shortstack{\textbf{Cross-RAG}\\ (Ours)}}} &
\multicolumn{2}{c}{\multirow{2}{*}{\shortstack{\textbf{Chronos-bolt}\\ (TMLR 2024)}}} &
\multicolumn{2}{c}{\multirow{2}{*}{\shortstack{\textbf{MOMENT}\\ (ICML 2024)}}} &
\multicolumn{2}{c}{\multirow{2}{*}{\shortstack{\textbf{TTM}\\ (NeurIPS 2024)}}} &
\multicolumn{2}{c}{\multirow{2}{*}{\shortstack{\textbf{Moirai}\\ (ICML 2024)}}} &
\multicolumn{2}{c}{\multirow{2}{*}{\shortstack{\textbf{TimesFM}\\ (ICML 2024)}}} &
\multicolumn{2}{c}{\multirow{2}{*}{\shortstack{\textbf{Chronos}\\ (TMLR 2024)}}} &
\multicolumn{2}{c}{\multirow{2}{*}{\shortstack{\textbf{Time-MoE}\\ (ICLR 2025)}}} \\
\\
\cmidrule(lr){2-3} \cmidrule(lr){4-5} \cmidrule(lr){6-7} \cmidrule(lr){8-9} \cmidrule(lr){10-11} \cmidrule(lr){12-13} \cmidrule(lr){14-15} \cmidrule(lr){16-17}
\textbf{Metric} & MSE & MAE & MSE & MAE & MSE & MAE & MSE & MAE & MSE & MAE & MSE & MAE & MSE & MAE & MSE & MAE \\
\midrule
ETTh1 &
\cellcolor{LightYellow} \textbf{0.341} & \cellcolor{LightYellow} \underline{0.367} &
\underline{0.362} & \textbf{0.365} &
0.392 & 0.411 &
0.362 & 0.371 &
0.369 & 0.384 &
0.425 & 0.383 &
0.422 & 0.381 &
0.362 & \underline{0.367} \\
ETTh2 &
\cellcolor{LightYellow} \textbf{0.243} & \cellcolor{LightYellow} \textbf{0.299} &
\underline{0.252} & \textbf{0.299} &
0.274 & 0.333 &
0.253 & 0.303 &
0.255 & 0.305 &
0.289 & 0.323 &
0.266 & 0.314 &
0.252 & 0.322 \\
ETTm1 &
\cellcolor{LightYellow} \textbf{0.290} & \cellcolor{LightYellow} \textbf{0.319} &
\underline{0.311} & \textbf{0.319} &
0.351 & 0.383 &
0.315 & \underline{0.325} &
0.540 & 0.432 &
0.332 & 0.333 &
0.394 & 0.370 &
0.321 & 0.334\\

ETTm2 &
\cellcolor{LightYellow} \textbf{0.143} & \cellcolor{LightYellow} \textbf{0.224} &
\underline{0.149} & \textbf{0.224} &
0.170 & 0.258 &
0.151 & \underline{0.241} &
0.196 & 0.269 &
0.170 & 0.255 &
0.166 & 0.252 &
0.157 & 0.254
\\
Weather &
\cellcolor{LightYellow} \textbf{0.144} & \cellcolor{LightYellow} \textbf{0.178} &
\underline{0.153} & \underline{0.183} &
0.180 & 0.238 &
0.154 & 0.189 &
0.171 & 0.191 &
--- & --- &
0.190 & 0.211 &
0.149 & 0.184 \\
Electricity &
\cellcolor{LightYellow} \textbf{0.112} & \cellcolor{LightYellow} \textbf{0.200} &
\underline{0.113} & \textbf{0.200} &
0.197 & 0.303 &
0.172 & 0.264 &
0.183 & 0.281 &
--- & --- &
0.146 & \underline{0.224}  &
0.114 & 0.203\\
Exchange &
\cellcolor{LightYellow} \textbf{0.064} & \cellcolor{LightYellow} \underline{0.175} &
{0.067} & 0.178 &
0.098 & 0.206 &
\underline{0.066} & \textbf{0.173} &
\underline{0.066} & 0.172 &
0.070 & 0.180 &
0.083 & 0.188 &
0.085 & 0.206\\
\midrule
Average & 
\cellcolor{LightYellow} \textbf{0.191} &  \cellcolor{LightYellow} \textbf{0.252} & %ours
\underline{0.201} & \underline{0.253} & % Chronos-bolt
0.237 & 0.304 & %MOMENT
0.210 & 0.267 & % TTMb
0.254 & 0.291 & % MoiraiB
--- & --- & % TimesFM
0.238 & 0.277 & % ChronosB
0.206 & 0.267 \\      % TimeMOEB
\bottomrule
\end{NiceTabular}%
\end{adjustbox}
\label{tbl:compare_tsfm}
% \vspace{-12pt}
\end{table*}

\begin{table*}[t]
% \vspace{-10pt}
%%% Table 4: RAG comparison %%%
\begin{minipage}{\textwidth}
\centering
\caption{
\textbf{Results of zero-shot forecasting across TS RAG methods.}
The proposed method outperforms existing RAG methods.
}
% \vspace{-3pt}
\begin{adjustbox}{max width=1.000\textwidth}
\begin{NiceTabular}{lccccccc|c}
\toprule
Method&ETTh1& ETTh2&ETTm1& ETTm2& Weather& Electricity& Exchange& Average\\
\midrule
RAF (arXiv 2025)& 0.366& 0.252& 0.306& 0.148& 0.178& 0.119& \textbf{0.063}& 0.205 \\
TS-RAG (NeurIPS 2025)& \underline{0.357}& \underline{0.246}& \underline{0.298}& \underline{0.147}& \underline{0.151} & \textbf{0.112} & 0.071& \underline{0.197} \\
\rowcolor{LightYellow} Cross-RAG (Ours)& \textbf{0.341} & \textbf{0.243} & \textbf{0.290} & \textbf{0.143} & \textbf{0.144} & \textbf{0.112} & \underline{0.064} & \textbf{0.191} \\
\bottomrule
\end{NiceTabular}
\end{adjustbox}
\label{tbl:compare_rag}
\end{minipage}

\vspace{15pt}

%%% Table 5: Application to various TSFM backbones (Chronos-Bolt + MOMENT) %%%
\begin{minipage}{\textwidth}
\centering
\caption{\textbf{Application to various TSFM backbones.} Cross-RAG generalizes across both Chronos-Bolt and MOMENT, consistently outperforming the standalone backbones and TS-RAG.}
% \vspace{-3pt}
\begin{adjustbox}{max width=1.000\textwidth}
\begin{NiceTabular}{c|l|ccccccc|c}
\toprule
Backbone & Method & ETTh1 & ETTh2 & ETTm1 & ETTm2 & Weather & Electricity & Exchange & Average \\
\midrule
\multirow{3}{*}{\textbf{Chronos-Bolt}}
& (standalone)
& 0.362 & 0.252 & 0.311 & \underline{0.149} & 0.153 & \underline{0.113} & \underline{0.067} & 0.201 \\
& + TS-RAG
& \underline{0.357} & \underline{0.246} & \underline{0.298} & \underline{0.147} & \underline{0.151} & \textbf{0.112} & 0.071 & \underline{0.197} \\
\rowcolor{LightYellow}
& + Cross-RAG
& \textbf{0.341} & \textbf{0.243} & \textbf{0.290} & \textbf{0.143} & \textbf{0.144} & \textbf{0.112} & \textbf{0.064} & \textbf{0.191} \\
\midrule
\multirow{3}{*}{\textbf{MOMENT}}
& (standalone)
& 0.392 & 0.274 & \textbf{0.351} & \underline{0.170} & 0.180 & 0.197 & 0.098 & 0.237 \\
& + TS-RAG
& \underline{0.370} & \underline{0.272} & 0.366 & 0.171 & \underline{0.164} & \textbf{0.154} & \underline{0.095} & \underline{0.227} \\
\rowcolor{LightYellow}
& + Cross-RAG
& \textbf{0.367} & \textbf{0.269} & \underline{0.358} & \textbf{0.166} & \textbf{0.157} & \underline{0.159} & \textbf{0.089} & \textbf{0.224} \\
\bottomrule
\end{NiceTabular}
\end{adjustbox}
\label{tbl:multi_backbone}
\end{minipage}
\end{table*}

\subsection{Zero-shot Forecasting}
% To validate the effectiveness of Cross-RAG, we evaluate it from three perspectives:
We validate the effectiveness of Cross-RAG from three perspectives:
%under zero-shot forecasting settings:
\begin{itemize}
    \item (1) Comparison with TSFMs
    \item (2) Comparison with RAG methods
    \item (3) Application to various TSFMs
\end{itemize}
The baseline results are obtained both by reproducing experiments using the official implementations and by taking the reported results from the original paper~\cite{ning2025ts}.

\textbf{[1] Comparison with TSFMs.}
We compare our method with various
TSFMs under the zero-shot forecasting setting. All models are evaluated without any task-specific fine-tuning, following the standard protocol of zero-shot forecasting in RAG.
Table~\ref{tbl:compare_tsfm} presents the results, demonstrating the effectiveness of our method in incorporating retrieval information for zero-shot forecasting and showing consistent improvements over TSFMs that rely solely on pretrained representations.

\textbf{[2] Comparison with RAG methods.}
We compare our method with RAG-basedshot forecasting across approaches under the same
zero-shot forecasting setting.
Table~\ref{tbl:compare_rag} shows the results,
highlighting the effectiveness of selectively attending to query-rshot forecasting acrosselevant retrieved samples.
As visualized in Figure~\ref{fig:viz_ts}, Cross-RAG follows the ground truth more closely than TS-RAG.
Additional visualizations are provided in Section~\ref{sec:add_viz}.

\textbf{[3] Application to various TSFMs.}
To verify that Cross-RAG is backbone-agnostic, we evaluate it with two TSFM backbones: Chronos-Bolt~\cite{ansari2024chronos} and MOMENT~\cite{goswami2024moment}.
Table~\ref{tbl:multi_backbone} presents the results, comparing each backbone without RAG, with TS-RAG, and with Cross-RAG.
Cross-RAG achieves the best average MSE across both backbones (0.191 for Chronos-Bolt and 0.224 for MOMENT), consistently outperforming both baselines across most datasets.
These results confirm that the proposed cross-attention mechanism generalizes beyond a single backbone and is effective across different TSFMs.

% %%%%%%%%%%%%%%%%%%%%%%%%%%%%%%%%%%%%%%%%%%%%%%%%%%%%%%%%%%
% % NEW: Electricity leakage analysis
% %%%%%%%%%%%%%%%%%%%%%%%%%%%%%%%%%%%%%%%%%%%%%%%%%%%%%%%%%%
% \rebuttal{
% \textbf{Data leakage analysis.}
% It has been noted that the full Electricity dataset is included in the Chronos pretraining corpus, which may raise concerns about test data leakage. To investigate this, we conduct an experiment where we remove all Electricity-related data from the retrieval knowledge base and re-evaluate on the remaining six datasets. The results are identical to the full-KB setting (e.g., ETTh1: 0.341, Weather: 0.144), confirming that the presence of Electricity in the KB does not inflate performance on other datasets. We further note that the Electricity result (MSE 0.112) is consistent with the backbone performance (Chronos-Bolt: 0.113), indicating that the marginal improvement stems from the RAG mechanism rather than memorization. Full results are reported in Appendix~\ref{sec:electricity_leakage}.
% }

\begin{table*}[t]
\begin{minipage}{\textwidth}
\centering
\caption{\textbf{Ablation on fusion components.}
$Q$ denotes the \textit{skip-connection} from the \textbf{query} representation,
$R$ denotes retrieval \textit{self-attention} that summarizes \textbf{retrieved samples},
and $Q \leftrightarrow R$ denotes \textbf{query--retrieval} \textit{cross-attention}.
% between the query and retrieved samples.
}
% \vspace{-4pt}
\begin{adjustbox}{max width=1.000\textwidth}
\begin{NiceTabular}{ccc|ccccccc|c}
\toprule
$Q$ & $R$ & $Q \leftrightarrow R$ &ETTh1& ETTh2&ETTm1& ETTm2& Weather& Electricity& Exchange& Average\\
\midrule
\cmark &  &  & 0.423 & 0.321 & 0.359 & 0.204 & 0.195 & 0.184 & 0.172 & 0.265 \\
 \cmark & \cmark &  &  0.361 & \second{0.249} & 0.308 & \second{0.148} & 0.151 & \second{0.113} & \second{0.067} & 0.200 \\
\cmark &  & \cmark &  \second{0.360} & 0.246 & \second{0.300} & \second{0.148} & \second{0.147} & \second{0.113} & \first{0.066} & \second{0.197} \\
\rowcolor{LightYellow} \cmark & \cmark & \cmark & \textbf{0.341} & \textbf{0.243} & \textbf{0.290} & \textbf{0.143} & \textbf{0.144} & \textbf{0.112} & \textbf{0.064} & \textbf{0.191} \\
\bottomrule
\end{NiceTabular}
\end{adjustbox}
\label{tbl:ablation_brief}
\end{minipage}
\end{table*}

%% file: icml_04_experiments_ablation_modules.tex
\subsection{Ablation Studies}
We conduct ablation studies to evaluate the contributions of Cross-RAG's components:
query skip-connection ($Q$),
retrieval self-attention ($R$),
and query--retrieval cross-attention ($Q \leftrightarrow R$).
Table~\ref{tbl:ablation_brief} shows the results, demonstrating that using only the query representation ($Q$) yields limited performance, while adding either retrieval self-attention ($R$) or query--retrieval cross-attention ($Q \leftrightarrow R$) consistently improves results.
Combining all three components achieves the best performance across all datasets, highlighting the importance of jointly modeling the query, retrieved samples, and their input-level interactions.
Additional ablation combinations are reported in Section~\ref{sec:ablation_full}.

%% file: icml_05_analysis1.tex
\section{Analysis}
\subsection{Various Retrieval Scenarios}
To validate the effectiveness of our method under various retrieval scenarios, 
we evaluate it across four different settings as follows:
\begin{itemize}
\item (R1) Retrieval with \textbf{various numbers of samples ($k$)}
\item (R2) Retrieval of \textbf{random samples}
\item (R3) Retrieval using a \textbf{smaller knowledge base}
\item (R4) Retrieval from a \textbf{different dataset}
\end{itemize}

\textbf{(R1) Retrieval with various $k$s.}
Table \ref{tbl:various_k_simple} and Figure \ref{fig:motivation_k} show the average MSE over seven datasets for various numbers of retrieved samples ($k$).
Unlike prior work \cite{ning2025ts}, Cross-RAG consistently improves as $k$ increases.
These results indicate that simply increasing the number of retrieved samples does not improve performance unless relevance is considered, and that Cross-RAG effectively leverages additional samples by focusing on relevant ones.
Full results for all datasets are reported in Section~\ref{sec:various_k_full}.

\begin{figure}[t]
\vspace{10pt}
\centering
\begin{adjustbox}{max width=\linewidth}
\includegraphics[width=\columnwidth]{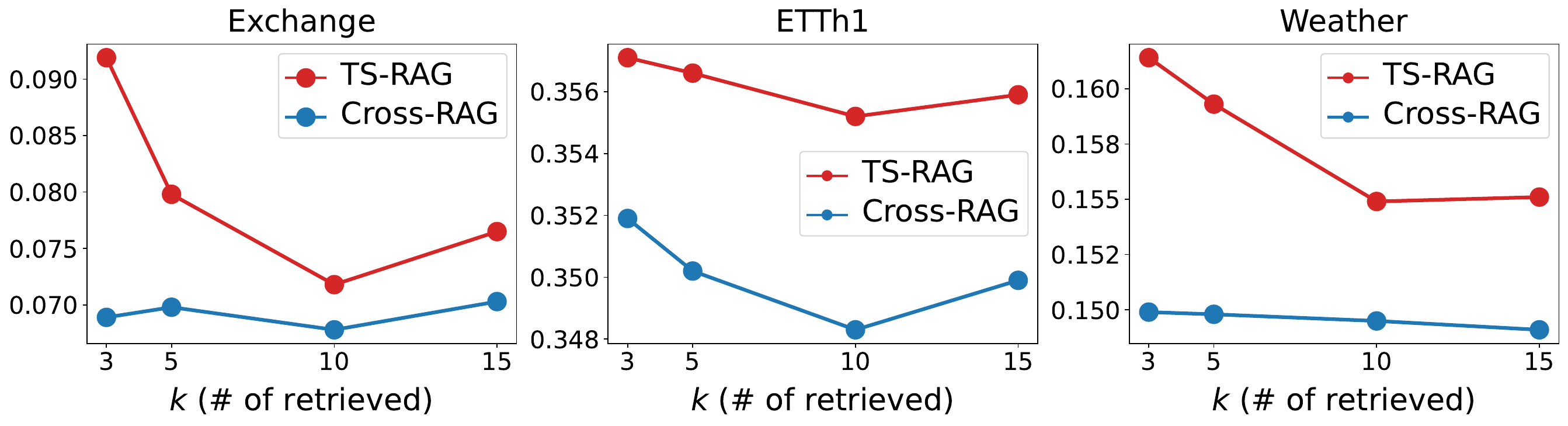}
\end{adjustbox}
\caption{\textbf{(R2) MSE under random retrieval.} Selective attention enables Cross-RAG to filter out irrelevant samples even under random retrieval.}
\label{fig:random_retrieval}
\end{figure}

\textbf{(R2) Retrieval of random samples.}
To evaluate Cross-RAG's ability to selectively attend to relevant retrieved samples,
we conduct an experiment using \textit{random} samples from the knowledge base, rather than selecting them based on similarity.
Figure~\ref{fig:random_retrieval} presents the MSE over three datasets, showing that Cross-RAG consistently outperforms the SoTA method~\cite{ning2025ts} for all values of $k$.

% --- (R1) Table and (R2) Figure side by side ---
% --- (R1) MSE/Rank under various k (standalone one-column table) ---
\begin{table}[t]
\centering
\caption{\textbf{(R1) MSE/Rank under various $k$s.} Cross-attention enables Cross-RAG to benefit from larger $k$.
%without performance degradation.
}
\begin{adjustbox}{max width=\linewidth}
\begin{NiceTabular}{c|ccccccc}
\toprule
$k$ & 2 & 4 & 6 & 8 & 10 & 12 & 14 \\
\midrule
\multirow{2}{*}{TS-RAG} &
.1975&.2011&.1968&.1975&.1967&.1974&.2013\\
& 4 & 6 & 2 & 4 & 1 & 3 & 7 \\
\midrule
\rowcolor{LightYellow}
\multirow{2}{*}{Cross-RAG} &
\textbf{.1944}&\textbf{.1938}&\textbf{.1936}&\textbf{.1934}&\textbf{.1933}&\textbf{.1930}&\textbf{.1927}\\
& 7 & 6 & 5 & 4 & 3 & 2 & 1 \\
\bottomrule
\end{NiceTabular}%
\end{adjustbox}
\label{tbl:various_k_simple}
\end{table}

% --- (R2) MSE under random retrieval (standalone one-column figure) ---

% --- (R3) MSE under smaller knowledge base (standalone one-column table) ---
\begin{table}[t]
\centering
\caption{\textbf{(R3) MSE under smaller knowledge base.} Cross-RAG remains effective when only a fraction of the training data is available for retrieval.}
\begin{adjustbox}{max width=\linewidth}
\begin{NiceTabular}{cccc|cc}
\toprule
\multicolumn{4}{c}{Train period (\%)} &\multicolumn{2}{c}{Methods}\\
\cmidrule(lr){1-4} \cmidrule(lr){5-6}
Q1 & Q2 & Q3 & Q4 & TS-RAG & Cross-RAG  \\
\midrule
\cellcolor{blue!10} \cmark &  \cellcolor{orange!15} \cmark  & \cellcolor{green!15} \cmark  & \cellcolor{red!15} \cmark  & .1470 & \textbf{.1435} \\
\cellcolor{LightGray3} \xmark &  \cellcolor{orange!15} \cmark  & \cellcolor{green!15} \cmark  & \cellcolor{red!15} \cmark  & .1485 & \textbf{.1463} \\
\cellcolor{LightGray3} \xmark &  \cellcolor{LightGray3} \xmark  & \cellcolor{green!15} \cmark  & \cellcolor{red!15} \cmark  & .1486 & \textbf{.1458} \\
\cellcolor{LightGray3} \xmark &  \cellcolor{LightGray3} \xmark  & \cellcolor{LightGray3} \xmark & \cellcolor{red!15} \cmark  & .1489 & \textbf{.1467} \\
\bottomrule
\end{NiceTabular}
\end{adjustbox}
\label{tbl:small_kb}
\end{table}

% --- (R3) Visualization of TS segments (standalone one-column figure) ---
\begin{figure}[t]
\centering
\begin{adjustbox}{max width=\linewidth}
\includegraphics[width=\columnwidth]{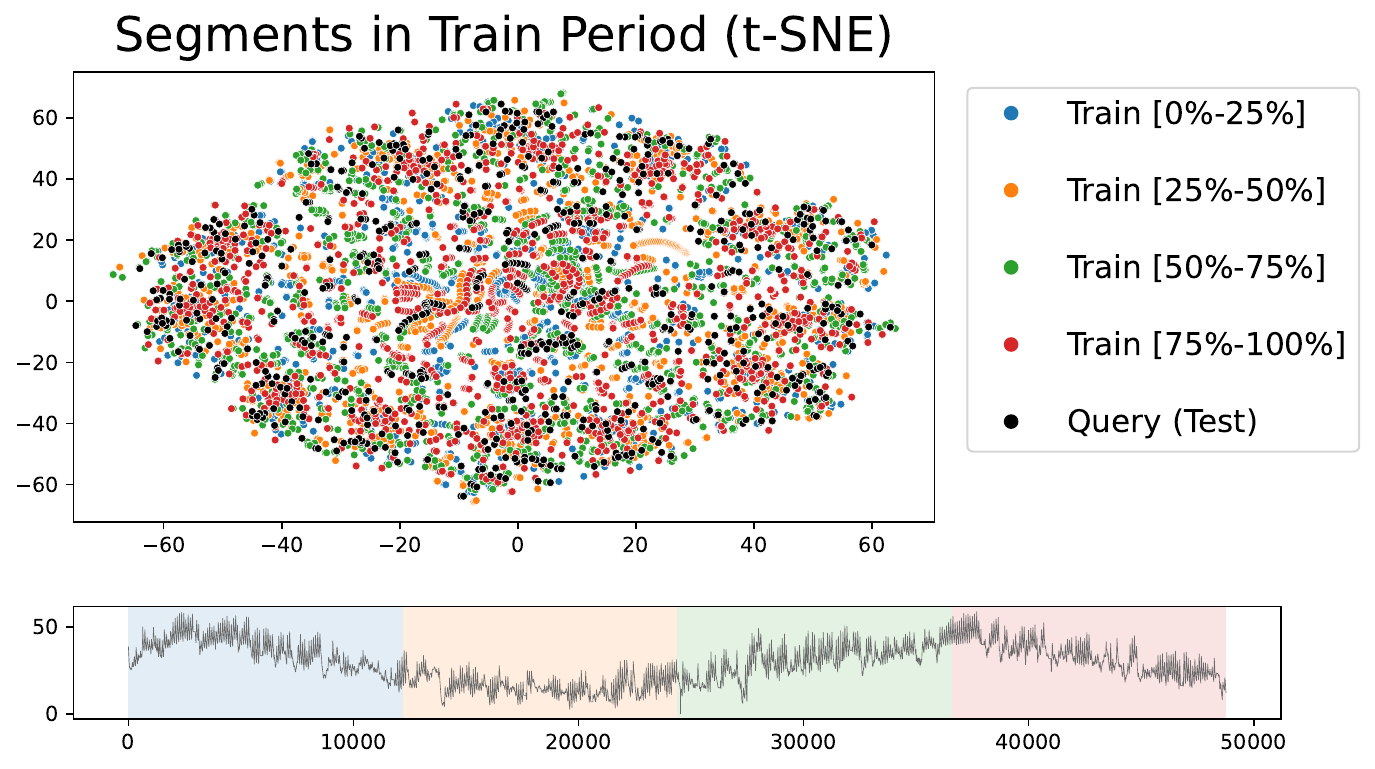}
\end{adjustbox}
\caption{\textbf{(R3) Visualization of TS segments.} The figure illustrates that historical TS segments contain redundant patterns, supporting Cross-RAG's robustness under smaller KBs.}
\label{fig:small_kb}
\end{figure}

\textbf{(R3) Retrieval using a smaller knowledge base.}
To evaluate the robustness of Cross-RAG under limited knowledge, we construct knowledge bases (KBs) using \textit{different fractions of the training dataset} using ETTm2,
as shown in Figure~\ref{fig:small_kb}.
Table~\ref{tbl:small_kb}
demonstrates
that Cross-RAG consistently outperforms TS-RAG by effectively leveraging the available information even when the KB is small.

Furthermore, we attribute this to the high redundancy of similar segments in historical data, as illustrated in Figure~\ref{fig:small_kb}. The figure shows that as recent segments cover a large portion of the relevant patterns, even a reduced KB provides sufficient context for accurate retrieval, allowing Cross-RAG to maintain strong performance.
We further conduct a sensitivity analysis by varying the KB size from 5\% to 100\% across all datasets, reported in Appendix~\ref{sec:kb_sensitivity}. The results confirm that Cross-RAG degrades gracefully even when the KB is reduced to 5\% of its original size.

\begin{table}[t]
% \caption{\textbf{(R4) Retrieval from a different dataset.} Cross-RAG generalizes across datasets, with stronger performance when the knowledge base shares the same temporal granularity as the target.}
\caption{\textbf{(R4) Retrieval from a different dataset.} Selective attention enables Cross-RAG to transfer useful signals across datasets, supporting retrieval-augmented forecasting even without target-domain KBs.}
\centering
\begin{adjustbox}{max width=\linewidth}
\begin{NiceTabular}{c|c|c|c|c|c}
\toprule
\multicolumn{2}{c}{\multirow{2.5}{*}{\shortstack{KB $\rightarrow$ Target\\ (KB: Knowledge Base)}}}& \multicolumn{4}{c}{Target} \\
\cmidrule{3-6}
\multicolumn{2}{c}{ } & ETTh1 & ETTh2 & ETTm1 & ETTm2\\
\midrule
\multirow{6}{*}{KB} & ETTh1 & \first{0.341} & \underline{0.246} & 0.325 & 0.149 \\
\cmidrule{2-6}
& ETTh2 &  \second{0.351} & \first{0.243} & 0.318 & 0.148 \\
\cmidrule{2-6}
& ETTm1 &  0.370 & {0.250} & \first{0.290} & \second{0.146} \\
\cmidrule{2-6}
& ETTm2 &  0.363 & 0.252 & \second{0.304} & \first{0.143} \\
\midrule
\midrule
\multicolumn{2}{c}{w/o Cross-RAG} &  0.362 & 0.252 & 0.311 & 0.149 \\
\bottomrule
\end{NiceTabular}
\end{adjustbox}
\label{tbl:transfer_setting}
\end{table}
\textbf{(R4) Retrieval from a different dataset.}
We further evaluate Cross-RAG in a setting where \textit{no knowledge base is available from the target dataset} by constructing the knowledge base from \textit{different} datasets.
Specifically, we treat one of the four ETT datasets as the target and use the remaining three as the knowledge base.
Table~\ref{tbl:transfer_setting} reports the results,
where Cross-RAG consistently outperforms the strong baseline (Chronos-Bolt) in most cases.
In addition, stronger performance is observed when the knowledge base comes from the same or a similar domain (e.g., ETT hourly or minutely), indicating more effective information transfer across aligned datasets.
An extended transfer analysis with additional target datasets (Weather, Electricity, Exchange) is provided in Appendix~\ref{sec:transfer_extended}.

%% file: icml_05_analysis2.tex
% \clearpage
\subsection{Other Analyses}
\textbf{Performance across various $T,L$s.}
Table~\ref{tbl:various_TL} reports the average MSE across various input windows ($T$) and forecast horizons ($L$).
We fix $L=64$ when varying the input window size ($T$),
and fix $T=512$ when varying the forecast horizon ($L$).
The proposed method consistently outperforms TS-RAG across all settings, with improvements pronounced for smaller input windows and longer forecast horizons.
We further extend the evaluation to longer $L$ with a fixed $T{=}512$, as reported in Appendix~\ref{sec:various_horizons}. Cross-RAG maintains consistent improvements across all extended horizons.

% --- Table 10: Average MSE across various T,L (standalone one-column) ---
\begin{table}[t]
\centering
\caption{\textbf{Average MSE across various $T$,$L$.} Cross-RAG consistently improves over TS-RAG across 
% all input windows and forecast horizons.
all $T$s and $L$s.
}
\begin{adjustbox}{max width=\linewidth}
\begin{NiceTabular}{l|cccc}
\toprule
\multicolumn{5}{l}{(1) \textbf{Input window} ($T$) with $L{=}64$}\\
\midrule
$T$ & 64 & 128 & 256 & 512 \\
\midrule
TS-RAG & 0.463 & 0.376 & 0.217 & 0.197 \\
\rowcolor{LightYellow} Cross-RAG & \first{0.258} & \first{0.246} & \first{0.196} & \first{0.191} \\
\midrule
% Imp. (\%) & \first{+44.3} & \first{+34.6} & \first{+9.7} & \first{+3.0} \\
Improvement (\%) & \first{+44.3} & \first{+34.6} & \first{+9.7} & \first{+3.0} \\
\midrule
\midrule
\multicolumn{5}{l}{(2) \textbf{Forecast horizon} ($L$) with $T{=}512$} \\
\midrule
$L$ & 24 & 36 & 48 & 64 \\
\midrule
TS-RAG  & 0.140 & 0.164 & 0.181 & 0.197\\
\rowcolor{LightYellow} Cross-RAG & \first{0.138} & \first{0.163} & \first{0.178} & \first{0.191} \\
\midrule
% Imp. (\%)  & \first{+1.4} & \first{+0.6} & \first{+1.7} &  \first{+3.0}\\
Improvement (\%)  & \first{+1.4} & \first{+0.6} & \first{+1.7} &  \first{+3.0}\\
\bottomrule
\end{NiceTabular}
\end{adjustbox}
\label{tbl:various_TL}
\end{table}

% --- Table 11: Robustness to similarity metrics (standalone one-column) ---
\begin{table}[t]
\centering
\caption{\textbf{Robustness to similarity metrics.} Performance remains stable across metrics.}
\begin{adjustbox}{max width=\linewidth}
\begin{NiceTabular}{c|ccc|c}
\toprule
& \multicolumn{3}{c}{Data space} & \multirow{2.5}{*}{Latent} \\
\cmidrule(lr){2-4}
 & Cosine & Euclidean & Correlation & \\
\cmidrule{1-1}  \cmidrule(lr){2-2} \cmidrule(lr){3-3} \cmidrule(lr){4-4} \cmidrule(lr){5-5}
ETTh1 & 0.341 & 0.343 & 0.348 & 0.347 \\
ETTh2 & 0.243 & 0.248 & 0.245 & 0.244 \\
ETTm1 & 0.290 & 0.294 & 0.297 & 0.294 \\
ETTm2 & 0.143 & 0.143 & 0.144 & 0.144 \\
Weather & 0.144 & 0.146 & 0.146 & 0.145 \\
Electricity & 0.112 & 0.114 & 0.115 & 0.114 \\
Exchange & 0.064 & 0.067 & 0.069 & 0.067 \\
\midrule
Average & \first{0.191} & \second{0.193} & 0.194 & \second{0.193} \\
\bottomrule
\end{NiceTabular}
\end{adjustbox}
\label{tbl:robust_metric}
\end{table}

% --- Figure 8: Attention of query and retrieved samples (standalone one-column) ---
\begin{figure}[t]
\centering
\begin{adjustbox}{max width=\linewidth}
\includegraphics[width=\columnwidth]{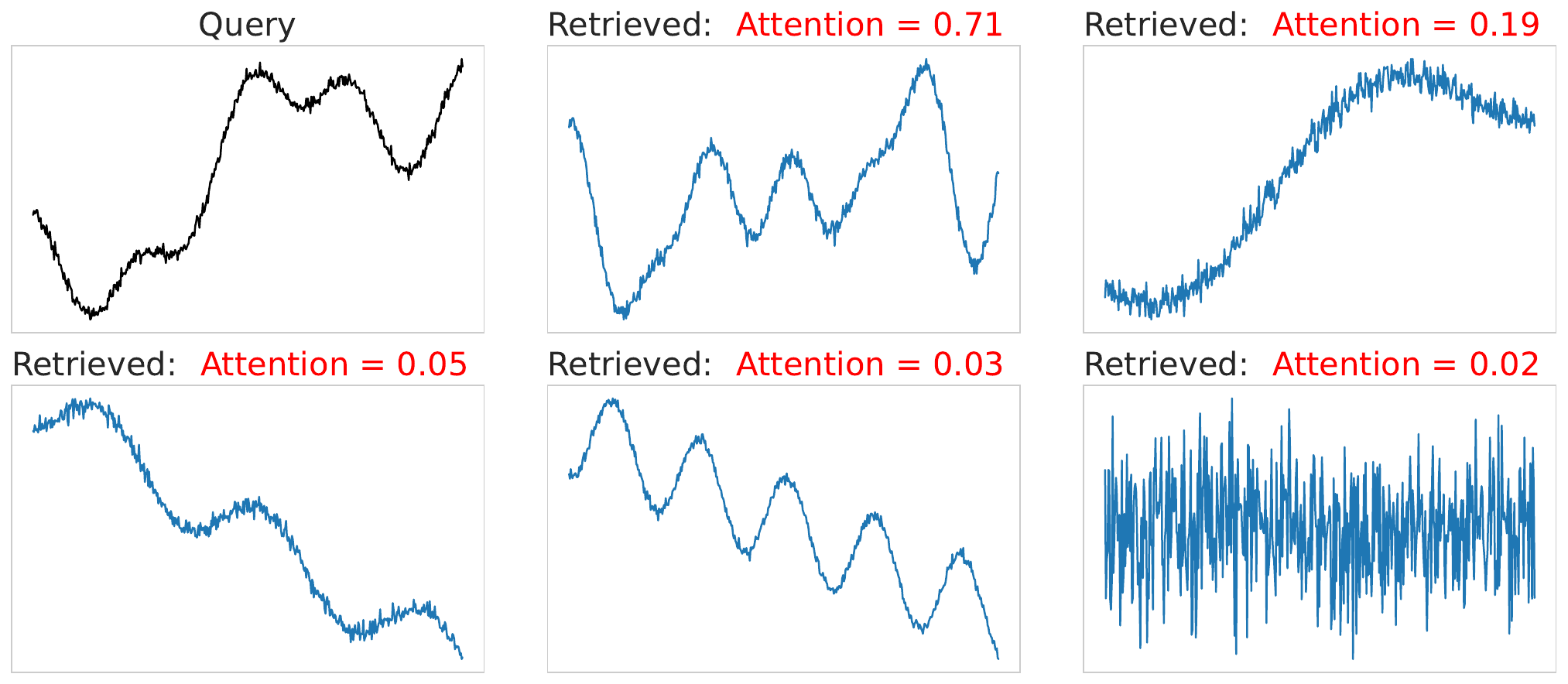}
\end{adjustbox}
\caption{\textbf{Attention of query and retrieved samples.}
% Cross-attention assign high attention to relevant samples and low attention to irrelevant ones.}
Cross-attention assign higher attention to more relevant samples.}
\label{fig:attention_ex}
\end{figure}

% --- Table 12: Robustness to lambda (standalone one-column) ---
\textbf{Attention weights across retrieved samples.}
To examine how our proposed method attends to each retrieved sample for a given query, we construct a toy dataset and manually compose the retrieved set to include both relevant and irrelevant samples for each query.
Figure~\ref{fig:attention_ex} visualizes the results along with the (cross-) attention weights, showing that samples similar to the query receive higher weights, while irrelevant or noisy samples receive lower weights.
More examples are visualized in Section~\ref{sec:q_r_att_viz_full}.

\textbf{Robustness to similarity metrics.}
To examine the robustness to the choice of similarity metric used for retrieval,
we conduct experiments under various similarity metrics across seven datasets.
Specifically, we consider cosine similarity, Euclidean distance, correlation, and a latent-space metric (Euclidean).
Table~\ref{tbl:robust_metric} reports the results,
demonstrating that the performance remains stable across similarity metrics,
suggesting that retrieval in the data space can be conducted without introducing an additional retrieval encoder.

\textbf{Robustness to $\lambda$.}
To evaluate the robustness to $\lambda$,
which controls the fusion of retrieved samples and the relational information between the query and these samples,
we perform a sensitivity analysis
% across different $\lambda$ values.
across various $\lambda$s.
We also test a learnable version of this gating mechanism,
where two types of representations are concatenated and passed through an MLP followed by a sigmoid function.

Table~\ref{tbl:robust_lambda} reports the results,
demonstrating that the performance is stable across $\lambda$ and that the fixed gating mechanism outperforms the learnable version.
We attribute this to the high variability introduced by instance-level gating: while a learnable $\lambda$ can in principle adapt to each query, the limited training signal per instance leads to noisy gradient estimates that prevent stable convergence. In contrast, a fixed $\lambda$ acts as a regularizer that stabilizes the fusion across diverse retrieval scenarios. 
A detailed per-dataset analysis of the learnable gating mechanism is provided in Appendix~\ref{sec:learnable_lambda}.
% Detailed analysis of the learnable gating mechanism is provided in Appendix~\ref{sec:learnable_lambda}.

\begin{table}[t]
\centering
\caption{\textbf{Robustness to $\lambda$.} A simple fixed $\lambda$ outperforms learnable gating and remains stable across a wide range of values.}
\begin{adjustbox}{max width=\linewidth}
\begin{NiceTabular}{c|ccccc}
\toprule
\multirow{2.5}{*}{$\lambda$} & \multicolumn{5}{c}{Fixed}\\
\cmidrule(lr){2-6}
& 0.4 & 0.5 & 0.6 & 0.7 & 0.8 \\
\midrule
MSE & 0.196 & 0.194 & 0.193 & \textbf{0.191} & 0.194 \\
\midrule
\midrule
& \multicolumn{5}{c}{Learnable} \\
\midrule
MSE & \multicolumn{5}{c}{0.196} \\
\bottomrule
\end{NiceTabular}
\end{adjustbox}
\label{tbl:robust_lambda}
\end{table}

\subsection{Efficiency Analyses}
\label{sec:efficiency}

% --- Efficiency table as wraptable ---
\begin{table}[t]
\caption{
\textbf{Efficiency - Retrieval.}
Data-space retrieval achieves a 339$\times$ speed-up by performing retrieval directly on raw inputs, without any embedding overhead.
}
\centering
\begin{adjustbox}{max width=\linewidth}
\begin{NiceTabular}{c|c|c}
\toprule
Space & Procedures & Time (s) \\
\midrule
\multirow{3}{*}{Latent}
& Embedding & 381.6  \\
& Search & 2.1 \\
\cmidrule(lr){2-3}
& Total & 383.7 \\
\midrule
\multirow{3}{*}{Data}
& \cellcolor{LightYellow} Embedding & \cellcolor{LightYellow} --- \\
& \cellcolor{LightYellow} Search & \cellcolor{LightYellow} 1.1 \\
\cmidrule(lr){2-3}
& \cellcolor{LightYellow} Total & \cellcolor{LightYellow} 1.1 \\
\midrule
\multicolumn{2}{c}{Speed-up} & \textbf{339$\times$}  \\
\bottomrule
\end{NiceTabular}
\end{adjustbox}
\label{tbl:eff_time}
\end{table}

\textbf{Efficiency analysis 1 - Retrieval.}
Table~\ref{tbl:eff_time} compares retrieval in the data space 
%(without embeddings) 
and latent space 
%(with embeddings) 
using ETTh1.
%The results show that data space retrieval is significantly faster, achieving a 339.5$\times$ speed-up over latent space retrieval.
%In contrast to latent space retrieval, data space retrieval does not require any embedding computation, as retrieval is directly performed on the raw input space.
The results show that data-space retrieval is significantly faster, achieving a 339.5$\times$ speed-up over latent-space retrieval, 
%since it directly operates on the raw input space without requiring 
as it does not require
any embedding computation.
Search with FAISS corresponds to the offline construction of the retrieval index, where similar historical TS across all timestamps and variables are precomputed and stored. During training and inference, only the stored index is accessed, without performing FAISS search repeatedly. Results for all datasets are discussed in Section~\ref{sec:eff_time_all}.

\textbf{Efficiency analysis 2 - Inference.}
Table~\ref{tbl:eff_flops} compares inference efficiency on ETTh1 under different combinations of Cross-RAG components in terms of 1) inference time per instance (ms, averaged across 1000 runs) and 2) FLOPs.
The inference cost remains nearly unchanged when incorporating the proposed components, demonstrating minimal computational overhead.
%The inference cost remains largely unchanged when incorporating the proposed components, indicating negligible computational overhead at inference time.

\textbf{Efficiency analysis 3 - Parameters.}
%As shown in Table~\ref{tbl:eff_params}, the proposed attention modules and projectors account for about 4\% of the total parameters, with all other modules frozen, resulting in high parameter efficiency. Note that these learnable parameters are pretrained \textit{only on the pretraining datasets} and remain fixed across all target datasets, corresponding to the zero-shot setting.
As shown in Table~\ref{tbl:eff_params}, the proposed modules (e.g., attention and projector) account for approximately 4\% of the total parameters, with all other modules frozen, ensuring high parameter efficiency. These learnable parameters are pretrained \textit{only on the pretraining datasets} and remain fixed across all target datasets, consistent with the zero-shot setting.

% --- Table 14: Efficiency - Inference (standalone one-column) ---
\begin{table}[t]
\centering
\caption{
\textbf{Efficiency - Inference.}
Cross-RAG incurs negligible inference overhead, demonstrating that input-aware fusion can be achieved without sacrificing efficiency.
}
\begin{adjustbox}{max width=\linewidth}
\begin{NiceTabular}{ccc|cc}
\toprule
$Q$ & $R$ & $Q \leftrightarrow R$ & Time (ms) & FLOPS \\
\midrule
\cmark &  &  & 8.90 & 751.3M \\
 \cmark & \cmark &  &  9.69 & 755.5M \\
\cmark &  & \cmark &  9.68 & 751.8M \\
\cellcolor{LightYellow} \cmark & \cellcolor{LightYellow} \cmark & \cellcolor{LightYellow} \cmark & \cellcolor{LightYellow} 9.73 & \cellcolor{LightYellow} 756.5M \\
\midrule
\multicolumn{3}{c}{TS-RAG} & 9.91 & 754.2M \\
\bottomrule
\end{NiceTabular}
\end{adjustbox}
\label{tbl:eff_flops}
\end{table}

% --- Table 15: Efficiency - Parameters (standalone one-column) ---
\begin{table}[t]
\centering
\caption{
\textbf{Efficiency - Parameters.} With only $\sim$4\% \colorbox{LightYellow}{trainable} parameters, Cross-RAG operates as a lightweight plug-in over the frozen TSFM backbone.
}
\begin{adjustbox}{max width=\linewidth}
\begin{NiceTabular}{c|c|c|cc}
\toprule
\multicolumn{3}{c}{Components} & \# Params & (\%) \\
\midrule
\midrule
\multicolumn{2}{c}{Patch Embedding} & \cellcolor{LightGray3} Input / Output & \cellcolor{LightGray3} 7.1M & \cellcolor{LightGray3} 3.3 \\
\midrule
\multicolumn{2}{c}{Backbone} & \cellcolor{LightGray3} Enc. / Dec. & \cellcolor{LightGray3} 198M & \cellcolor{LightGray3} 92.6 \\
\midrule
\multirow{4}{*}{\textbf{+Ours}} & \multirow{2}{*}{Attention} & \cellcolor{LightYellow} Self & \cellcolor{LightYellow} 3.5M & \cellcolor{LightYellow} \textbf{1.6} \\
&  & \cellcolor{LightYellow} Cross & \cellcolor{LightYellow} 3.5M & \cellcolor{LightYellow} \textbf{1.6} \\
\cmidrule{2-5}
&  \multirow{2}{*}{Projector} & \cellcolor{LightYellow} Input & \cellcolor{LightYellow} 1.0M & \cellcolor{LightYellow} \textbf{0.5} \\
& & \cellcolor{LightYellow} Output & \cellcolor{LightYellow} 0.6M & \cellcolor{LightYellow} \textbf{0.3} \\
\bottomrule
\end{NiceTabular}
\end{adjustbox}
\label{tbl:eff_params}
\end{table}

%% file: icml_06_conclusion.tex
\section{Conclusion}
In this work, we introduce Cross-RAG, a zero-shot retrieval-augmented framework for TS forecasting that selectively attends to query-relevant samples and models their input-level interactions via cross-attention.
Our results show consistent improvements in zero-shot settings, demonstrating effective and efficient selective retrieval that is robust to retrieval settings, distance metrics, and the choice of top-$k$.
We further validate generalizability across TSFM backbones (e.g., Chronos-Bolt and MOMENT) and provide analyses on knowledge base sensitivity and extended forecast horizons.

\textbf{Limitation and future works.}
While Cross-RAG relies on input-level relationships for selective retrieval, it does not exploit auxiliary information such as metadata.
Furthermore, 
following the common evaluation protocols used in prior RAG-based time series forecasting methods, 
our evaluation focuses on conventional forecasting datasets, and extending to large-scale benchmarks (e.g., GIFT-Eval~\cite{aksu2024gift}) would further strengthen the findings.
Future work includes adaptive retrieval and interaction mechanisms that incorporate such auxiliary information across datasets.
We also provide an empirical analysis showing that learnable gating introduces instability due to noisy per-instance gradients, and a deeper theoretical understanding of the optimal fusion dynamics remains an open problem.
Finally, as Cross-RAG is a backbone-agnostic framework, we hope it will naturally extend to future TSFMs, as it relies on input-level interactions without requiring architecture-specific modifications.
% \rebuttal{We also acknowledge the following limitations: (1) the current evaluation primarily uses conventional forecasting benchmarks, and evaluation on large-scale benchmarks such as GIFT-Eval would further strengthen the findings; (2) while we demonstrate backbone generalization with two architectures (Chronos-Bolt and MOMENT), extending to additional TSFMs (e.g., TimesFM, Moirai) would provide broader empirical evidence; (3) the knowledge base is constructed from the Chronos pretraining corpus, and exploring the effect of knowledge base composition on downstream performance remains an open question; (4) the fixed scalar $\lambda$ for fusion, while empirically effective and stable, lacks a formal theoretical justification. We have provided an empirical analysis showing that learnable gating introduces instability due to noisy per-instance gradients, but a deeper theoretical understanding of the optimal fusion dynamics is left for future work.}

% \clearpage

%% file: icml_99_appendix.tex
\clearpage
\appendix

\section{Experimental Settings}
\label{sec:exp_setting}
\subsection{Dataset Statistics}
\label{sec:data}
\textbf{Inference dataset for zero-shot forecasting.}
We conduct experiments on seven datasets spanning diverse domains,
with dataset statistics summarized in Table~\ref{tab:dataset_stat}, where $C$ and $T$ denote the number of channels and timesteps, respectively.
The training split is used to construct the knowledge base, while the test split is used for zero-shot inference.

\begin{table}[h]
\centering
\caption{Statistics of inference dataset for zero-shot forecasting.}
\begin{adjustbox}{max width=\linewidth}
\begin{NiceTabular}{l|c|c|c}
\toprule
Dataset & $C$ & $T$ & $(N_\text{train},N_\text{val},N_\text{test})$ \\
\midrule
ETTh1 \cite{zhou2021informer} & 7 & 17420 & (8545, 2881, 2881) \\
ETTh2 \cite{zhou2021informer} &  7& 17420 & (8545, 2881, 2881)\\
 ETTm1 \cite{zhou2021informer} &  7& 69680 & (34465, 11521, 11521)  \\
 ETTm2 \cite{zhou2021informer} & 7 & 69680 & (34465, 11521, 11521)  \\
Exchange \cite{wu2021autoformer} & 8 & 7588 & (5120, 665, 1422) \\
Weather \cite{wu2021autoformer} & 21 & 52696 & (36792, 5271, 10540)  \\
Electricity \cite{wu2021autoformer}& 321 & 26304 & (18317, 2633, 5261) \\
\bottomrule
\end{NiceTabular}
\end{adjustbox}
\label{tab:dataset_stat}
\end{table}

\subsection{Experimental Details}
\label{sec:exp_details}
We follow the experimental settings of TS-RAG \cite{ning2025ts}, where the backbone TSFM remains frozen during pretraining, and only the parameters introduced by Cross-RAG, including attention modules and projectors, are updated.
Chronos-Bolt \cite{ansari2024chronos} serves as the backbone, and we adopt its original quantile regression loss.
All experiments are conducted on a single NVIDIA L40-48G GPU.
Details of training hyperparameters are summarized in Table~\ref{tab:training_hyperparameters}.

\begin{table}[h]
\centering
\caption{Details of training hyperparameters.}
\begin{adjustbox}{max width=\linewidth}
\begin{NiceTabular}{l|c}
\toprule
Optimizer & AdamW \\
Learning rate & $3\times10^{-4}$ \\
Weight decay & 0.01 \\
Batch size & 256 \\
Training steps & 10{,}000 \\
Dropout rate & 0.2 \\
Retrieved sequences ($k$) & 15\\
\bottomrule
\end{NiceTabular}
\end{adjustbox}
\label{tab:training_hyperparameters}
\end{table}

\vspace{30pt}

\section{Baseline Methods}
\label{sec:baseline}
\begin{itemize}
    \item \textbf{Chronos} \cite{ansari2024chronos}: A probabilistic TSFM that tokenizes quantized TS and processes them with a T5 backbone \cite{raffel2020exploring}.
    \item \textbf{Chronos-Bolt} \cite{ansari2024chronos}: A Chronos variant that supports patch-based modeling and generates multi-step quantile forecasts using decoder representations.
    \item \textbf{MOMENT} \cite{goswami2024moment}: A zero-shot TSFM applies masked modeling by appending a masked forecast horizon to the lookback sequence.
	\item \textbf{TTM} \cite{ekambaram2024tiny}: A compact TSFM based on TSMixer \cite{chen2023tsmixer} with adaptive patching and resolution-aware pretraining.
	\item \textbf{Moirai} \cite{woo2024unified}: A Transformer encoder pretrained on LOTSA via horizon masking and reconstruction across target channels.
	\item \textbf{TimesFM} \cite{das2024decoder}: A decoder-style attention model pretrained on large-scale real and synthetic TS.
	\item \textbf{Time-MoE} \cite{shi2024time}: A decoder-only TSFM with a mixture-of-experts (MoE) architecture for autoregressive forecasting with long contexts.
\end{itemize}

\vspace{30pt}

\section{Similarity Metrics}
For the experiments, we use four different similarity metrics for retrieving similar samples:
\begin{itemize}
\item \textbf{Cosine similarity (data space)} measures the similarity between two TS based on the angle between their vectors in the data space, capturing pattern-level similarity while being invariant to scale.
\item \textbf{Euclidean distance (data space)} computes the $\ell_2$  distance between raw TS, directly reflecting point-wise magnitude differences.
\item \textbf{Pearson correlation (data space)} quantifies the linear relationship between two TS, providing a normalized similarity measure invariant to mean and variance.
\item \textbf{Euclidean distance (latent space)} measures the $\ell_2$ distance between latent representations produced by a retrieval encoder, capturing similarity in the learned representation space.
\end{itemize}

For cosine similarity and Euclidean distance in the data space, each TS is first normalized to the $[0,1]$ range using min--max scaling before computing the similarity.
As robustness is consistently observed with these similarity metrics, Dynamic Time Warping (DTW) is omitted due to its high computational complexity.

\vspace{30pt}

\section{Ablation Studies}
\label{sec:ablation_full}
We conduct an ablation study to assess the effect of the query skip-connection ($Q$), retrieval self-attention ($R$), and query--retrieval cross-attention ($Q \leftrightarrow R$).
Table~\ref{tbl:ablation_full} reports the full results, where combining all three components yields the best performance.

\begin{table}[h]
\caption{\textbf{Full results of ablation on fusion components.}
$Q$ denotes the \textit{skip-connection} from the \textbf{query} representation,
$R$ denotes retrieval \textit{self-attention} that summarizes \textbf{retrieved samples.}}
\centering
\begin{adjustbox}{max width=\linewidth}
\begin{NiceTabular}{ccc|ccccccc|c}
\toprule

$Q$ & $R$ & $Q \leftrightarrow R$ &ETTh1& ETTh2&ETTm1& ETTm2& Weather& Electricity& Exchange& Average\\
\midrule
\cmark &  &  & 0.423 & 0.321 & 0.359 & 0.204 & 0.195 & 0.184 & 0.172 & 0.265 \\
 & \cmark & &  0.717 & 0.364 & 0.697 & 0.284 & 0.263 & 0.848 & 0.364 & 0.505 \\
 &  & \cmark &  0.405 & 0.282 & 0.348 & 0.177 & 0.171 & 0.168 & 0.128 & 0.240 \\
 \cmark & \cmark &  &  0.361 & 0.249 & 0.308 & 0.148 & 0.151 & 0.113 & 0.067 & 0.200 \\
\cmark &  & \cmark &  0.360 & 0.246 & 0.300 & 0.148 & 0.147 & 0.113 & 0.066 & 0.197 \\
 & \cmark & \cmark &  0.393 & 0.279 & 0.329 & 0.171 & 0.164 & 0.164 & 0.115 & 0.231 \\
\cmark & \cmark & \cmark & \textbf{0.341} & \textbf{0.243} & \textbf{0.290} & \textbf{0.143} & \textbf{0.144} & \textbf{0.112} & \textbf{0.064} & \textbf{0.191} \\
\bottomrule
\end{NiceTabular}
\end{adjustbox}
\label{tbl:ablation_full}
\end{table}

\clearpage

\section{Various Number of Retrieved Sequences ($k$s)}
\label{sec:various_k_full}
To assess zero-shot forecasting performance under different numbers of retrieved samples ($k$), we conduct an evaluation with varying values of $k$.
Table~\ref{tbl:various_K_full} reports the results, showing that the average performance across seven datasets improves as $k$ increases.

\begin{table}[h]
\caption{
Zero-shot forecasting performance (MSE) under various $k$s.
}
\centering
\begin{adjustbox}{max width=\linewidth}
\begin{NiceTabular}{c|ccccccc|c}
\toprule
$k$ & ETTh1 & ETTh2 & ETTm1 & ETTm2 & Weather & Electricity & Exchange & Average \\
\midrule
1  & 0.3513 & 0.2457 & 0.2986 & 0.1441 & 0.1462 & 0.1158 & 0.0656 & 0.1953 \\
2  & 0.3476 & 0.2464 & 0.2976 & 0.1435 & 0.1458 & 0.1145 & 0.0652 & 0.1944 \\
3  & 0.3466 & 0.2467 & 0.2966 & 0.1433 & 0.1454 & 0.1143 & 0.0648 & 0.1940 \\
5  & 0.3461 & 0.2468 & 0.2945 & 0.1435 & 0.1455 & 0.1141 & 0.0653 & 0.1937 \\
8  & 0.3444 & 0.2469 & 0.2925 & 0.1436 & 0.1449 & 0.1139 & 0.0658 & 0.1934 \\
10 & 0.3451 & 0.2467 & 0.2922 & 0.1425 & 0.1447 & 0.1142 & 0.0675 & 0.1933 \\
12 & 0.3449 & 0.2462 & 0.2912 & 0.1434 & 0.1450 & 0.1141 & 0.0663 & 0.1930 \\
15 & 0.3411 & 0.2427 & 0.2896 & 0.1435 & 0.1437 & 0.1123 & 0.0645 & \textbf{0.1911} \\
\bottomrule
\end{NiceTabular}%
\end{adjustbox}
\label{tbl:various_K_full}
\end{table}

\vspace{30pt}

\section{Efficiency Analysis}
\label{sec:eff_time_all}
Table~\ref{tbl:efficiency_full} compares retrieval
in the data space and latent space using five different datasets.
Since ETTh1 and ETTh2 share the same data size, as do ETTm1 and ETTm2, we report only one dataset from each pair.
The results demonstrate that data-space retrieval is significantly faster than latent-space retrieval.

\begin{table}[h]
\caption{
Efficiency analysis under five different datasets.
}
\centering
\begin{adjustbox}{max width=\linewidth}
\begin{NiceTabular}{c|c|ccccc}
\toprule
\multirow{2.5}{*}{Space} & \multirow{2.5}{*}{Retrieval procedures} &  \multicolumn{5}{c}{Dataset}\\
\cmidrule(lr){3-7}
 &  & ETTh1 & ETTm1 & Exchange & Weather & ECL \\
\midrule
\midrule
\multirow{3}{*}{Latent space ($Z$)}
& 1) Embedding & 381.6  & 1547.4 & 196.7 & 3530.1 & 27514.5 \\
& 2) Search (FAISS) & 2.1 & 19.4 & 0.7 & 37.1 & 253.5 \\
\cmidrule(lr){2-7}
& Total & 383.7 & 1566.8 & 197.4 & 3587.2 & 27768.0 \\
\midrule
\multirow{3}{*}{Data space ($X$)}
& \cellcolor{LightYellow} 1) Embedding & \cellcolor{LightYellow} --- & \cellcolor{LightYellow} --- & \cellcolor{LightYellow} --- & \cellcolor{LightYellow} --- & \cellcolor{LightYellow} --- \\
& \cellcolor{LightYellow} 2) Search (FAISS) & \cellcolor{LightYellow} 1.1 & \cellcolor{LightYellow} 16.5 & \cellcolor{LightYellow} 0.4 & \cellcolor{LightYellow} 22.2 & \cellcolor{LightYellow} 129.3 \\
\cmidrule(lr){2-7}
& \cellcolor{LightYellow} Total & \cellcolor{LightYellow} 1.1 & \cellcolor{LightYellow} 16.5 & \cellcolor{LightYellow} 0.4 & \cellcolor{LightYellow} 22.2 & \cellcolor{LightYellow} 129.3\\
\bottomrule
\end{NiceTabular}%
\end{adjustbox}
\label{tbl:efficiency_full}
\end{table}

%%%%%%%%%%%%%%%%%%%%%%%%%%%%%%%%%%%%%%%%%%%%%%%%%%%%%%%%%%
% Inference and Parameter Efficiency tables moved to main body (icml_05_analysis2.tex)
%%%%%%%%%%%%%%%%%%%%%%%%%%%%%%%%%%%%%%%%%%%%%%%%%%%%%%%%%%
% \vspace{25pt}

%%%%%%%%%%%%%%%%%%%%%%%%%%%%%%%%%%%%%%%%%%%%%%%%%%%%%%%%%%
% NEW APPENDIX SECTIONS FOR REBUTTAL
%%%%%%%%%%%%%%%%%%%%%%%%%%%%%%%%%%%%%%%%%%%%%%%%%%%%%%%%%%

\vspace{30pt}

\section{Electricity Data Leakage Analysis}
\label{sec:electricity_leakage}

The full Electricity dataset is included in the Chronos pretraining corpus, which may raise concerns about unfair evaluation. To address this, we remove all Electricity-related data from the retrieval knowledge base and re-evaluate on the remaining six datasets. Table~\ref{tbl:no_electricity} compares the results with and without Electricity in the knowledge base. The results on all six non-Electricity datasets are nearly identical, confirming that the presence of Electricity data in the knowledge base does not inflate performance on other datasets.

\begin{table}[h]
\caption{\textbf{Data leakage analysis.} Results with and without Electricity in the knowledge base.}
% \vspace{10pt}
\centering
\begin{adjustbox}{max width=\linewidth}
\begin{NiceTabular}{l|cccccc}
\toprule
KB Setting & ETTh1 & ETTh2 & ETTm1 & ETTm2 & Weather & Exchange \\
\midrule
Full KB & 0.341 & 0.244 & 0.290 & 0.144 & 0.144 & 0.065 \\
w/o Electricity & 0.341 & 0.243 & 0.290 & 0.143 & 0.144 & 0.064 \\
\bottomrule
\end{NiceTabular}
\end{adjustbox}
\label{tbl:no_electricity}
\end{table}

\vspace{30pt}
% \clearpage
\section{Knowledge Base Sensitivity Analysis}
\label{sec:kb_sensitivity}

To assess the sensitivity of Cross-RAG to the size of the retrieval knowledge base, we vary the KB size from 5\% to 100\% of the full corpus and evaluate across all seven datasets. Table~\ref{tbl:kb_sensitivity} presents the results. Performance degrades gracefully as the KB size decreases: even at 5\% of the original KB, the average MSE is 0.212, compared to 0.191 at 100\%---representing only a ${\sim}$10\% relative degradation. Notably, ETTm2 is almost unaffected across all KB sizes, while Electricity shows the most sensitivity due to its large number of channels and diverse patterns.

\begin{table}[h]
\caption{\textbf{KB sensitivity analysis.} MSE across varying knowledge base sizes ($\alpha$\% of the full KB).}
\centering
\begin{adjustbox}{max width=\linewidth}
% \vspace{10pt}
\begin{NiceTabular}{l|ccccccc}
\toprule
$\alpha$ (\%) & 5 & 10 & 20 & 25 & 50 & 75 & 100 \\
\midrule
ETTh1 & 0.365 & 0.360 & 0.355 & 0.352 & 0.349 & 0.354 & 0.341 \\
ETTh2 & 0.251 & 0.246 & 0.246 & 0.247 & 0.248 & 0.250 & 0.243 \\
ETTm1 & 0.310 & 0.318 & 0.309 & 0.307 & 0.306 & 0.303 & 0.290 \\
ETTm2 & 0.147 & 0.146 & 0.147 & 0.147 & 0.146 & 0.147 & 0.143 \\
Weather & 0.157 & 0.154 & 0.149 & 0.155 & 0.152 & 0.148 & 0.144 \\
Electricity & 0.175 & 0.220 & 0.200 & 0.217 & 0.226 & 0.177 & 0.112 \\
Exchange & 0.079 & 0.070 & 0.075 & 0.078 & 0.069 & 0.074 & 0.064 \\
\midrule
Average & 0.212 & 0.216 & 0.212 & 0.215 & 0.213 & 0.207 & \textbf{0.191} \\
\bottomrule
\end{NiceTabular}
\end{adjustbox}
\label{tbl:kb_sensitivity}
\end{table}

\vspace{30pt}

\section{Learnable $\lambda$ Analysis}
\label{sec:learnable_lambda}

We provide a per-dataset breakdown of the learnable vs.\ fixed $\lambda$ comparison in Table~\ref{tbl:learnable_lambda_detail}. While the learnable $\lambda$ achieves competitive results on certain datasets (e.g., ETTh2: 0.243 vs.\ 0.246, Electricity: 0.113 vs.\ 0.113), it underperforms on others (e.g., ETTh1: 0.355 vs.\ 0.341). We attribute this to the high variance of instance-level gating gradients: since each training instance produces a different gating signal, the gradient updates for the gating network are noisy, preventing stable convergence. In contrast, a fixed $\lambda$ provides consistent regularization across instances.

\begin{table}[h]
\caption{\textbf{Per-dataset comparison of learnable vs.\ fixed $\lambda$.}}
\centering
\begin{adjustbox}{max width=\linewidth}
% \vspace{10pt}
\begin{NiceTabular}{l|ccccccc|c}
\toprule
$\lambda$ & ETTh1 & ETTh2 & ETTm1 & ETTm2 & Weather & Elec. & Exch. & Avg. \\
\midrule
Learnable & 0.355 & \textbf{0.243} & 0.295 & 0.147 & 0.147 & 0.113 & 0.065 & 0.195 \\
Fixed ($\lambda{=}0.7$) & \textbf{0.341} & \textbf{0.243} & \textbf{0.290} & \textbf{0.143} & \textbf{0.144} & \textbf{0.112} & \textbf{0.064} & \textbf{0.191} \\
\bottomrule
\end{NiceTabular}
\end{adjustbox}
\label{tbl:learnable_lambda_detail}
\end{table}

\clearpage
\section{Various Forecast Horizons}
\label{sec:various_horizons}

To evaluate Cross-RAG under extended forecast horizons, we conduct experiments with $L \in \{96, 128, 192\}$ using a fixed context length of $T{=}512$. Table~\ref{tbl:various_horizons_ext} reports the results. As expected, MSE increases with horizon length, but the degradation is smooth and the method maintains reasonable forecasting quality even at $L{=}192$. These results demonstrate that Cross-RAG is applicable beyond the default setting and generalizes to varying prediction requirements.

\begin{table}[h]
\caption{\textbf{Extended forecast horizons.} MSE with $T{=}512$ and varying prediction length $L$.}
\centering
\begin{adjustbox}{max width=\linewidth}
\begin{NiceTabular}{c|ccccccc|c}
\toprule
$L$ & ETTh1 & ETTh2 & ETTm1 & ETTm2 & Weather & Elec. & Exch. & Avg. \\
\midrule
64 & 0.341 & 0.243 & 0.290 & 0.143 & 0.144 & 0.112 & 0.064 & 0.191 \\
96 & 0.374 & 0.285 & 0.326 & 0.174 & 0.168 & 0.160 & 0.094 & 0.226 \\
128 & 0.398 & 0.311 & 0.356 & 0.202 & 0.188 & 0.187 & 0.124 & 0.252 \\
192 & 0.447 & 0.353 & 0.408 & 0.245 & 0.217 & 0.245 & 0.188 & 0.300 \\
\bottomrule
\end{NiceTabular}
\end{adjustbox}
\label{tbl:various_horizons_ext}
\end{table}

\vspace{30pt}
\section{Extended Transfer Learning Analysis}
\label{sec:transfer_extended}
We extend the transfer learning analysis from Section~5.1 by evaluating Cross-RAG with same-domain knowledge bases for Weather, Electricity, and Exchange datasets, in addition to the ETT datasets already reported. Table~\ref{tbl:transfer_extended} presents the results. Using same-domain knowledge bases yields competitive performance across most datasets, with Weather and Electricity showing slight degradation compared to the full cross-domain KB, indicating that these datasets benefit from the diversity of the cross-domain knowledge. The ETT datasets perform nearly identically under both settings, confirming that the model generalizes well to target-specific knowledge bases.

\begin{table}[h]
\caption{\textbf{Extended transfer learning.} MSE with same-domain vs.\ full cross-domain knowledge base.}
\centering
\begin{adjustbox}{max width=\linewidth}
\begin{NiceTabular}{l|ccccccc}
\toprule
KB Setting & ETTh1 & ETTh2 & ETTm1 & ETTm2 & Weather & Elec. & Exch. \\
\midrule
Same-domain & 0.345 & 0.247 & 0.296 & 0.146 & 0.154 & 0.157 & 0.066 \\
Cross-domain (full KB) & 0.341 & 0.243 & 0.290 & 0.143 & 0.144 & 0.112 & 0.064 \\
\bottomrule
\end{NiceTabular}
\end{adjustbox}
\label{tbl:transfer_extended}
\end{table}

\vspace{30pt}
\section{Visualization of Inputs of Query and Retrieved Samples}
\label{sec:q_r_att_viz_full}
\begin{figure}[h]
% \vspace{30pt}
    \centering
    \begin{subfigure}[t]{0.48\linewidth}
        \centering
        \begin{adjustbox}{max width=\linewidth}
        \includegraphics[width=\textwidth]{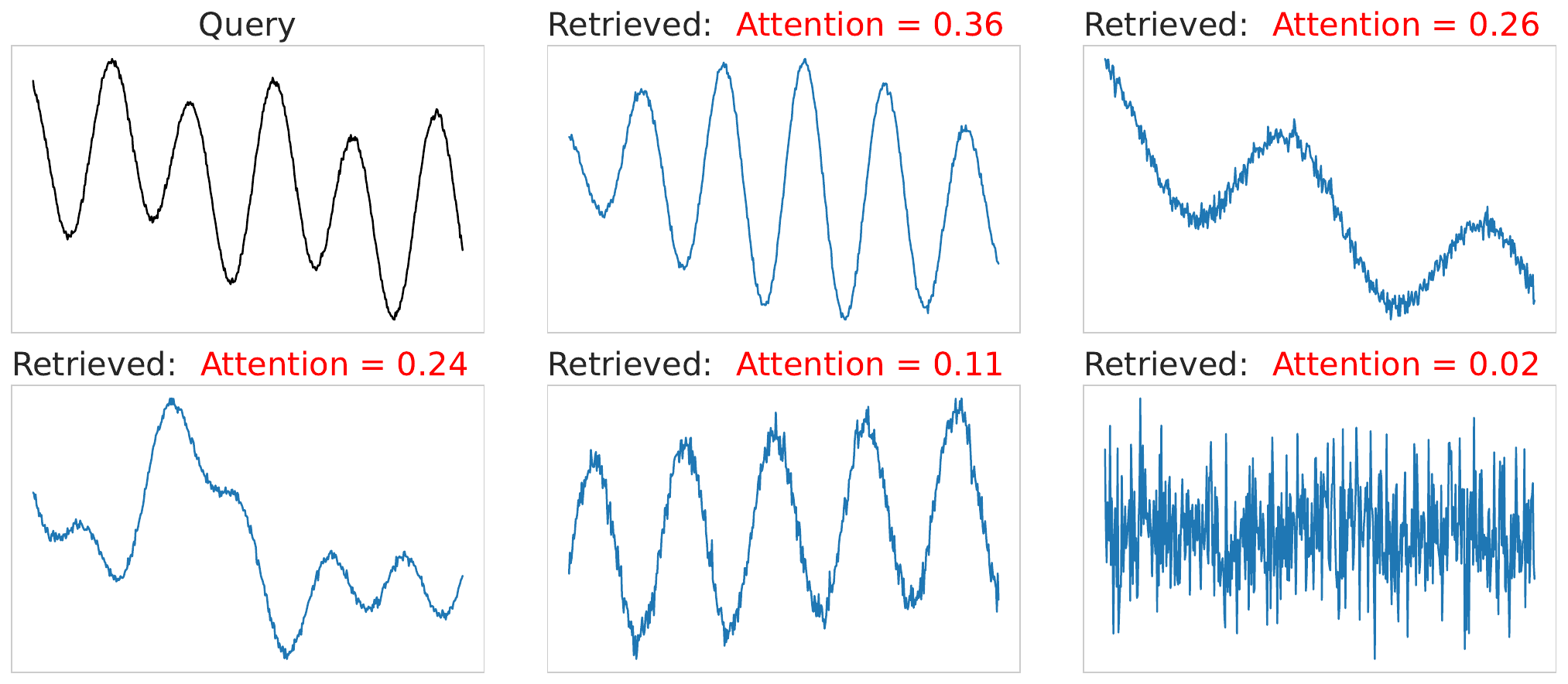}
        \end{adjustbox}
        \caption{Example 1}
    \end{subfigure}
    \hfill
    \begin{subfigure}[t]{0.48\linewidth}
        \centering
        \begin{adjustbox}{max width=\linewidth}
        \includegraphics[width=\textwidth]{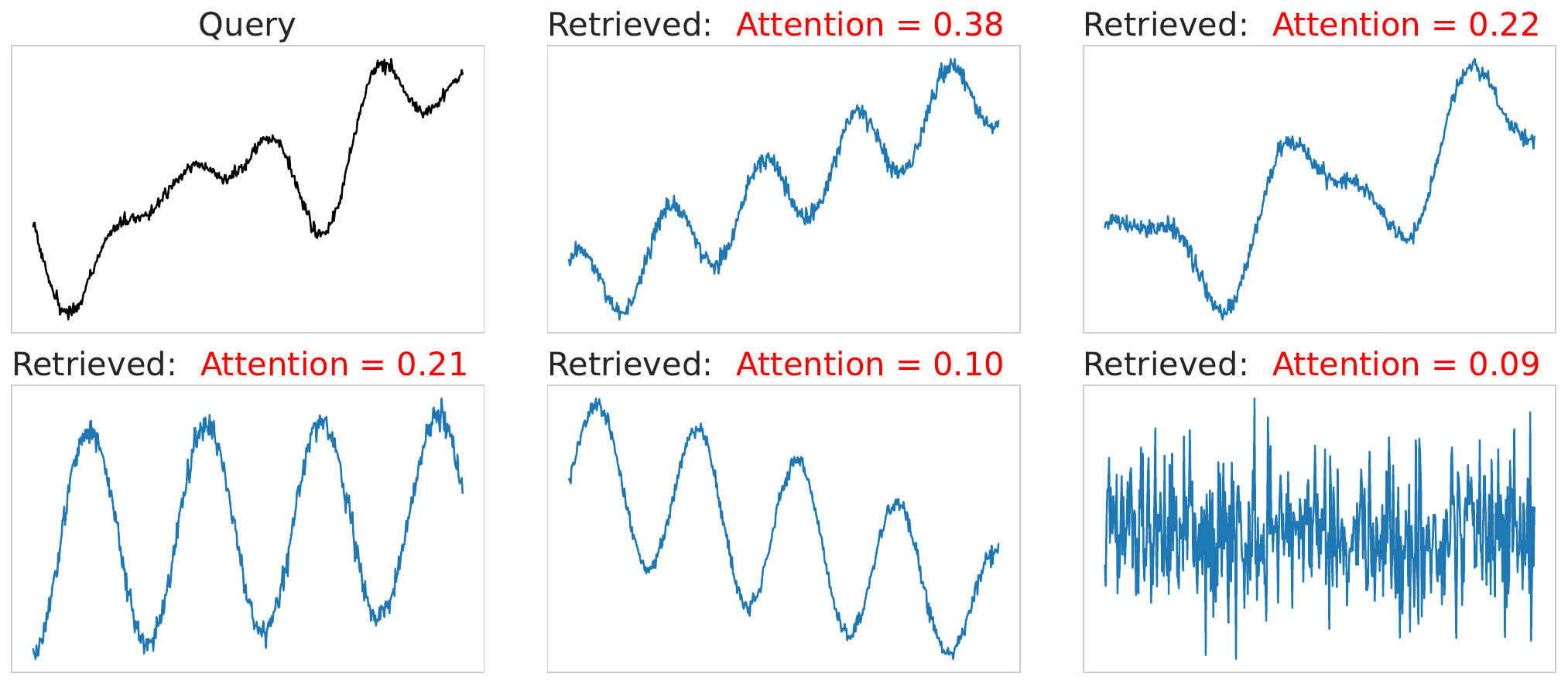}
        \end{adjustbox}
        \caption{Example 2}
    \end{subfigure}
    \begin{subfigure}[t]{0.48\linewidth}
        % \vspace{30pt}
        \centering
        \begin{adjustbox}{max width=\linewidth}
        \includegraphics[width=\textwidth]{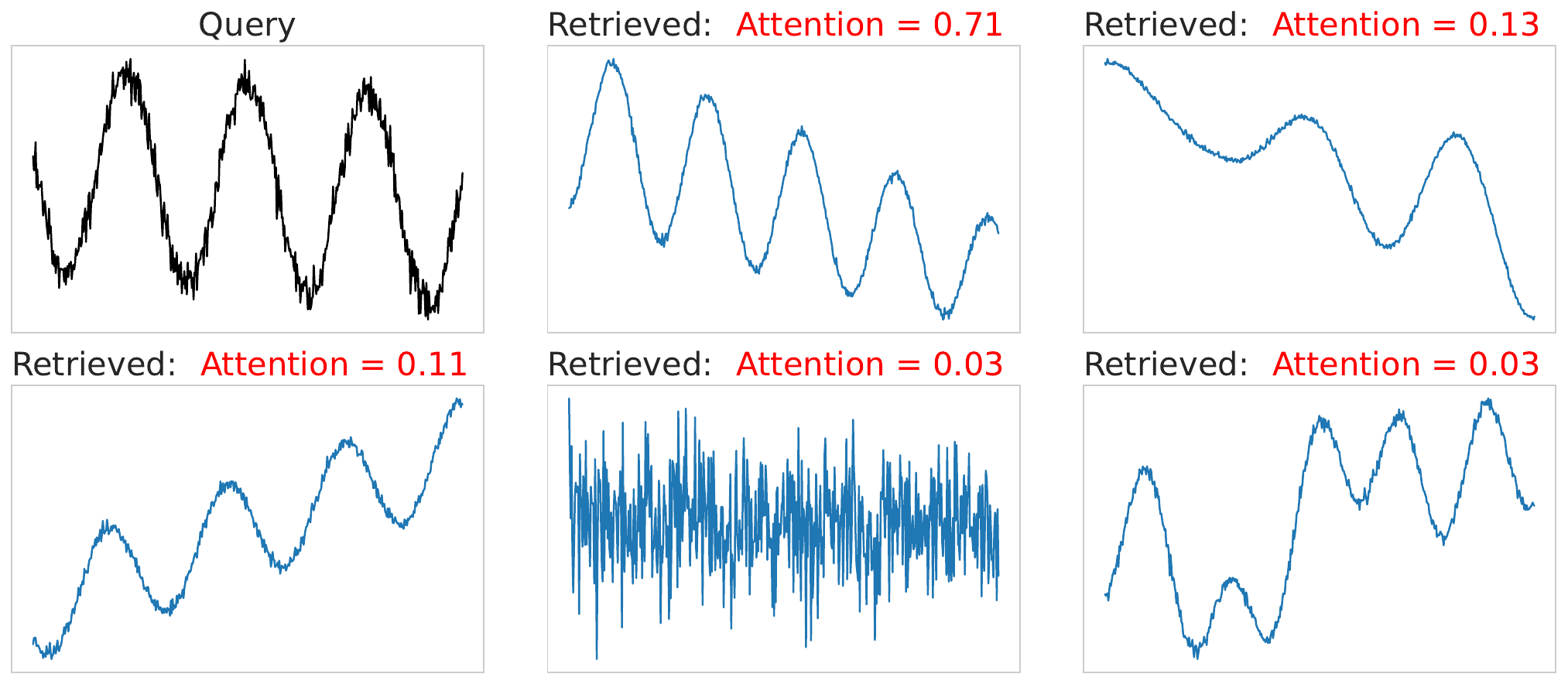}
        \end{adjustbox}
        \caption{Example 3}
    \end{subfigure}
    \hfill
    \begin{subfigure}[t]{0.48\linewidth}
        % \vspace{30pt}
        \centering
        \begin{adjustbox}{max width=\linewidth}
        \includegraphics[width=\textwidth]{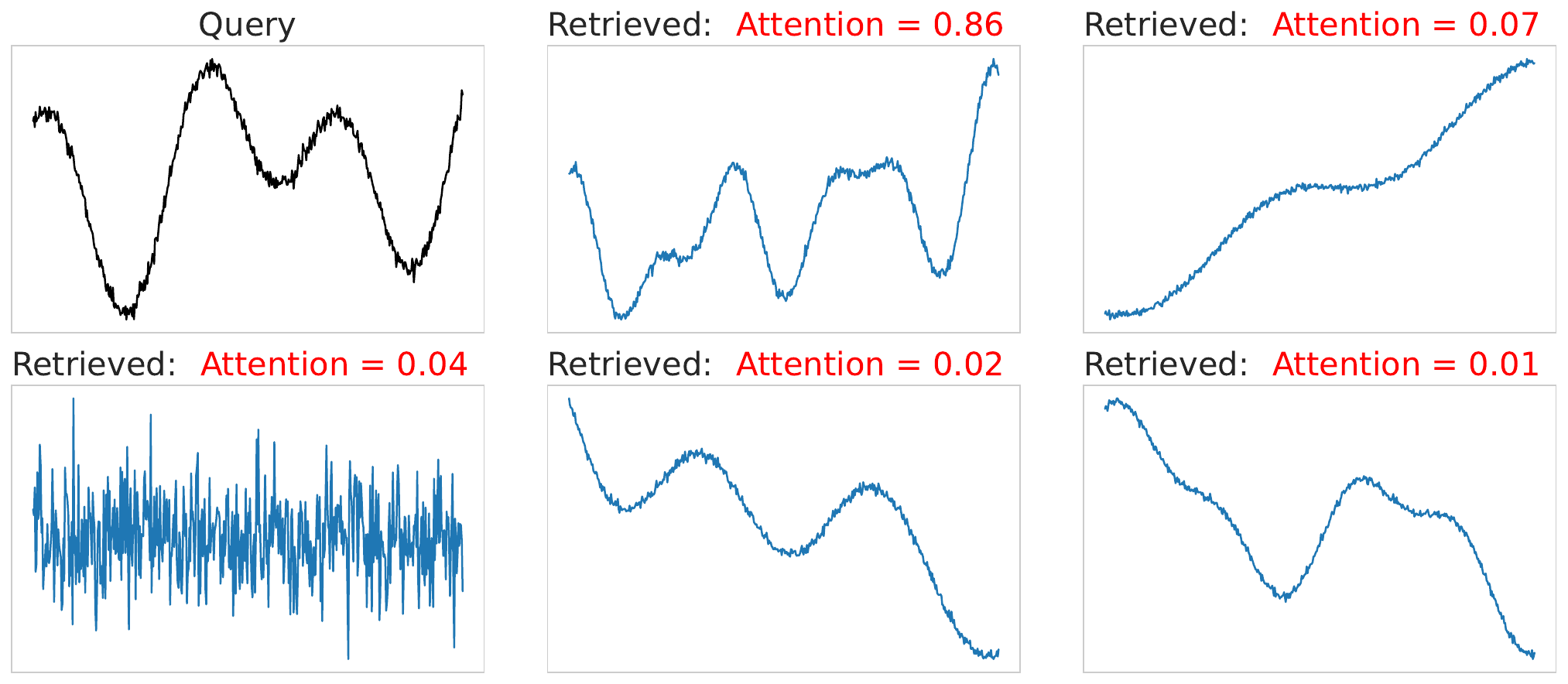}
        \end{adjustbox}
        \caption{Example 4}
    \end{subfigure}
    \caption{
    Visualization of inputs of query and retrieved samples with toy datasets.
    }
    \label{fig:full_attention}
\end{figure}

To analyze how Cross-RAG assigns attention to retrieved samples for a given query, we design a toy dataset containing both relevant and irrelevant retrieved sequences.
Figure~\ref{fig:full_attention} presents the resulting cross-attention weights,
where retrieved samples that resemble the query receive higher weights, whereas unrelated or noisy samples are assigned lower weights.

\vspace{100pt}
% \clearpage
\section{Visualization of Zero-shot Forecasting}
\label{sec:add_viz}
We visualize the predicted results on six different datasets and compare them with TS-RAG\cite{ning2025ts}.

\begin{figure}[H]
% \vspace{12pt}
    \centering
    \begin{subfigure}[t]{0.48\linewidth}
        \centering
        \begin{adjustbox}{max width=\linewidth}
        \includegraphics[width=\textwidth]{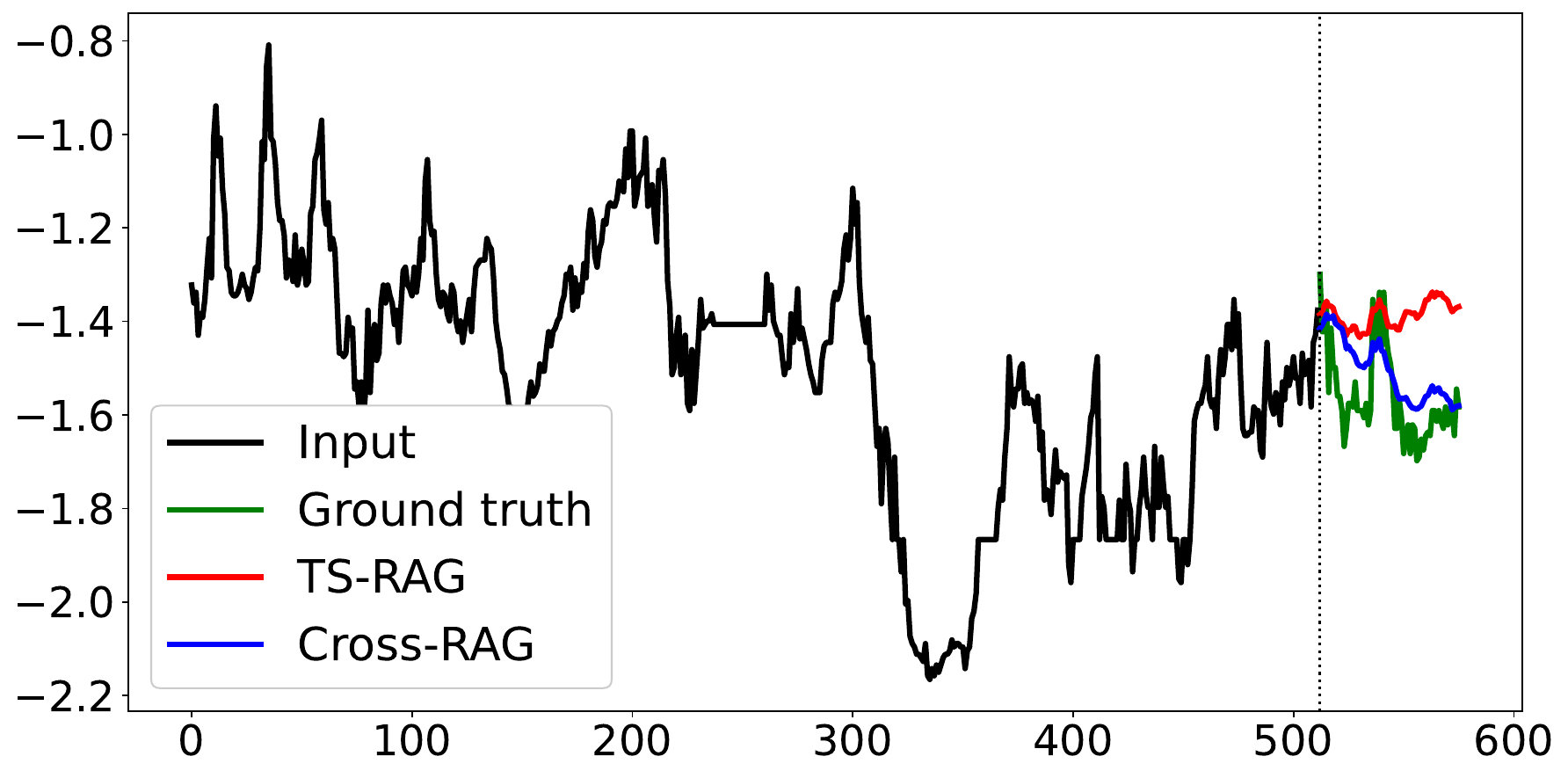}
        \end{adjustbox}
        \caption{Example 2}
    \end{subfigure}
    \begin{subfigure}[t]{0.48\linewidth}
        \centering
        \begin{adjustbox}{max width=\linewidth}
        \includegraphics[width=\textwidth]{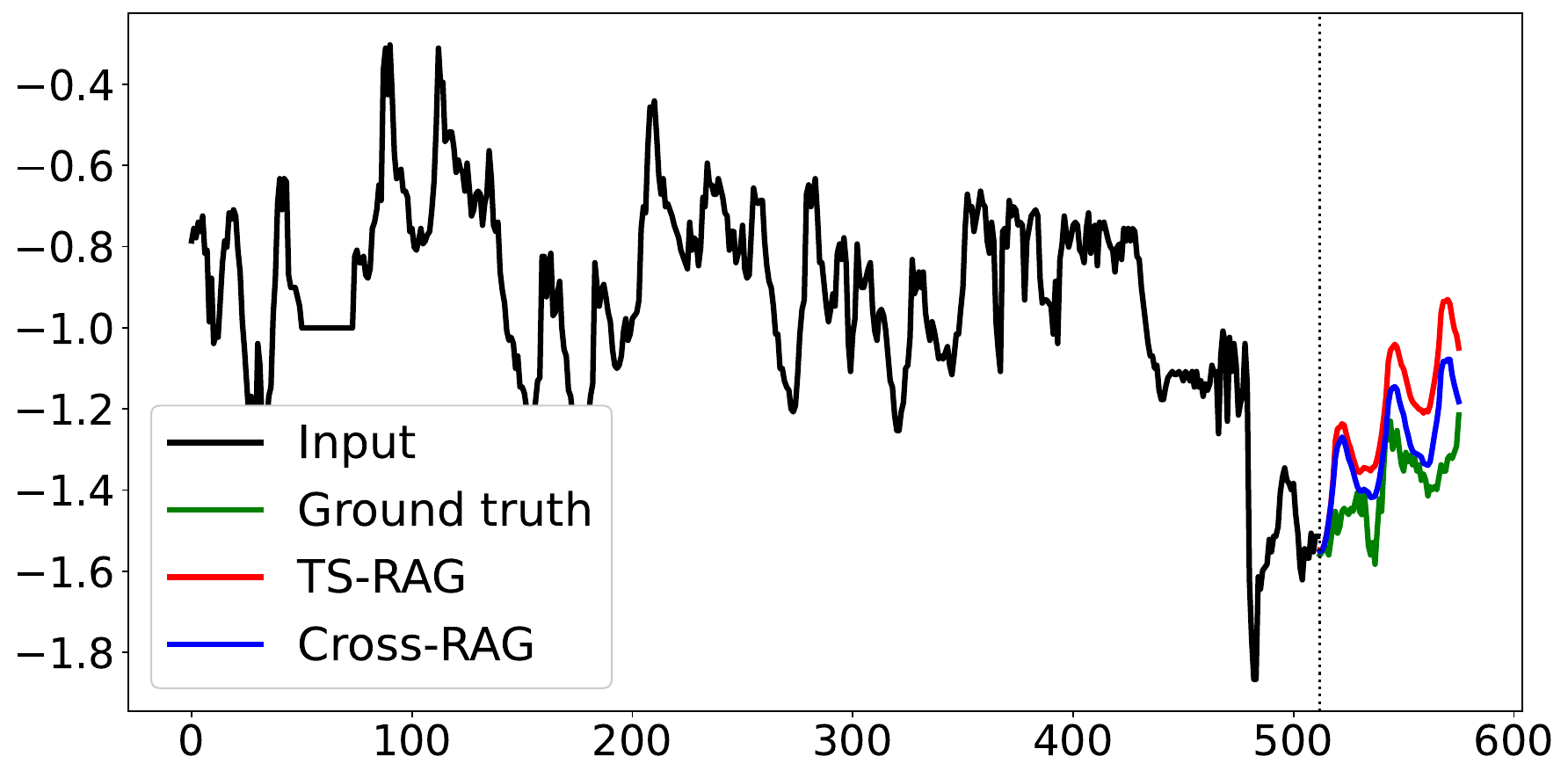}
        \end{adjustbox}
        \caption{Example 3}
    \end{subfigure}
    \begin{subfigure}[t]{0.48\linewidth}
        \centering
        \begin{adjustbox}{max width=\linewidth}
        \includegraphics[width=\textwidth]{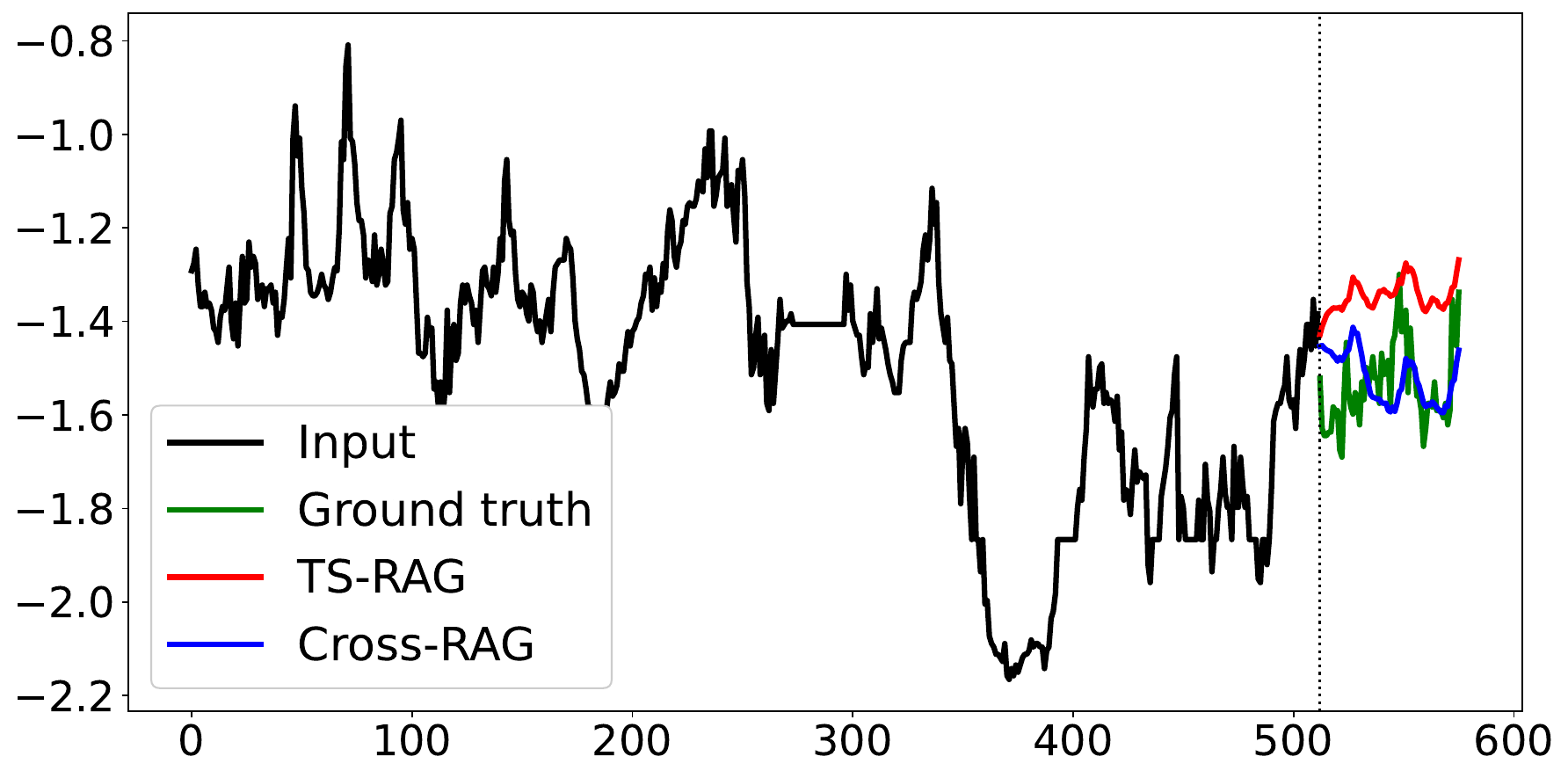}
        \end{adjustbox}
        \caption{Example 4}
    \end{subfigure}
    \begin{subfigure}[t]{0.48\linewidth}
        \centering
        \begin{adjustbox}{max width=\linewidth}
        \includegraphics[width=\textwidth]{figures/etth1/ETTh1_18809.pdf}
        \end{adjustbox}
    \end{subfigure}
    \caption{
    Visualization of zero-shot forecasting on \textbf{ETTh1}.
    }
\end{figure}

\begin{figure}[H]
% \vspace{12pt}
    \centering
    \begin{subfigure}[t]{0.48\linewidth}
        \centering
        \begin{adjustbox}{max width=\linewidth}
        \includegraphics[width=\textwidth]{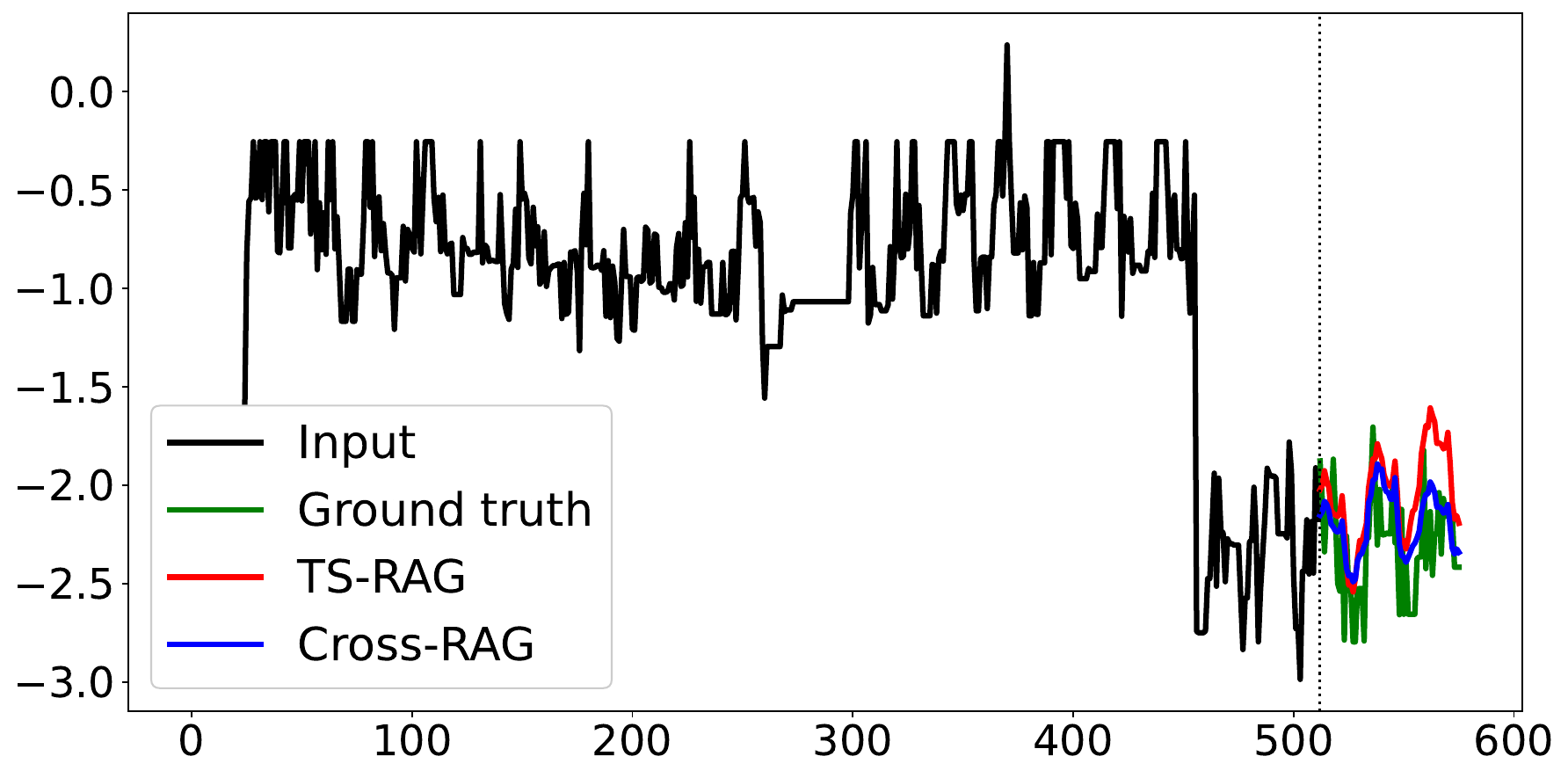}
        \end{adjustbox}
    \end{subfigure}
    \begin{subfigure}[t]{0.48\linewidth}
        \centering
        \begin{adjustbox}{max width=\linewidth}
        \includegraphics[width=\textwidth]{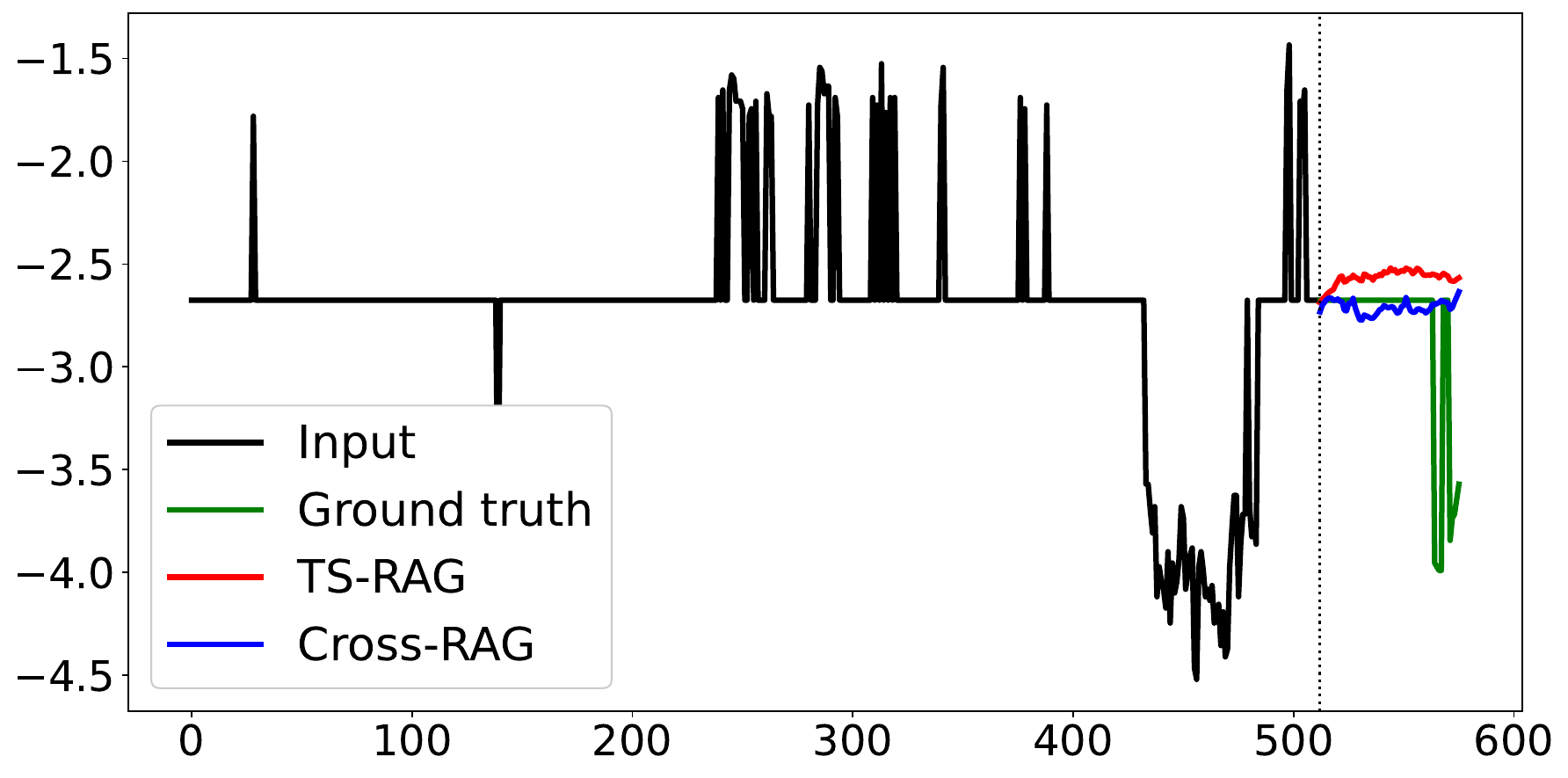}
        \end{adjustbox}
    \end{subfigure}
    \begin{subfigure}[t]{0.48\linewidth}
        \centering
        \begin{adjustbox}{max width=\linewidth}
        \includegraphics[width=\textwidth]{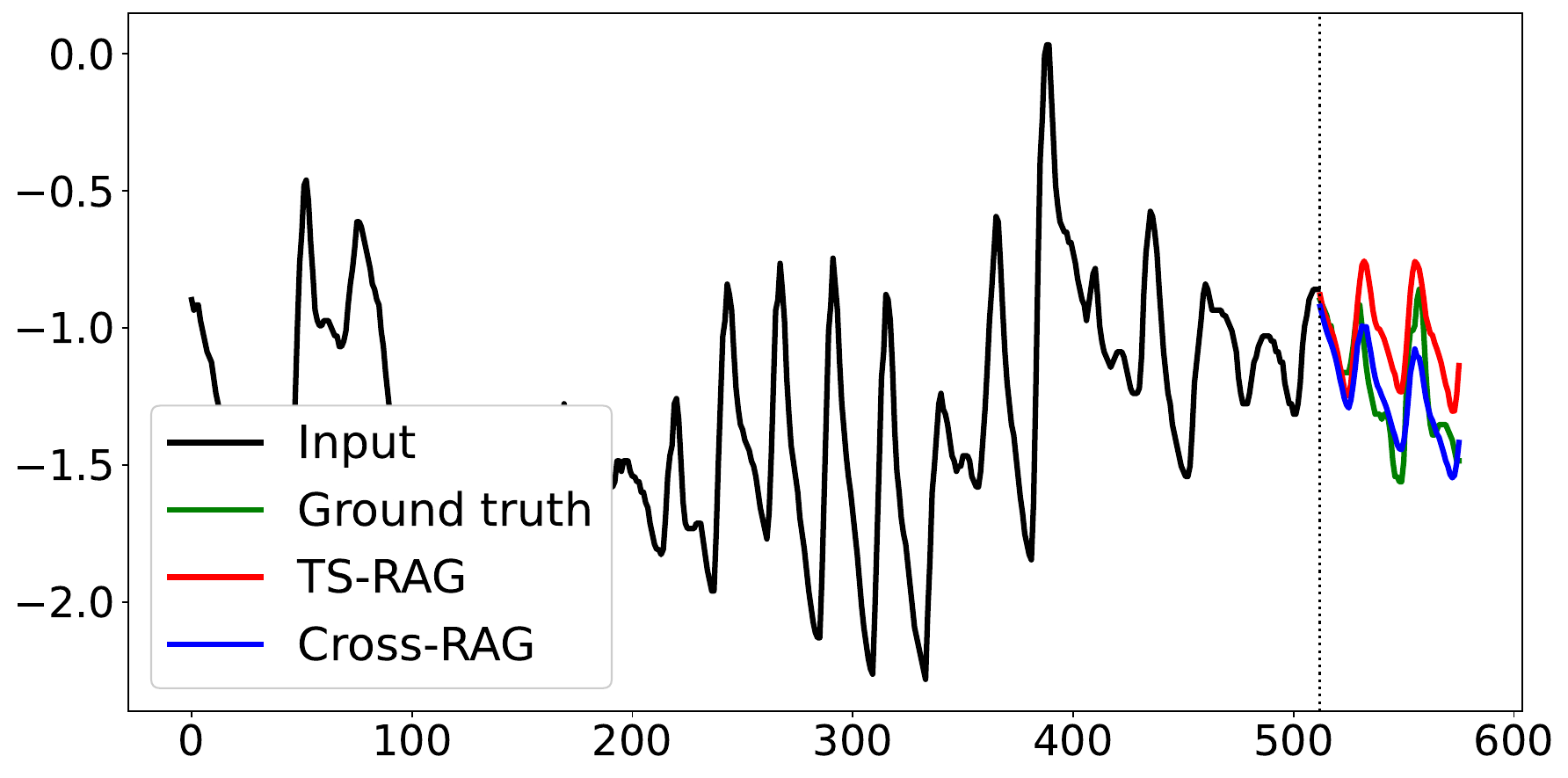}
        \end{adjustbox}
    \end{subfigure}
    \begin{subfigure}[t]{0.48\linewidth}
        \centering
        \begin{adjustbox}{max width=\linewidth}
        \includegraphics[width=\textwidth]{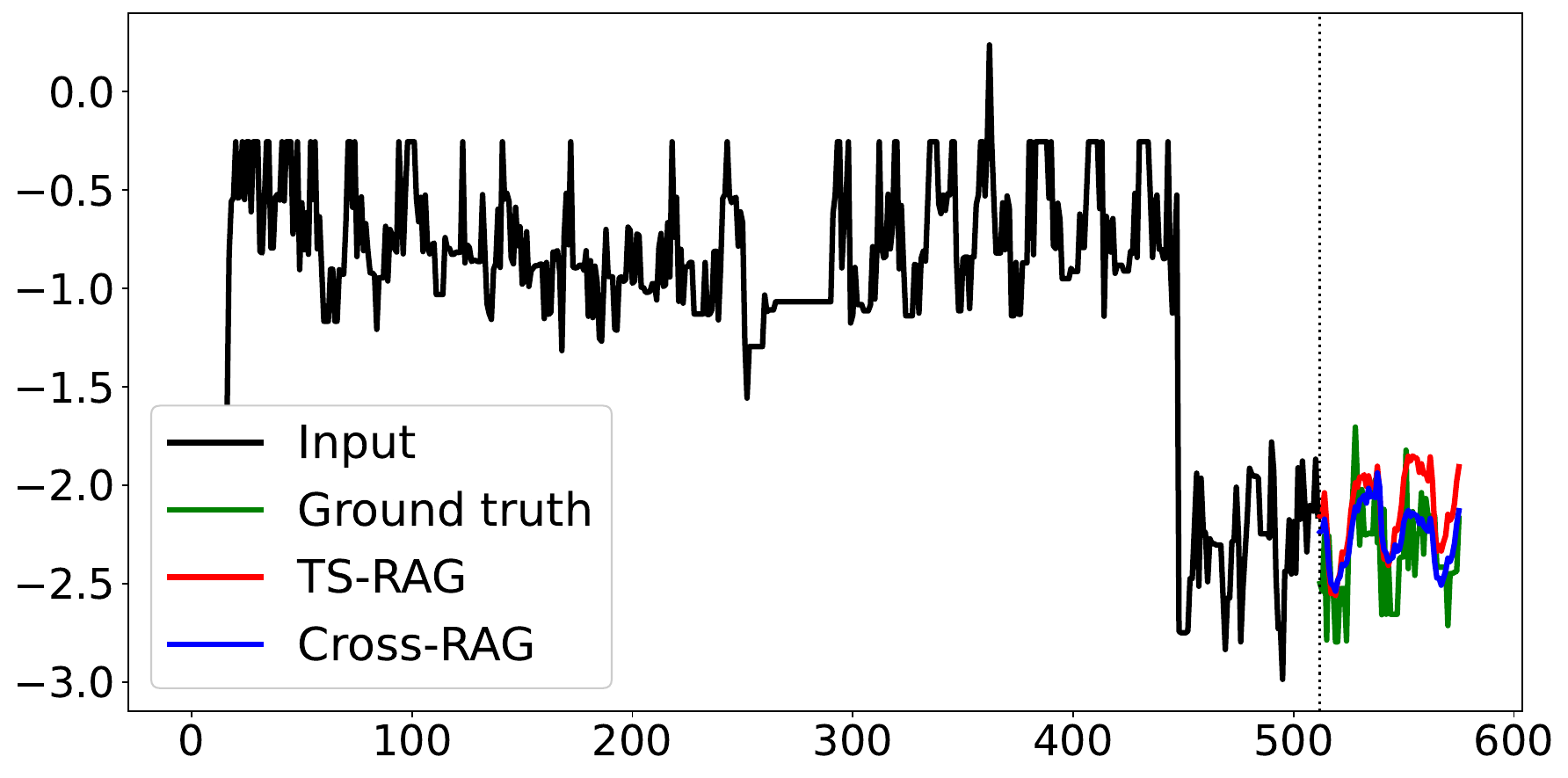}
        \end{adjustbox}
    \end{subfigure}
    \caption{
    Visualization of zero-shot forecasting on \textbf{ETTh2}.
    }
\end{figure}

\begin{figure}[H]
% \vspace{30pt}
    \centering
    \begin{subfigure}[t]{0.48\linewidth}
        \centering
        \begin{adjustbox}{max width=\linewidth}
        \includegraphics[width=\textwidth]{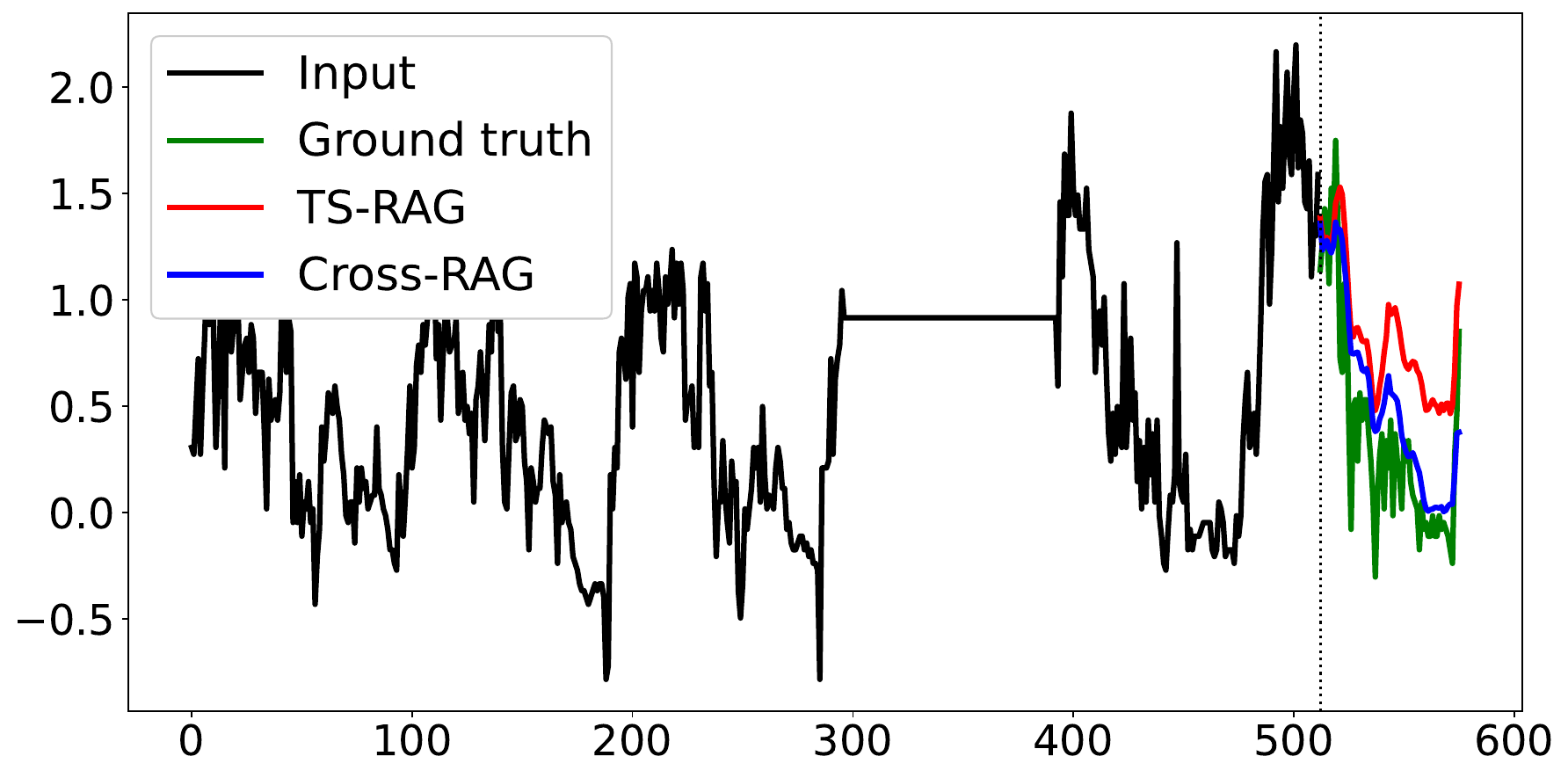}
        \end{adjustbox}
    \end{subfigure}
    \begin{subfigure}[t]{0.48\linewidth}
        \centering
        \begin{adjustbox}{max width=\linewidth}
        \includegraphics[width=\textwidth]{figures/ettm1/ETTm1_52712.pdf}
        \end{adjustbox}
    \end{subfigure}
    \begin{subfigure}[t]{0.48\linewidth}
        \centering
        \begin{adjustbox}{max width=\linewidth}
        \includegraphics[width=\textwidth]{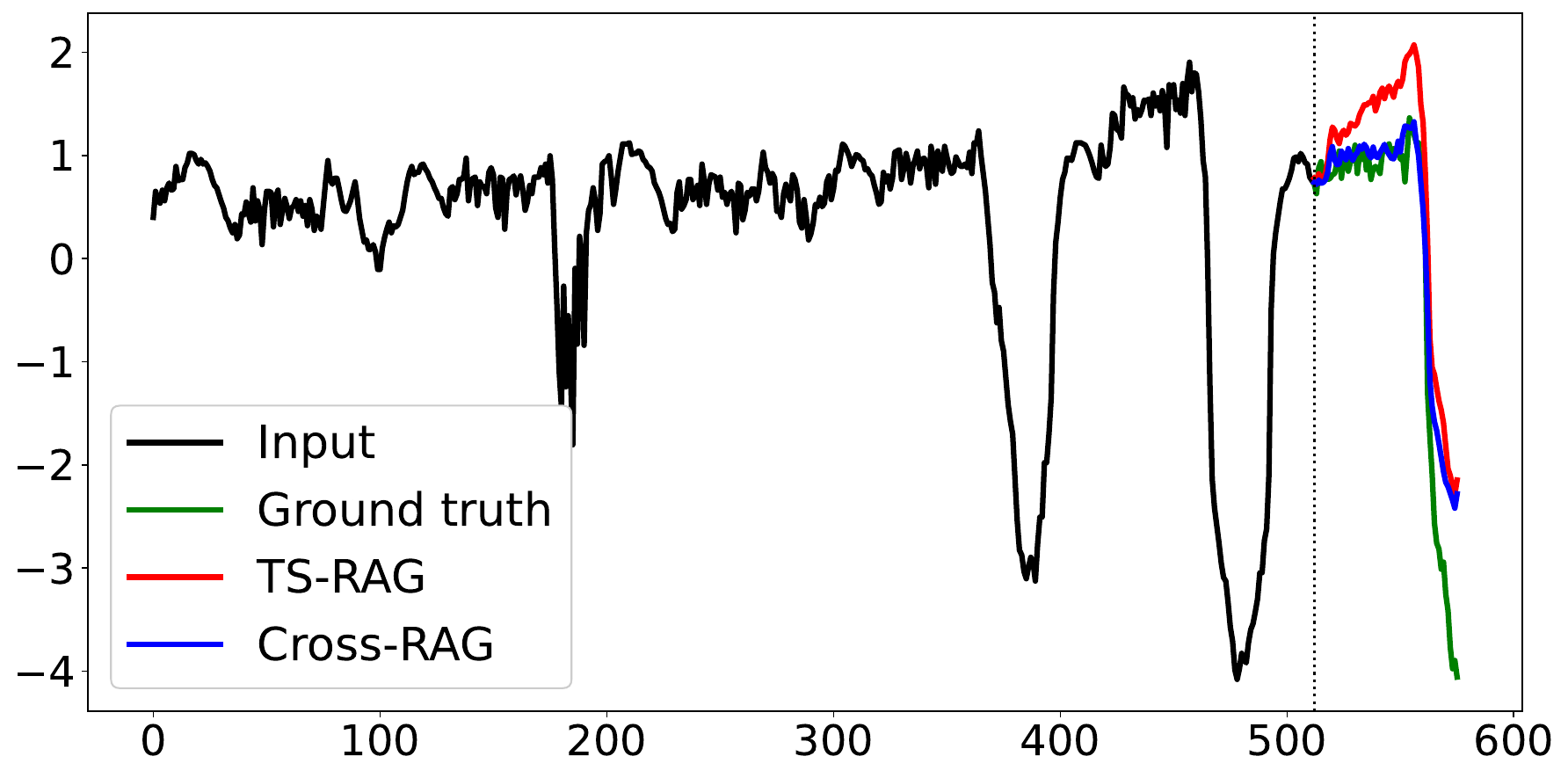}
        \end{adjustbox}
    \end{subfigure}
    \begin{subfigure}[t]{0.48\linewidth}
        \centering
        \begin{adjustbox}{max width=\linewidth}
        \includegraphics[width=\textwidth]{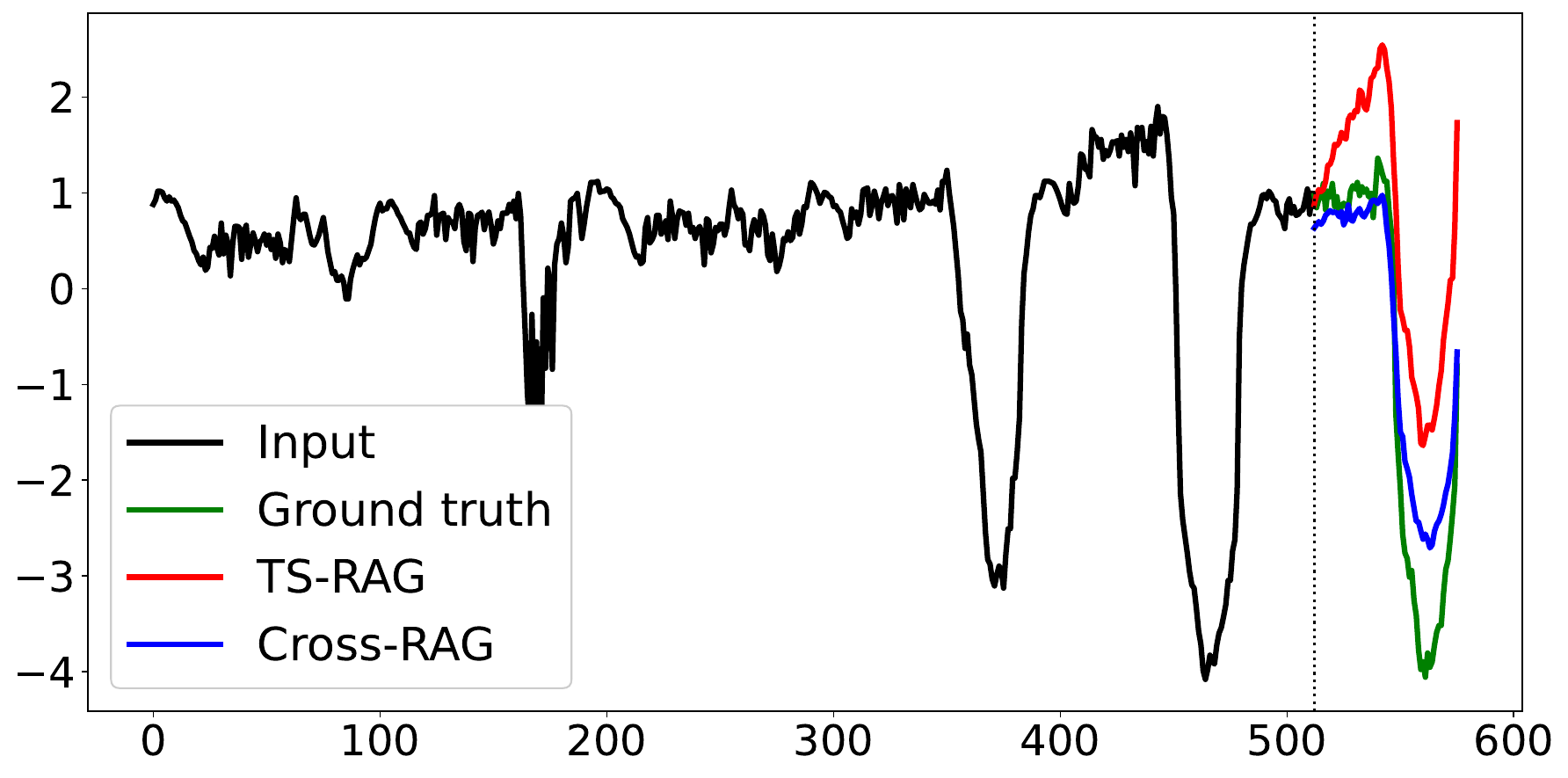}
        \end{adjustbox}
    \end{subfigure}
    \caption{
    Visualization of zero-shot forecasting on \textbf{ETTm1}.
    }
\end{figure}

\begin{figure}[H]
% \vspace{12pt}
    \centering
    \begin{subfigure}[t]{0.48\linewidth}
        \centering
        \begin{adjustbox}{max width=\linewidth}
        \includegraphics[width=\textwidth]{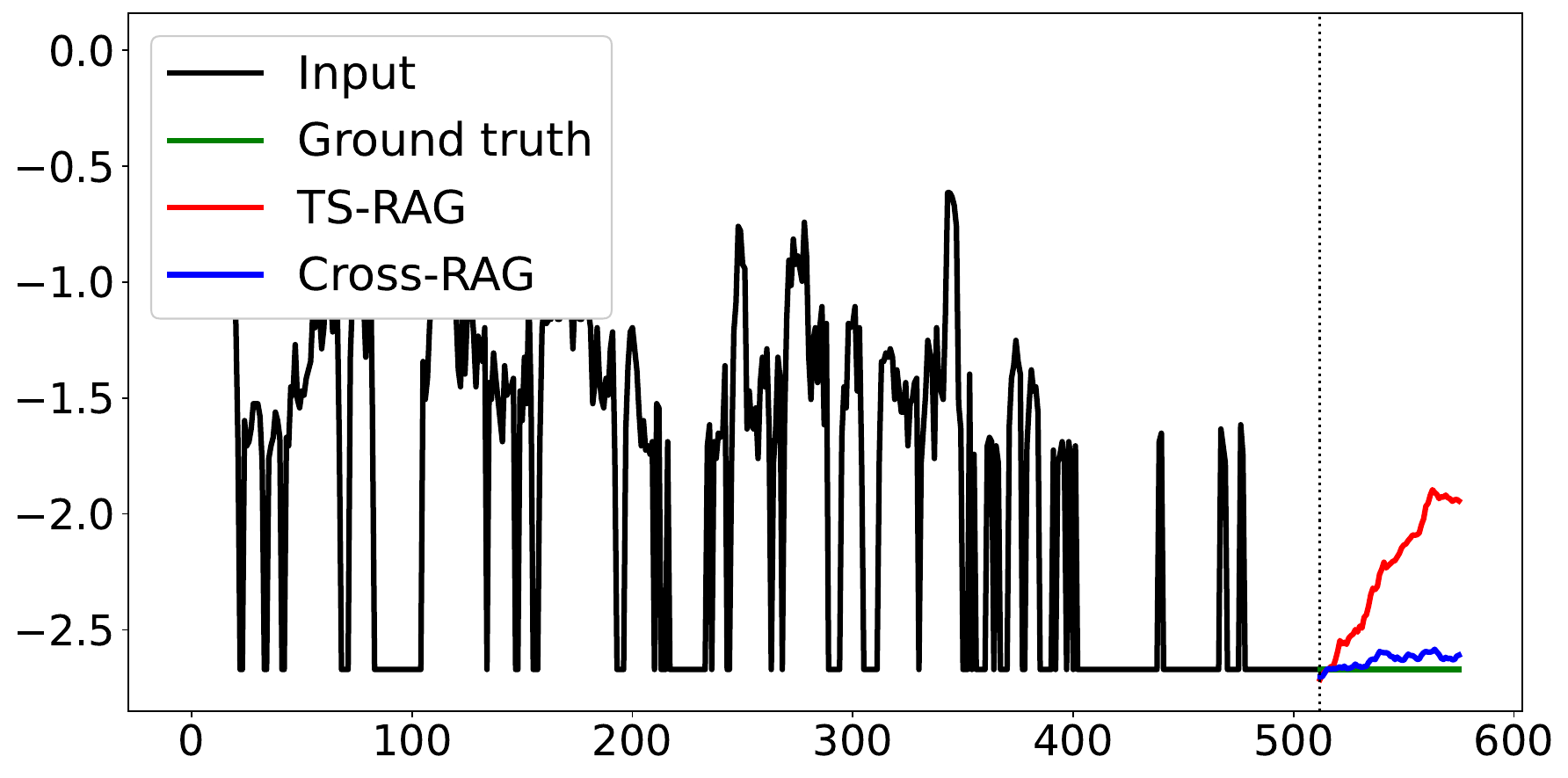}
        \end{adjustbox}
    \end{subfigure}
    \begin{subfigure}[t]{0.48\linewidth}
        \centering
        \begin{adjustbox}{max width=\linewidth}
        \includegraphics[width=\textwidth]{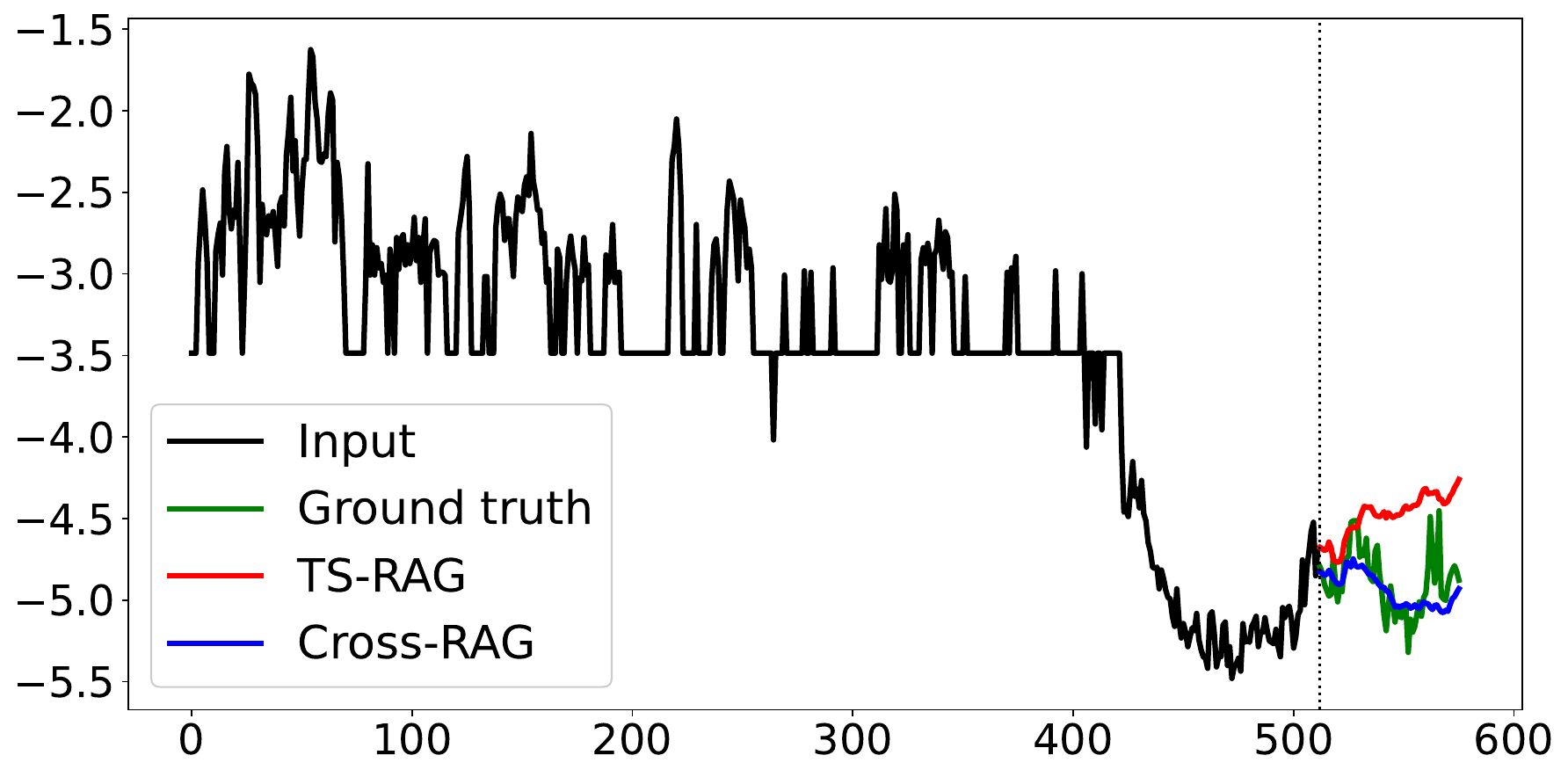}
        \end{adjustbox}
    \end{subfigure}
    \begin{subfigure}[t]{0.48\linewidth}
        \centering
        \begin{adjustbox}{max width=\linewidth}
        \includegraphics[width=\textwidth]{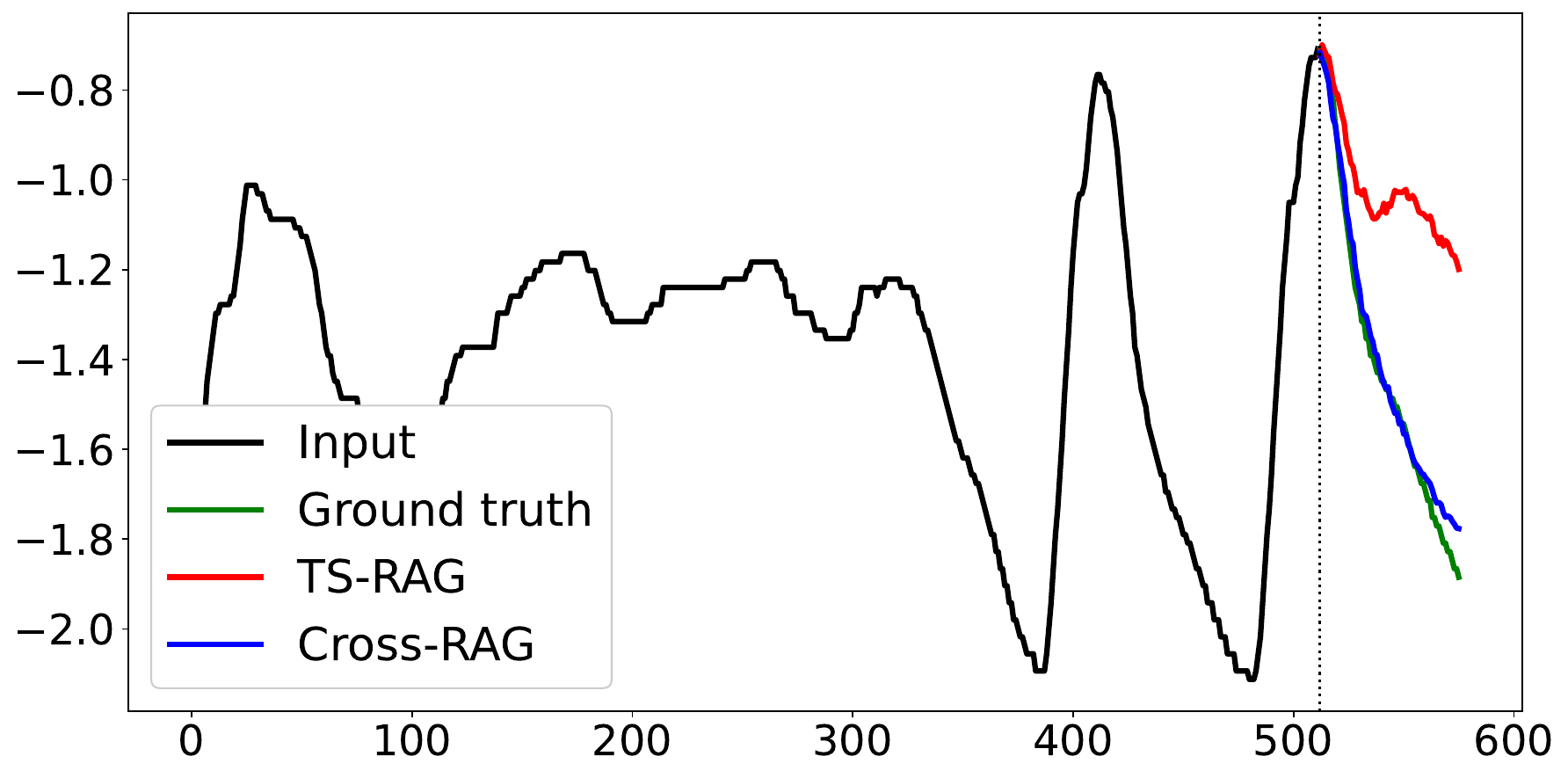}
        \end{adjustbox}
    \end{subfigure}
    \begin{subfigure}[t]{0.48\linewidth}
        \centering
        \begin{adjustbox}{max width=\linewidth}
        \includegraphics[width=\textwidth]{figures/ettm2/ETTm2_75688.pdf}
        \end{adjustbox}
    \end{subfigure}
    \caption{
    Visualization of zero-shot forecasting on \textbf{ETTm2}.
    }
\end{figure}

\begin{figure}[H]
% \vspace{30pt}
    \centering
    \begin{subfigure}[t]{0.48\linewidth}
        \centering
        \begin{adjustbox}{max width=\linewidth}
        \includegraphics[width=\textwidth]{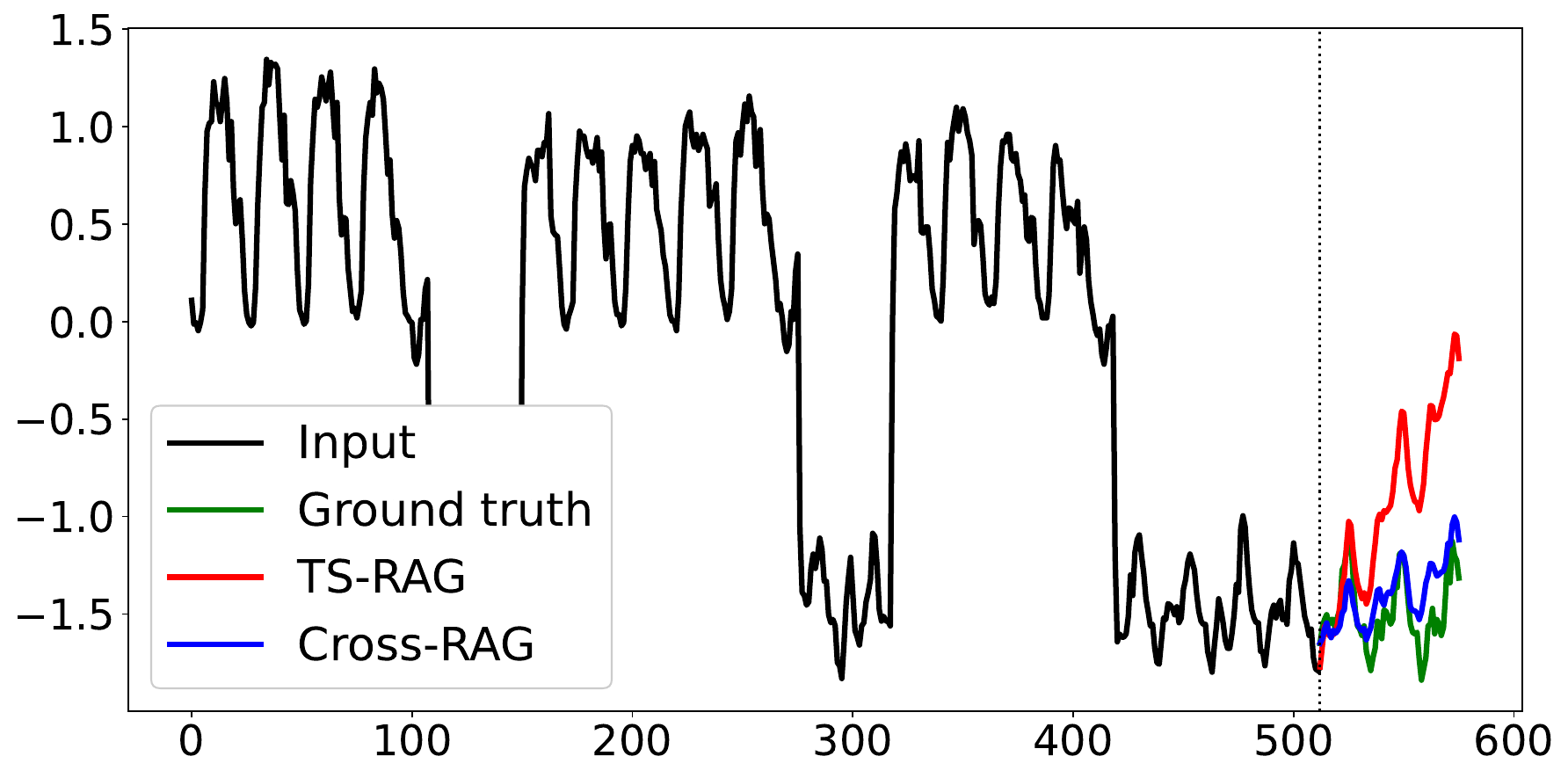}
        \end{adjustbox}
    \end{subfigure}
    \begin{subfigure}[t]{0.48\linewidth}
        \centering
        \begin{adjustbox}{max width=\linewidth}
        \includegraphics[width=\textwidth]{figures/electricity/electricity_33051.pdf}
        \end{adjustbox}
    \end{subfigure}
    \begin{subfigure}[t]{0.48\linewidth}
        \centering
        \begin{adjustbox}{max width=\linewidth}
        \includegraphics[width=\textwidth]{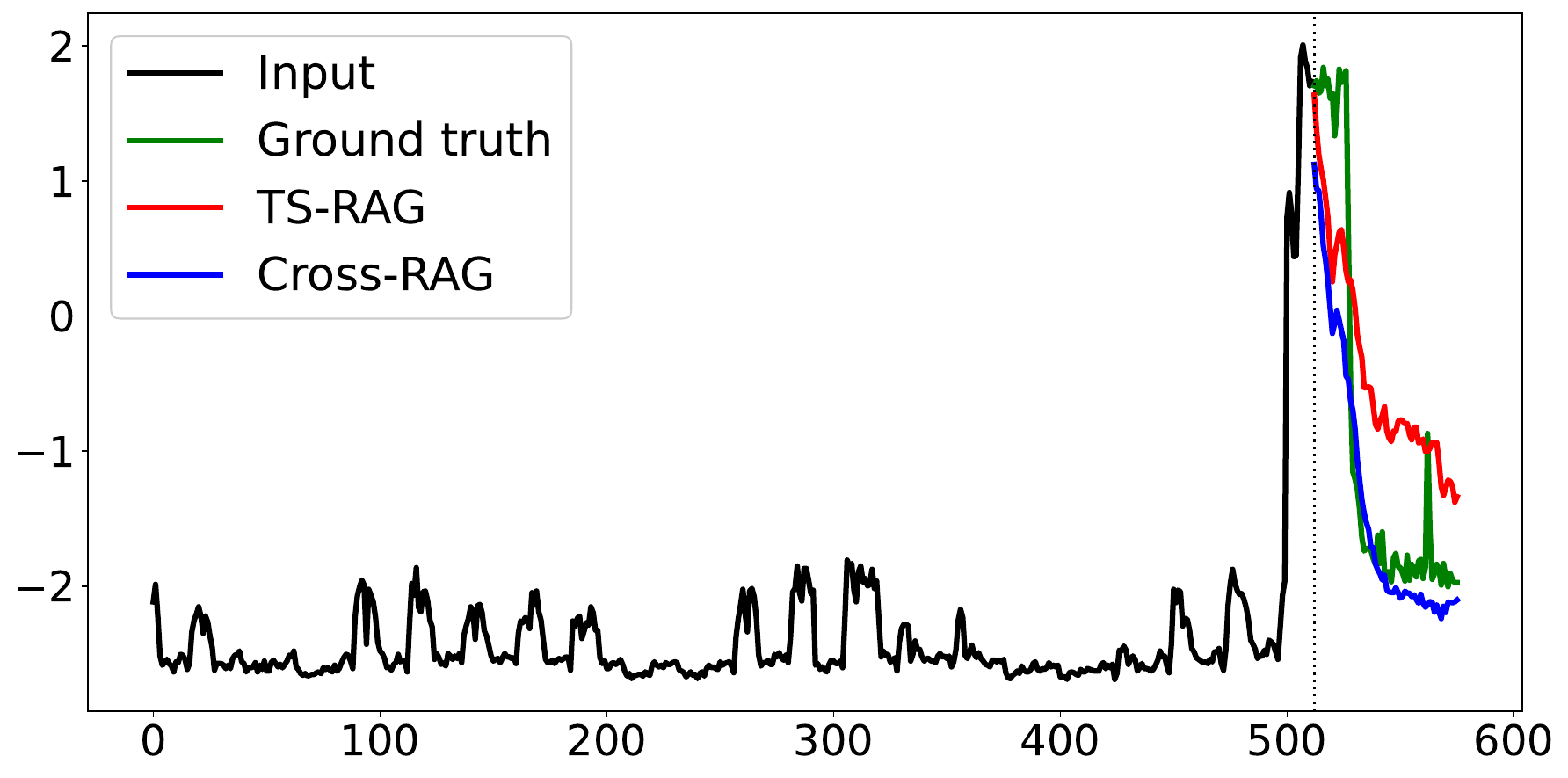}
        \end{adjustbox}
    \end{subfigure}
    \begin{subfigure}[t]{0.48\linewidth}
        \centering
        \begin{adjustbox}{max width=\linewidth}
        \includegraphics[width=\textwidth]{figures/electricity/electricity_544091.pdf}
        \end{adjustbox}
    \end{subfigure}
    \caption{
    Visualization of zero-shot forecasting on \textbf{Electricity}.
    }
\end{figure}

\begin{figure}[H]
% \vspace{12pt}
    \centering
    \begin{subfigure}[t]{0.48\linewidth}
        \centering
        \begin{adjustbox}{max width=\linewidth}
        \includegraphics[width=\textwidth]{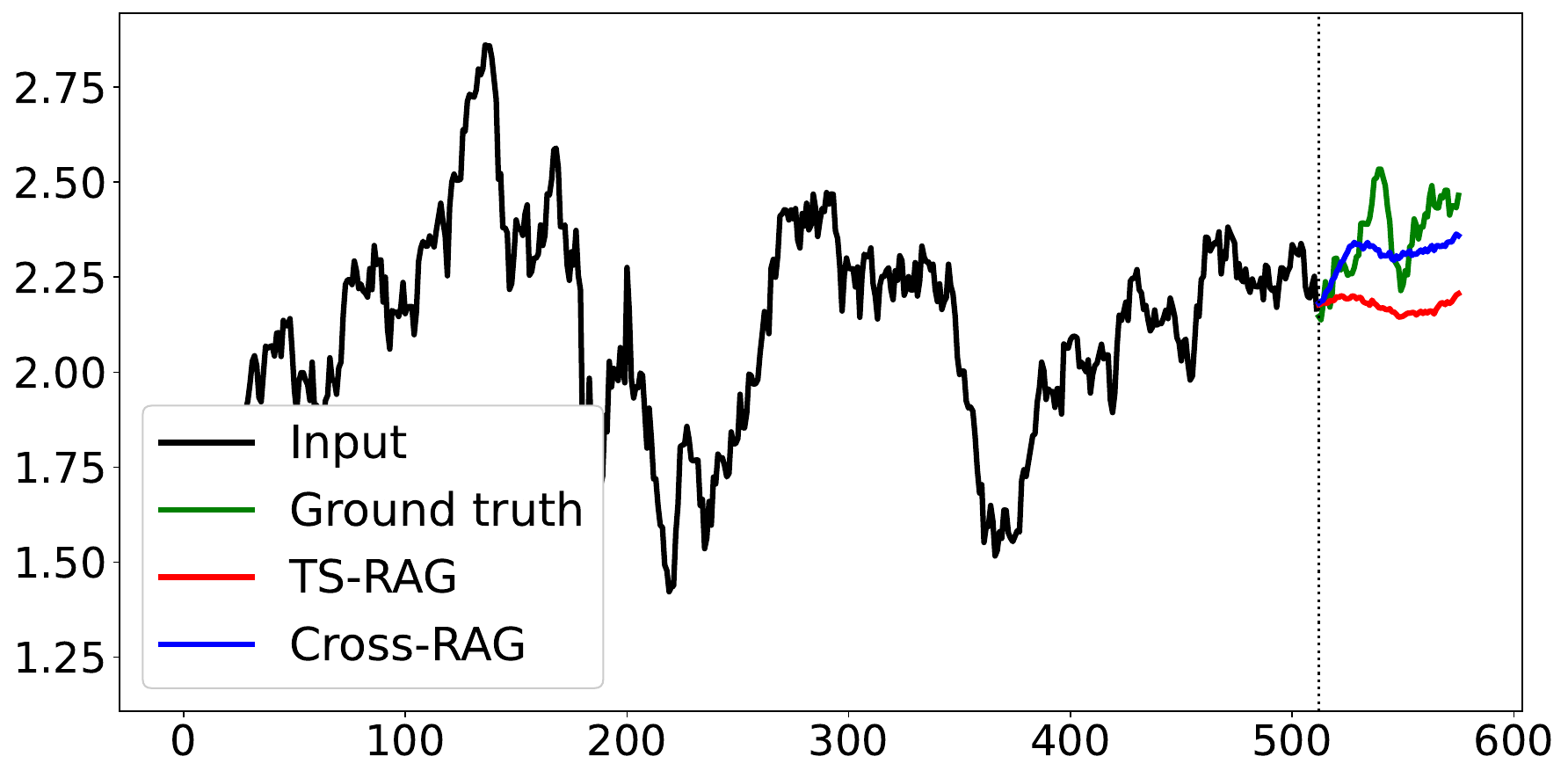}
        \end{adjustbox}
    \end{subfigure}
    \begin{subfigure}[t]{0.48\linewidth}
        \centering
        \begin{adjustbox}{max width=\linewidth}
        \includegraphics[width=\textwidth]{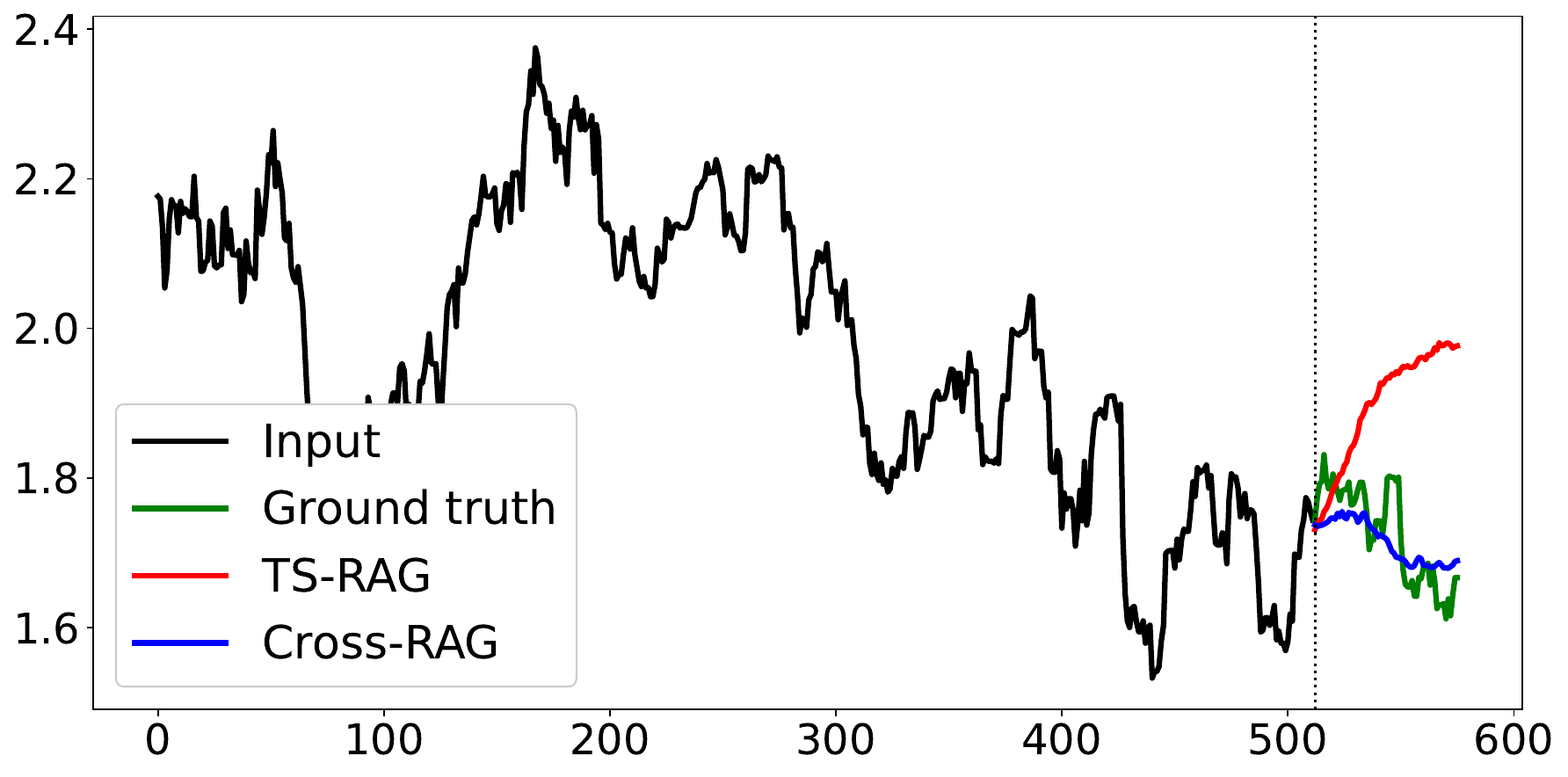}
        \end{adjustbox}
    \end{subfigure}
    \begin{subfigure}[t]{0.48\linewidth}
        \centering
        \begin{adjustbox}{max width=\linewidth}
        \includegraphics[width=\textwidth]{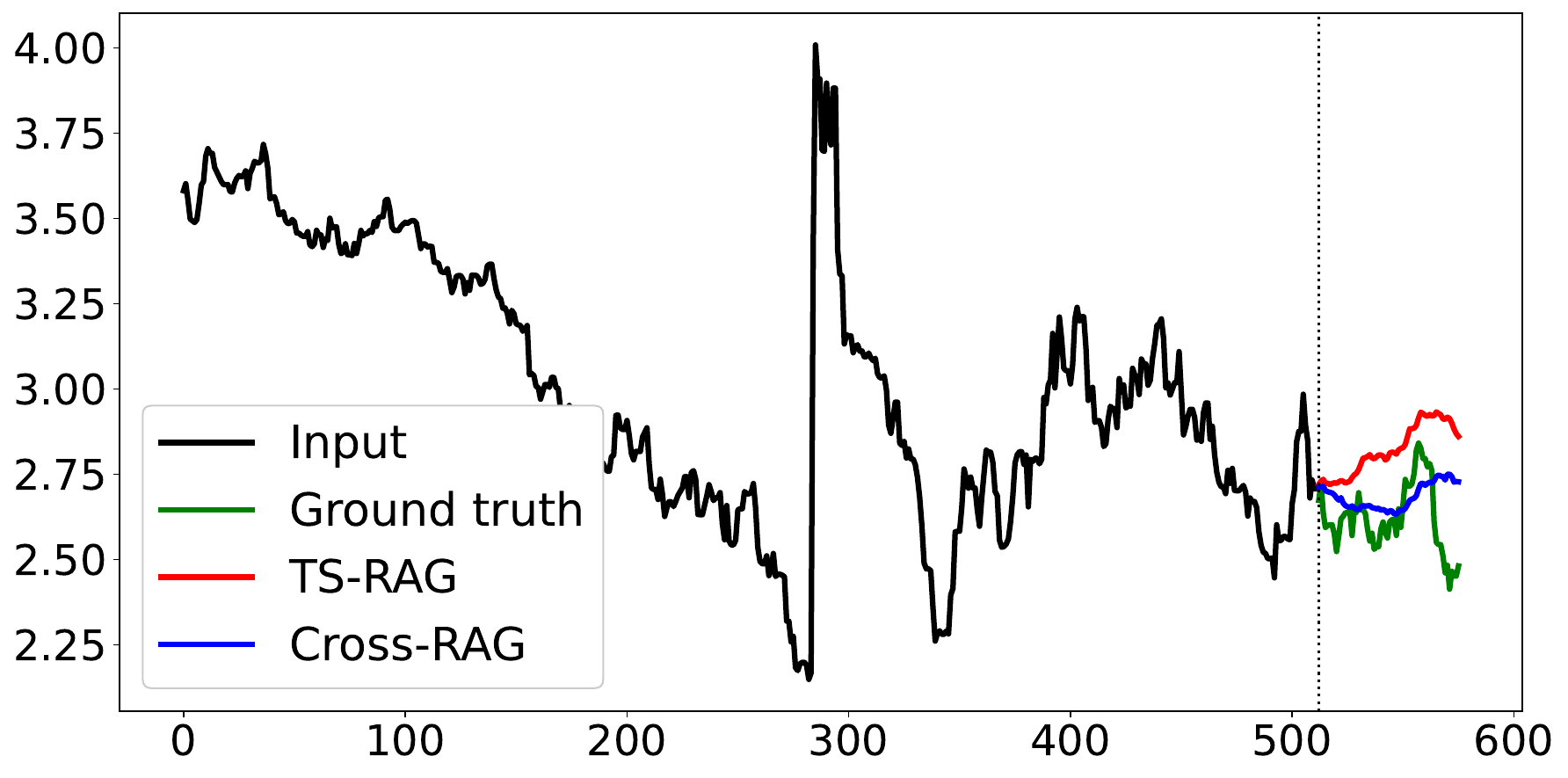}
        \end{adjustbox}
    \end{subfigure}
    \begin{subfigure}[t]{0.48\linewidth}
        \centering
        \begin{adjustbox}{max width=\linewidth}
        \includegraphics[width=\textwidth]{figures/exchange/exchange_7549.pdf}
        \end{adjustbox}
    \end{subfigure}
    \caption{
    Visualization of zero-shot forecasting on \textbf{Exchange}.
    }
\end{figure}

\begin{figure}[H]
% \vspace{30pt}
    \centering
    \begin{subfigure}[t]{0.48\linewidth}
        \centering
        \begin{adjustbox}{max width=\linewidth}
        \includegraphics[width=\textwidth]{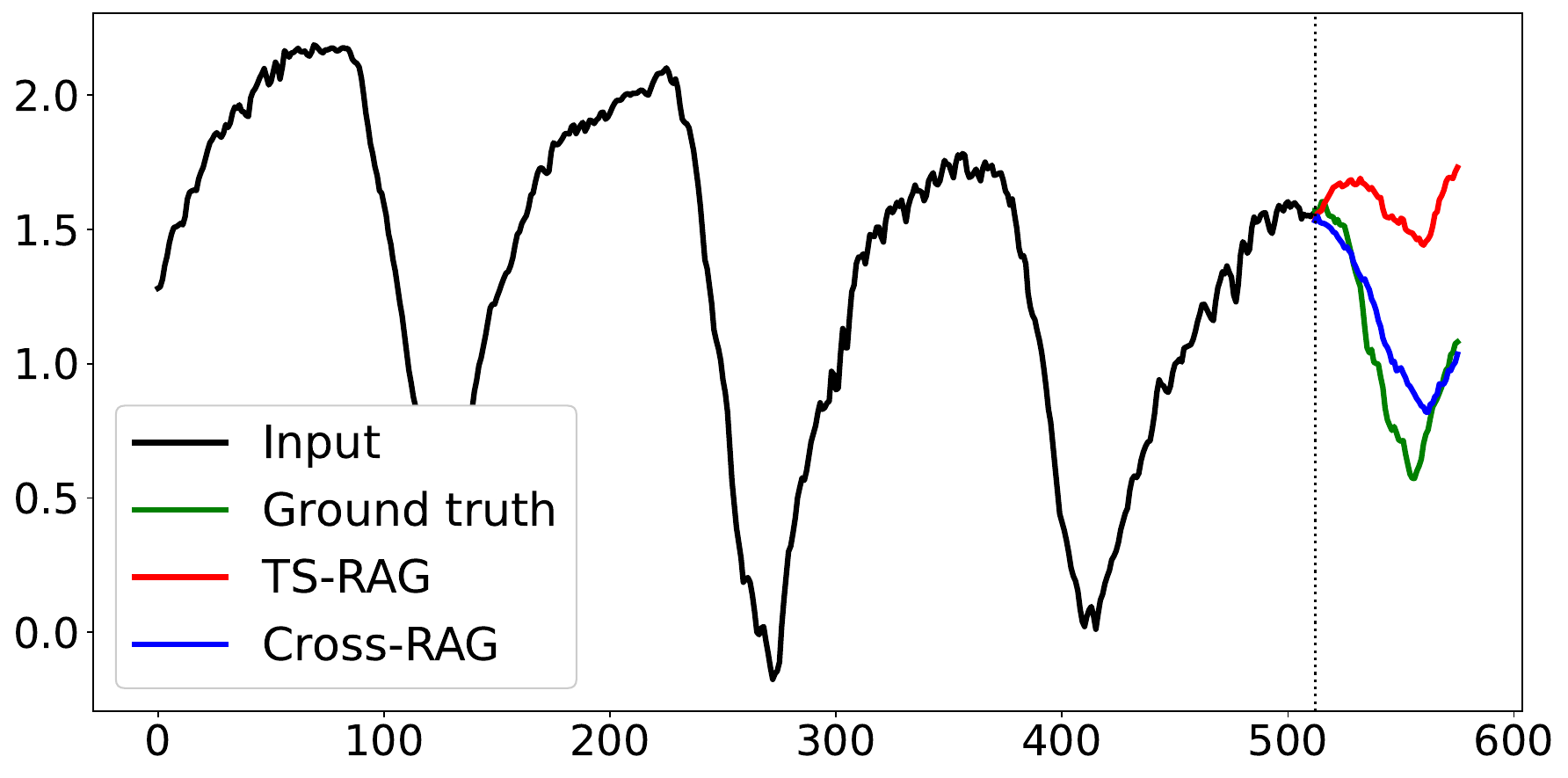}
        \end{adjustbox}
    \end{subfigure}
    \begin{subfigure}[t]{0.48\linewidth}
        \centering
        \begin{adjustbox}{max width=\linewidth}
        \includegraphics[width=\textwidth]{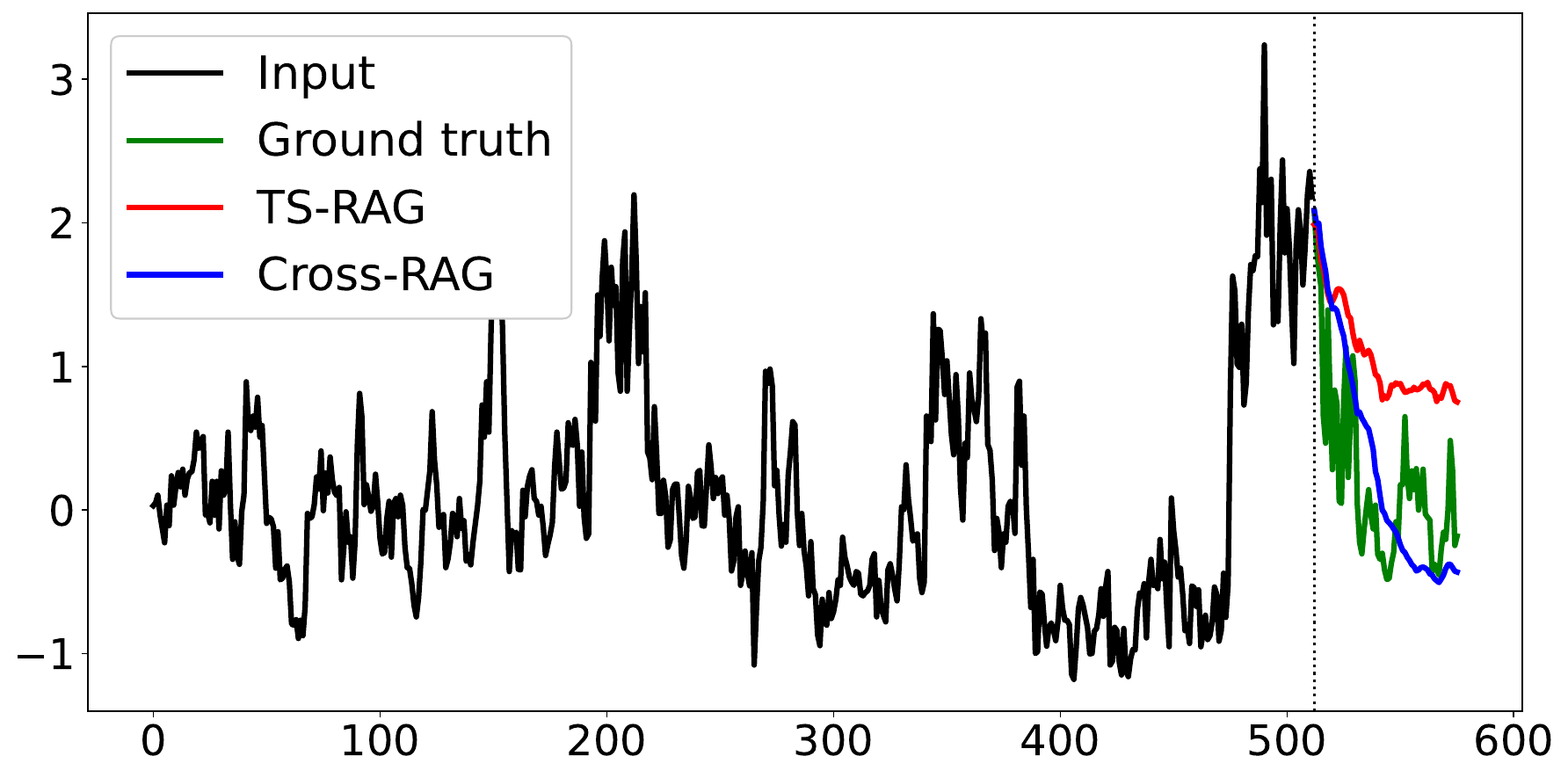}
        \end{adjustbox}
    \end{subfigure}
    \begin{subfigure}[t]{0.48\linewidth}
        \centering
        \begin{adjustbox}{max width=\linewidth}
        \includegraphics[width=\textwidth]{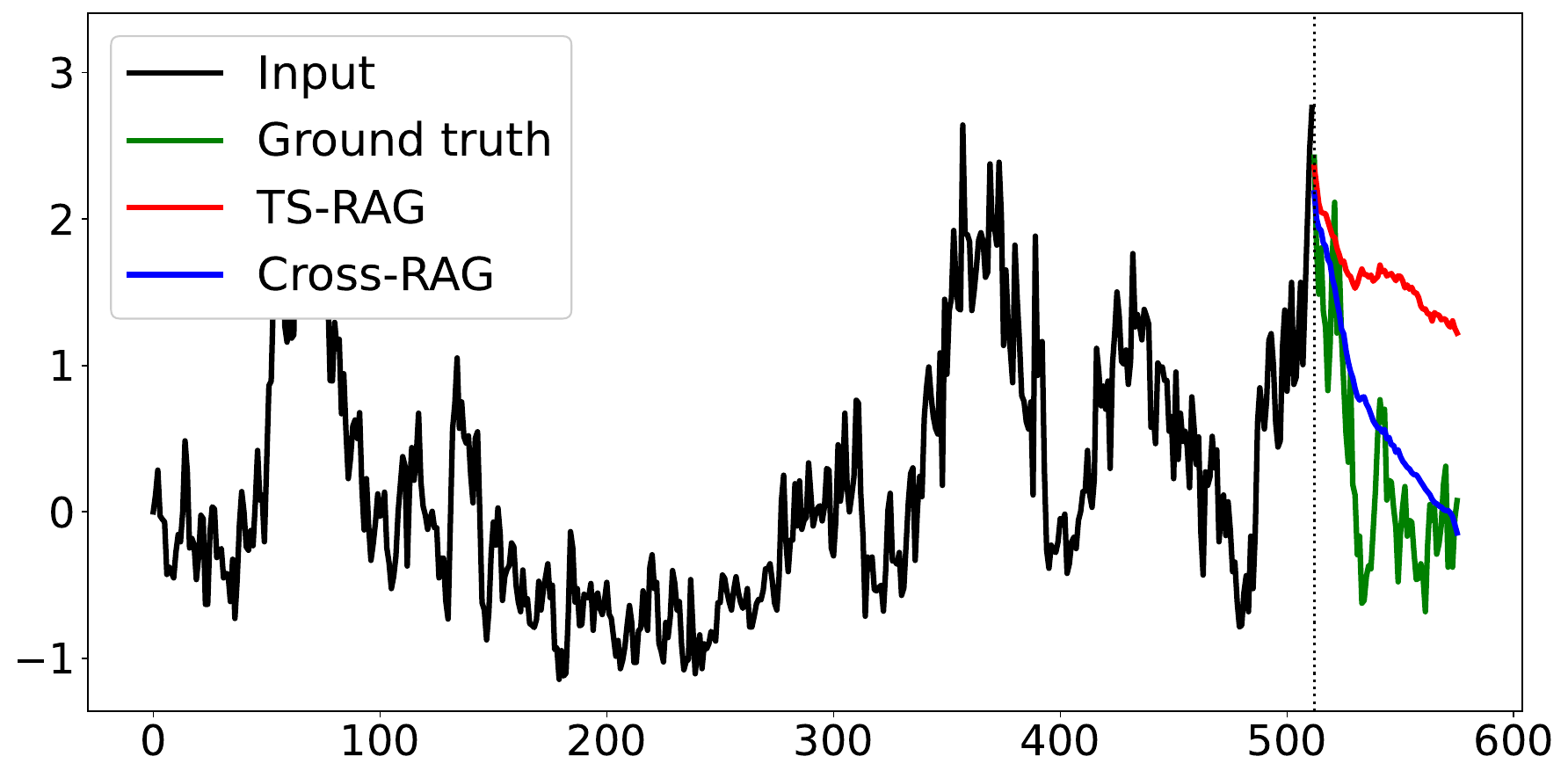}
        \end{adjustbox}
    \end{subfigure}
    \begin{subfigure}[t]{0.48\linewidth}
        \centering
        \begin{adjustbox}{max width=\linewidth}
        \includegraphics[width=\textwidth]{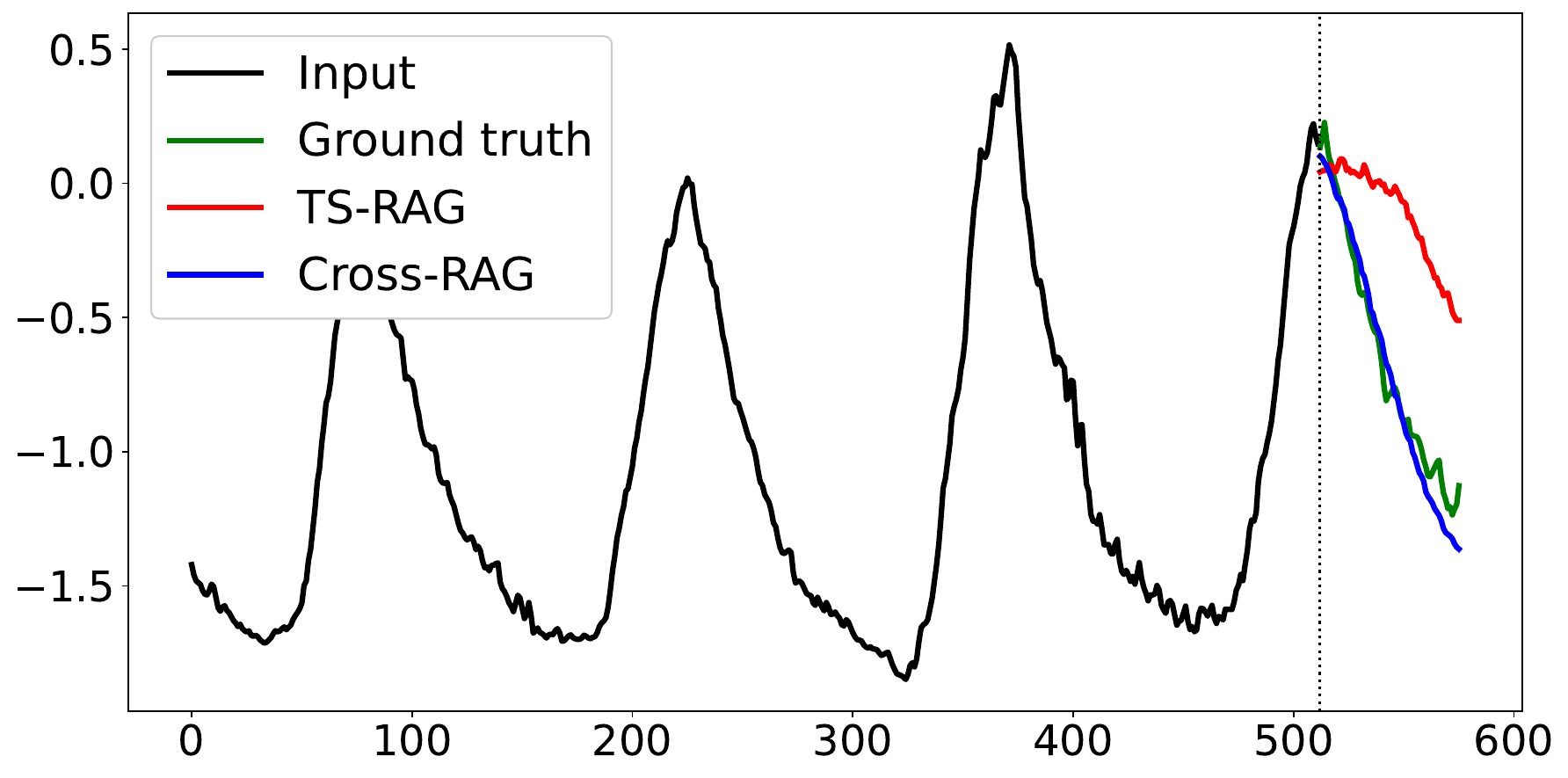}
        \end{adjustbox}
    \end{subfigure}
    \begin{subfigure}[t]{0.48\linewidth}
        \centering
        \begin{adjustbox}{max width=\linewidth}
        \includegraphics[width=\textwidth]{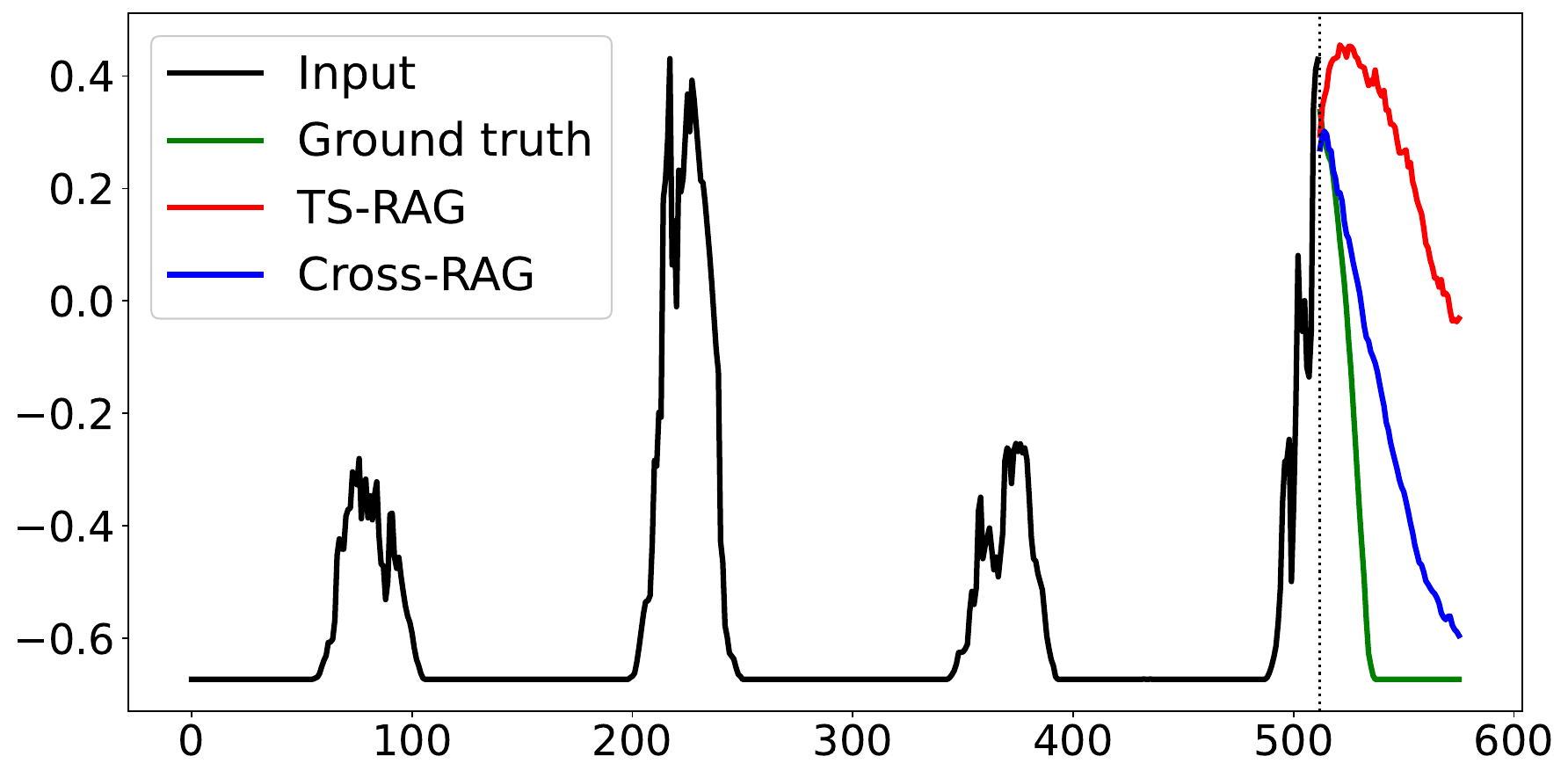}
        \end{adjustbox}
    \end{subfigure}
    \begin{subfigure}[t]{0.48\linewidth}
        \centering
        \begin{adjustbox}{max width=\linewidth}
        \includegraphics[width=\textwidth]{figures/weather/weather_21038.pdf}
        \end{adjustbox}
    \end{subfigure}
    \caption{
    Visualization of zero-shot forecasting on \textbf{Weather}.
    }
\end{figure}